%% file: main.tex
\documentclass{article} 
\usepackage{CJKutf8}
\usepackage{colm2024_conference}
\usepackage{microtype}
\usepackage{hyperref}
\usepackage{url}
\usepackage{booktabs}
\usepackage{CJKutf8}
\usepackage{graphicx}
\usepackage{multicol}
\usepackage{multirow}
\usepackage{multirow}
\usepackage{lineno}
\usepackage{array}
\usepackage[table]{xcolor}
\usepackage{booktabs}
\usepackage{enumitem}
\usepackage{graphicx}
\usepackage{wrapfig}
\usepackage{algorithm}
\usepackage{algpseudocode}
\usepackage{microtype}
\usepackage{amsmath}
\usepackage{amssymb}
\usepackage{colortbl}
\usepackage[table]{xcolor}
\usepackage[utf8]{inputenc}
\definecolor{lightgray}{rgb}{0.9,0.9,0.9}
\usepackage{caption}
\usepackage{subcaption}
\usepackage{xcolor}
\usepackage{setspace}
\usepackage{colortbl}
\usepackage{tabularx}
\usepackage{blindtext}
\usepackage{pgfplots}
\pgfplotsset{compat=1.18} 
\usepackage{tikz}
\usetikzlibrary{er,positioning,bayesnet}
\usepackage{makecell}
\usepackage{tipa}
\usepackage{siunitx}
\usepackage{nicefrac}
\usepackage{tocloft}
\usepackage{listings}
\usepackage{tcolorbox}
\usepackage{xltabular}
\usepackage{adjustbox}
\usepackage{xurl}
\usepackage{lipsum}

\AtEndPreamble{
    \usepackage[capitalize]{cleveref}
    \crefname{section}{Sec.}{Secs.}
    \Crefname{section}{Section}{Sections}
    \Crefname{table}{Table}{Tables}
    \crefname{table}{Tab.}{Tabs.}
}

\title{HunyuanOCR Technical Report}

\author{
	\bf \large Tencent Hunyuan Vision Team
}

\begin{document}
\begin{CJK*}{UTF8}{gbsn}
    \maketitle

    \begin{center}
        \begin{tabular}{c l}
            \raisebox{-0.15\height}{\includegraphics[height=12pt]{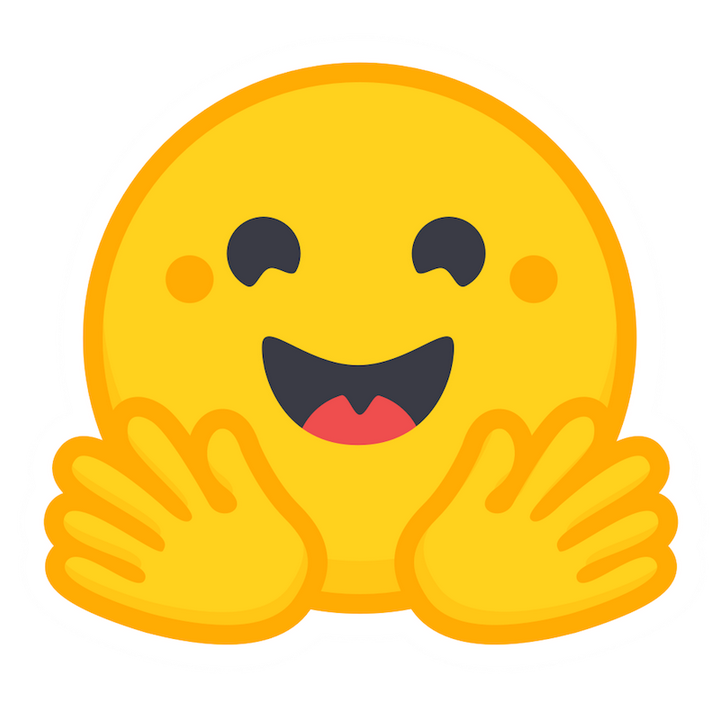}} &
            \href{https://huggingface.co/tencent/HunyuanOCR}{https://huggingface.co/tencent/HunyuanOCR}
            \\[6pt]
            \raisebox{-0.15\height}{\includegraphics[height=12pt]{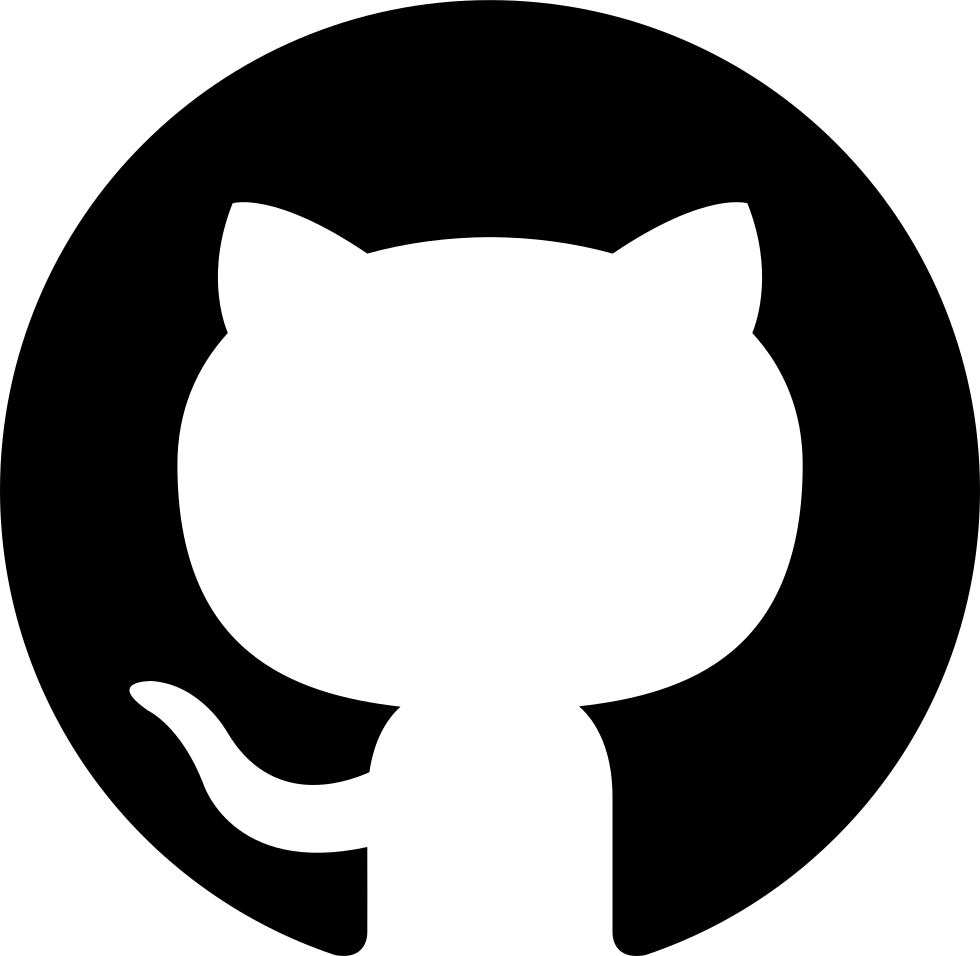}}       &
            \href{https://github.com/Tencent-Hunyuan/HunyuanOCR}{https://github.com/Tencent-Hunyuan/HunyuanOCR}
        \end{tabular}
    \end{center}

    \input{sec/abstract}

    \input{sec/introduction}

    \input{sec/relatedwork}

    \input{sec/model_design}

    \input{sec/data_construction}

    \input{sec/training_recipe}

    \input{sec/evaluation}

    \input{sec/conclusion}

    \bibliography{colm2024_conference}
    \bibliographystyle{colm2024_conference}

    \input{sec/appendix}

\end{CJK*}
\end{document}

%% file: sec/abstract.tex
\begin{figure}[!ht]
    \centering
    \includegraphics[width=0.9\linewidth]{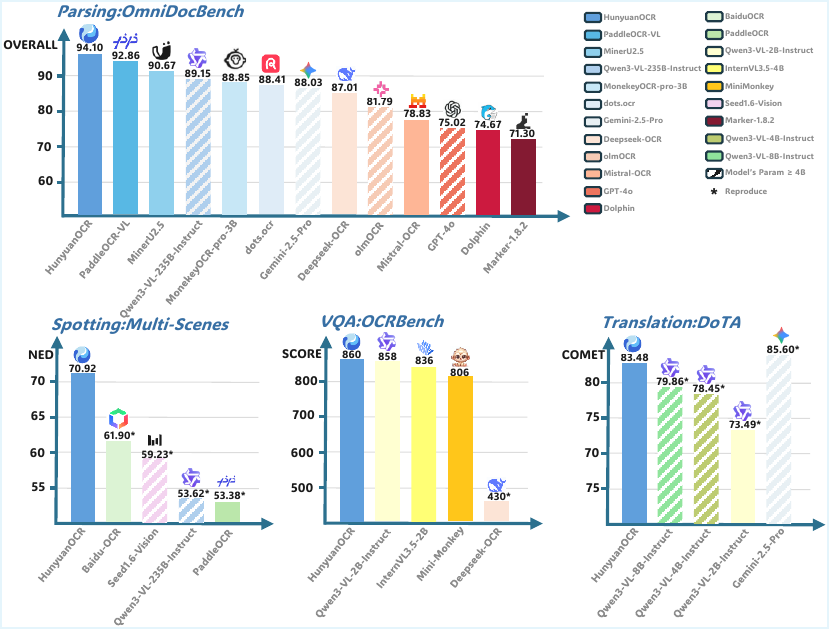}
    \vspace{-0.2em}
    \caption{Performance comparison of HunyuanOCR and other SOTA models.}
    \label{fig:pipeline}
\end{figure}

\begin{abstract}

    This paper presents HunyuanOCR, a commercial-grade, open-source, and lightweight (1B parameters) Vision-Language Model (VLM) dedicated to OCR tasks. The architecture comprises a Native Vision Transformer (ViT) and a lightweight LLM connected via an MLP adapter. HunyuanOCR demonstrates superior performance, outperforming commercial APIs, traditional pipelines, and larger models (e.g., Qwen3-VL-4B). Specifically, it surpasses current public solutions in perception tasks (Text Spotting, Parsing) and excels in semantic tasks (IE, Text Image Translation), securing first place in the ICDAR 2025 DIMT Challenge (Small Model Track). Furthermore, it achieves state-of-the-art (SOTA) results on OCRBench among VLMs with fewer than 3B parameters.

    HunyuanOCR achieves breakthroughs in three key aspects: 1) Unifying Versatility and Efficiency: We implement comprehensive support for core capabilities, including spotting, parsing, IE, VQA, and translation within a lightweight framework. This addresses the limitations of narrow ``OCR expert models" and inefficient ``General VLMs". 2) Streamlined End-to-End Architecture: Adopting a pure end-to-end paradigm eliminates dependencies on pre-processing modules (e.g., layout analysis). This fundamentally resolves error propagation common in traditional pipelines and simplifies system deployment. 3) Data-Driven and RL Strategies: We confirm the critical role of high-quality data and, for the first time in the industry, demonstrate that Reinforcement Learning (RL) strategies yield significant performance gains in OCR tasks.

    HunyuanOCR is officially open-sourced on HuggingFace. We also provide a high-performance deployment solution based on vLLM, placing its production efficiency in the top tier. We hope this model will advance frontier research and provide a solid foundation for industrial applications.
\end{abstract}

%% file: sec/introduction.tex
\section{Introduction}
Modern Optical Character Recognition (OCR)~\cite{long2021scene,qin2024parsing} is a fundamental technology of artificial intelligence that continues to play an important role in promoting digitalization and industrial automation. 
Traditionally, OCR has mainly focused on extracting text from scanned document images and converting it into machine-readable data.
In recent years, with the rapid development of deep learning and multimodal large language model technologies~\cite{zhou2017east,bgshi2017crnn,yin2024survey,liu2024ocrb}, advanced OCR systems have broken through the limitations of scanned documents, now handling diverse layouts, casually captured images, and multilingual as well as handwritten text.
Simultaneously, the scope of OCR tasks has expanded to include more challenging capabilities such as complex document parsing, end-to-end information extraction, text-centric visual question answering, and text image translation.
Driven by technological innovation, the applications of intelligent OCR have permeated various aspects of industry and everyday life.
For example, in office and educational settings~\cite{adeshola2024edu}, OCR enables functions such as literature translation and subject-specific tutoring.
In the healthcare field~\cite{wang2025medical},  OCR facilitates the digital archiving of medical records and correlation analysis, supporting the provision of valuable treatment and health management advice to patients.
Even more significantly, OCR systems fill a critical gap in acquiring high-quality corpora for Large Language Models, acting as an essential instrument for unlocking the content of specialized books and historical archives~\cite{zhang2024docparsing}.

To address diverse application requirements, the industry has long adopted pipeline-based frameworks, including PaddleOCR~\cite{cui2025paddleocr}, EasyOCR~\cite{jaidedai2020easyocr}, and MMOCR~\cite{kuang2021mmocr}. These approaches construct a sequential processing pipeline by integrating multiple compact expert models, offering benefits such as high modularity and the capacity for task-specific optimization. As a result, they exhibit considerable flexibility in applications like text spotting, document parsing, and translation. Nevertheless, the cascaded structure of multiple models introduces inherent drawbacks, including error propagation and elevated development and maintenance overhead.
Recently, with the progress in visual-language models (VLMs), a number of specialized open-source models for OCR and document parsing have been introduced, such as MonkeyOCR~\cite{li2025monkeyocr}, Dots.OCR~\cite{dot2024ocr}, MinerU2.5~\cite{niu2025mineru25}, and PaddleOCR-VL~\cite{cui2025paddleocrvl}. These efforts aim to enhance parsing accuracy through large-scale modeling. However, due to the limited robustness of current open-source models in handling complex layouts and lengthy text sequences, many still depend on a preliminary layout analysis module~\cite{sun2025pplayout,zhao2024doclayout} to detect document elements, with the VLM subsequently parsing content within localized regions. While this hybrid design improves usability to some degree, it has yet to fully exploit the potential of VLMs for end-to-end joint inference and unified multi-task modeling.

\begin{table}[!t]
    \small
    \centering
    \caption{Performance comparison of different VLMs and OCR systems across multiple tasks.
\includegraphics[width=8pt]{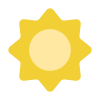} indicates Supported and High-Performing, \includegraphics[width=8pt]{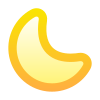} indicates Supported with Moderate Performance, and \includegraphics[width=8pt]{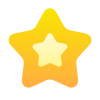} indicates Supported but Underperforming. Otherwise, it is Not Supported.}
    \vspace{-0.5em}
    \setlength{\abovecaptionskip}{0.2cm}
    \setlength{\tabcolsep}{0.55mm}{
        \begin{tabular}{l|ccc|ccccc}
            \toprule
            \multirow{2}{*}{\parbox{1.6cm}{\centering \textbf{Model\\Type}}} &
            \multirow{2}{*}{\parbox{1.9cm}{\centering \textbf{Inference\\Type}}} &
            \multirow{2}{*}{\parbox{1.6cm}{\centering \textbf{Model\\Name}}} &
            \multirow{2}{*}{\parbox{2.1cm}{\centering \textbf{Deploytment\\Cost}}} &
            \multicolumn{5}{c}{\textbf{Task}}   
            \\
            &&&&Spotting&Parsing&Text-VQA&IE&Translation
            \\
            \midrule
            \multirow{5}{*}{\parbox{1.6cm}{\centering \textbf{Casecade\\Pipeline}}}&
            \multirow{5}{*}{\parbox{1.6cm}{\centering \textbf{Multi-Step}}}&
            PaddleOCR-V5&low& \includegraphics[width=8pt]{imgs/moon.png} & - & - & - & -
            \\
            &&BaiduOCR&low& \includegraphics[width=8pt]{imgs/sun.png} & - & - & - & -
            \\
            &&Marker-1.8.2&low& - & \includegraphics[width=8pt]{imgs/star.png} & - & - & -
            \\
            &&PP-ChatOCR&medium& - & - & - & \includegraphics[width=8pt]{imgs/star.png} & -
            \\
            &&PP-DocTranslation&high& - & - & - & - & \includegraphics[width=8pt]{imgs/moon.png}
            \\
            \midrule
            \multirow{3}{*}{\parbox{1.6cm}{\centering \textbf{Specialized\\VLMs\\(Modular)}}}&
            \multirow{3}{*}{\parbox{1.6cm}{\centering \textbf{two-stage}}}&
            MonkeyOCR-pro-3B&medium& - & \includegraphics[width=8pt]{imgs/sun.png} & - & - & -
            \\
            &&MinerU2.5&low& - & \includegraphics[width=8pt]{imgs/sun.png} & - & - & -
            \\
            &&PaddleOCR-VL&low& - & \includegraphics[width=8pt]{imgs/sun.png} & - & - & -
            \\
            \midrule
            \multirow{3}{*}{\parbox{1.6cm}{\centering \textbf{General\\VLMs}}}&
            \multirow{3}{*}{\parbox{1.6cm}{\centering \textbf{One-Step}}}&
            Gemini-2.5-Pro&high& \includegraphics[width=8pt]{imgs/star.png} & \includegraphics[width=8pt]{imgs/moon.png} & \includegraphics[width=8pt]{imgs/sun.png} & \includegraphics[width=8pt]{imgs/sun.png} & \includegraphics[width=8pt]{imgs/sun.png}
            \\
            &&Seed-1.6-Vision&high& \includegraphics[width=8pt]{imgs/moon.png} & \includegraphics[width=8pt]{imgs/moon.png} & \includegraphics[width=8pt]{imgs/sun.png} & \includegraphics[width=8pt]{imgs/star.png} & \includegraphics[width=8pt]{imgs/sun.png}
            \\
            &&Qwen3-VL-235B-Instruct&high& \includegraphics[width=8pt]{imgs/moon.png} & \includegraphics[width=8pt]{imgs/moon.png} & \includegraphics[width=8pt]{imgs/sun.png} & \includegraphics[width=8pt]{imgs/moon.png} & \includegraphics[width=8pt]{imgs/sun.png}
            \\
            \midrule
            \multirow{4}{*}{\parbox{1.6cm}{\centering \textbf{Specialized\\VLMs\\(End2End)}}}&
            \multirow{4}{*}{\parbox{1.6cm}{\centering \textbf{One-Step}}}&
            Mistral-OCR&medium& - & \includegraphics[width=8pt]{imgs/moon.png} & - & - & -
            \\
            &&Deepseek-OCR &medium& - & \includegraphics[width=8pt]{imgs/moon.png} & \includegraphics[width=8pt]{imgs/star.png} & \includegraphics[width=8pt]{imgs/star.png} & -
            \\
            &&dots.ocr&medium& - & \includegraphics[width=8pt]{imgs/sun.png} & - & - & -
            \\
            &&HunyuanOCR&low& \includegraphics[width=8pt]{imgs/sun.png} & \includegraphics[width=8pt]{imgs/sun.png} & \includegraphics[width=8pt]{imgs/sun.png} & \includegraphics[width=8pt]{imgs/sun.png} & \includegraphics[width=8pt]{imgs/sun.png}
            \\
            
            \bottomrule
        \end{tabular}}
    \label{intro}
\end{table}

This report introduces HunyuanOCR, a novel open-source multilingual VLM designed for OCR that delivers commercial-grade performance. Departing from conventional pipeline-based frameworks, HunyuanOCR adopts an end-to-end VLM architecture, establishing a unified foundation for multi-task learning that effectively overcomes long-standing challenges such as error propagation and high maintenance costs. As summarized in Table~\ref{intro}, our model demonstrates significant advantages across four key dimensions:
1) Comprehensive Capability Coverage: HunyuanOCR supports an extensive range of tasks beyond basic document parsing, including text spotting, end-to-end receipt information extraction, video subtitle recognition, text-centric visual question answering (VQA), as well as multilingual recognition and translation. By integrating these diverse capabilities into a unified modeling framework, it addresses complex and varied application needs, establishing itself as one of the most comprehensive OCR expert models in the open-source community. 
2) High Inference Efficiency: Built upon the native Hunyuan VLM architecture, the model contains only 1B parameters while maintaining high computational efficiency. This compact design ensures low latency and makes it suitable for on-device deployment, meeting the practical requirements of resource-constrained environments.
3) Superior Performance: HunyuanOCR outperforms leading open-source alternatives on core benchmarks; for instance, it surpasses MinerU2.5 and PaddleOCR-VL on the OmniDocBench for document parsing. It also excels in specialized tasks—exceeding Qwen3-VL-4B~\cite{bai2025qwen25vl} in text image translation and information extraction, and outperforming PaddleOCR 3.0 and certain commercial Cloud OCR APIs in text spotting tasks. 
4) Enhanced Usability and Unified Modeling: The end-to-end VLM architecture enables unified task modeling within a single framework, allowing diverse OCR tasks to be accomplished through a single inference based on natural language instructions. This design eliminates the need for complex model cascading and post-processing, significantly lowering the technical barrier and offering a streamlined, user-friendly solution for diverse application scenarios.

The HunyuanOCR model adopts an efficient, compact architecture that connects a 0.4B-parameter native-resolution Vision Transformer (ViT)~\cite{tschannen2025siglip} to a 0.5B-parameter Hunyuan Large Language Model (LLM)~\cite{hunyuan2025llm} via a learnable pooling MLP adapter. The model is trained following the mainstream two-stage paradigm for VLMs. The first stage, pre-training, involves four steps: vision-language alignment, multi-modal pre-training, long-context pre-training, and application-oriented SFT.
This stage utilizes a mixture of large-scale open-source data, synthetic element-level data, and high-quality, end-to-end application-oriented data (e.g., complex long-document parsing and text image translation), totaling approximately 200 million high-quality samples. The second stage, post-training, employs the online reinforcement learning algorithm GRPO with task-specific reward mechanisms, significantly improving the model’s accuracy and stability in challenging scenarios such as complex document parsing and text image translation.

This study demonstrates the substantial potential of the end-to-end VLM paradigm when applied to OCR-specific tasks. We attribute the success of HunyuanOCR to two principal insights. First, during pre-training, exposing the model to high-quality, application-aligned data proves critical for performance—especially in complex and long-text document parsing, as well as in text image translation tasks. Second, the design of targeted online reinforcement learning strategies, combined with an emphasis on data diversity and quality, leads to significant gains in OCR-specific VLMs. These improvements are most pronounced in challenging settings such as intricate layout understanding and knowledge-intensive tasks including visual question answering and image-based translation.

%% file: sec/relatedwork.tex
\section{Related Work}
The evolution of Optical Character Recognition (OCR) technology, traceable to the 1950s, has exhibited distinct developmental phases. In the initial stage (1950s–1980s), OCR systems were primarily based on template matching and feature engineering, focusing on basic text recognition in scanned documents. The 1990s witnessed a significant breakthrough with the maturation of machine learning theory, as statistical methods such as Hidden Markov Models (HMMs)~\cite{eddy1996hmm} and Support Vector Machines (SVMs)~\cite{cortes1995svm} were widely adopted, substantially improving recognition accuracy. Entering the 21st century, rapid advances in deep learning catalyzed a paradigm shift in OCR: system architectures have progressively transitioned from traditional modular frameworks to the current paradigm of unified processing enabled by vision-language models.

\subsection{Traditional OCR Systems}
Traditional OCR systems typically employ a highly modularized pipeline architecture. Depending on the requirements of specific application scenarios, such systems often incorporate several core processing modules with distinct functionalities, primarily including, but not limited to: deep learning-based text detection, text recognition, document layout analysis, named entity recognition, and optional text translation modules. Over the past few decades, significant research efforts have been devoted to this direction. Through continuous innovation, numerous models have been developed~\cite{zhou2017east,liao2017textboxes,liao2022real,bgshi2017crnn,shi2018aster,lyu2018mask,li2023trocr,lyu2024maskocr,li2021structext,yustructextv2}, substantially enhancing the accuracy and robustness of each functional module.

Nevertheless, conventional OCR systems still suffer from two fundamental limitations that require urgent resolution. First, at the architectural level, these solutions generally rely on cascading multiple independent functional modules, resulting in highly complex system structures. Taking a typical document parsing task as an example, a fully functional system typically requires integrating at least five key subsystems: a high-precision text detection module, a multilingual text recognition engine, a fine-grained layout analysis component, a specialized mathematical formula recognition module, and a structured table recognition unit. This modular stacking design not only increases deployment complexity and maintenance costs but also requires specialized personnel to perform coordinated tuning of each component. Second, during inference, the multi-stage cascaded processing flow leads to progressive error amplification through a ``pipeline effect." Specifically, inaccuracies in text detection can degrade input quality for subsequent recognition modules, while layout analysis errors may cause incorrect ordering of text blocks. These early-stage inaccuracies ultimately compromise the accuracy and usability of the system's final output. Consequently, traditional OCR systems often fail to meet practical requirements when handling complex scenarios such as documents with overlapping text or non-standard layouts.

\subsection{Vision-Language Models}
With the rapid advancement of deep learning, large language models (LLMs)~\cite{devlin2019bert,radford2019language,brown2020language,liu2024deepseek,qwen3,comanici2025gemini} have achieved remarkable breakthroughs in natural language processing (NLP). Subsequently, VLMs~\cite{liu2023visual,achiam2023gpt,bai2025qwen25vl,comanici2025gemini,wang2025internvl3}, which align information across multiple modalities, have demonstrated exceptional capabilities in cross-modal understanding and generation. These models typically employ unified neural network architectures, enabling efficient handling of complex cognitive tasks such as visual recognition, textual comprehension, and multimodal reasoning.
The advantages of this paradigm are twofold. First, architecturally, the unified network design supports synergistic multi-task processing, allowing a single model to perform diverse tasks in an end-to-end manner. Second, by leveraging the inherent reasoning abilities of LLMs, this architecture achieves substantial performance gains, particularly in cognition-intensive applications.

\subsubsection{General Vision-Language Models}
Current mainstream general vision-language models, such as Gemini~\cite{comanici2025gemini} and Qwen-VL~\cite{bai2025qwen25vl}, have demonstrated strong OCR capabilities. These models exhibit robust text perception, accurately recognizing both printed and handwritten text while effectively handling complex scenarios involving irregular layouts, low-resolution images, and multilingual content. However, their large parameter size introduces two notable limitations in practical applications. First, inference requires substantial GPU memory and computational resources. Second, they often fail to meet the stringent low-latency requirements of real-world business scenarios.

\subsubsection{OCR-Specific Vision-Language Models}
To address the aforementioned technical constraints, the development of lightweight, specialized vision-language models for OCR has emerged as a promising solution. Pioneering approaches such as Nougat~\cite{blecher2023nougat} and StructText-V3~\cite{lyu2024structextv3} attempted to achieve end-to-end processing for document parsing and information extraction within a unified model. Subsequent models including Dolphin~\cite{feng2025dolphin}, MonkeyOCR~\cite{li2025monkeyocr}, Dots.OCR~\cite{dot2024ocr}, MinerU2.5~\cite{niu2025mineru25}, and PaddleOCR-VL~\cite{cui2025paddleocrvl} have drawn inspiration from traditional OCR pipelines. These methods typically first perform layout detection~\cite{zhao2024doclayout,sun2025pplayout} using a dedicated model or a repurposed vision-language model, followed by unified recognition of text blocks, formulas, and tables. While these approaches reduce system complexity and improve accuracy compared to traditional pipelines by leveraging the generalization capability of vision-language models, they remain susceptible to error propagation from the layout analysis stage and fail to fully exploit the benefits of end-to-end optimization.

In contrast, the proposed HunyuanOCR model demonstrates substantial advantages in both technical architecture and application effectiveness across three key dimensions:

1) Fully end-to-end architecture: HunyuanOCR employs a purely end-to-end design that eliminates error accumulation from cascaded processing. This architecture maximizes the potential of end-to-end learning through a systematically optimized training paradigm. From an engineering perspective, the model completes entire workflows in a single inference pass, significantly improving operational efficiency in real-world applications.

2) Comprehensive functional coverage: Leveraging the unified task-handling capability of vision-language models, HunyuanOCR supports not only basic document parsing and text spotting but also advanced functionalities, including information extraction, visual question answering, and cross-lingual translation. Notably, it provides extensive multilingual support for hundreds of global languages, making it one of the most functionally complete specialized OCR solutions available.

3) Superior performance benchmarking: HunyuanOCR achieves exceptional performance, with key metrics significantly surpassing current state-of-the-art models and matching or exceeding the standards of leading commercial OCR APIs.

%% file: sec/model_design.tex
\section{Model Design}

\begin{figure}[!ht]
    \centering
    \includegraphics[width=1\linewidth]{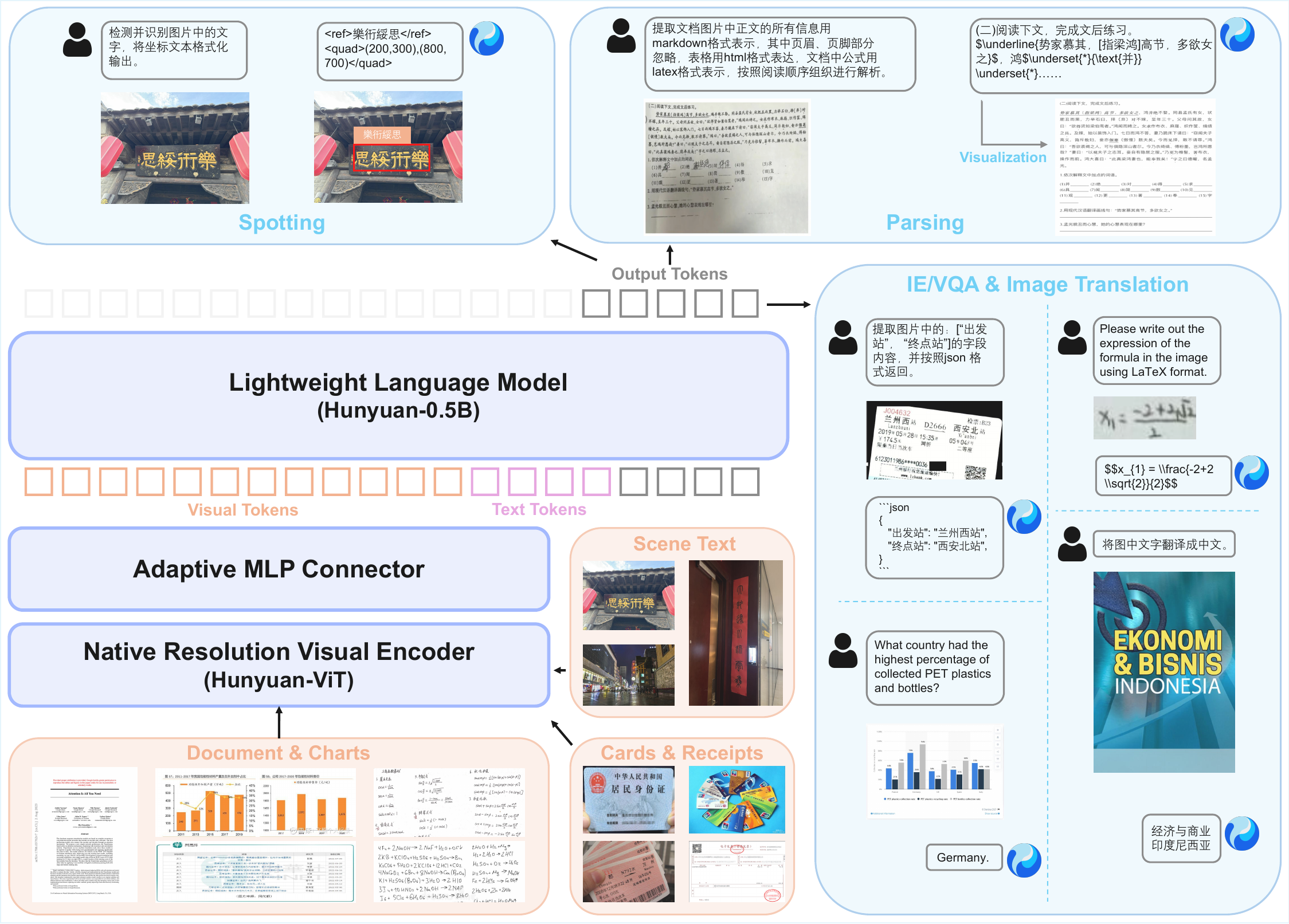}
    \caption{The Architecture of HunyuanOCR: An end-to-end framework integrating Native Resolution Visual Encoder, Adaptive MLP Connector, and a Lightweight Language Model for diverse OCR tasks, including: spotting, parsing, information extraction, visual question answering, and text image translation.}
    \label{fig:framework}
\end{figure}

HunyuanOCR features a collaborative architecture comprising three core modules: a Native Resolution Visual Encoder, an Adaptive MLP Connector, and a Lightweight Language Model.

\textbf{Native Resolution Visual Encoder (Hunyuan-ViT)} is built upon the SigLIP-v2-400M pre-trained model~\cite{tschannen2025siglip}. By incorporating a hybrid generative-discriminative joint training strategy, it significantly enhances the model's ability to comprehend complex visual semantics. The encoder natively supports arbitrary input resolutions through an adaptive patching mechanism that preserves the original aspect ratio, making it particularly suitable for challenging scenarios involving extreme aspect ratios such as long-text documents. The image is divided into patches according to its native proportions, and all patches are processed by the Vision Transformer (ViT) with global attention. This design avoids image distortion and detail loss, leading to notable improvements in text recognition accuracy for difficult cases, including long text lines, extensive documents, and low-quality scans.

\textbf{Adaptive MLP Connector} acts as a bridge between the visual and linguistic domains, implementing a core learnable pooling operation. It employs spatial-dimension adaptive content compression to reduce the sequence length of tokens generated from the visual encoder's high-resolution feature maps, effectively minimizing redundancy. During this process, the module preserves critical semantic information from key areas, such as text-dense regions, thereby achieving an efficient and precise projection of visual features into the input space of the language model.

\textbf{Lightweight Language Model} is based on the densely architected Hunyuan-0.5B model~\cite{hunyuan2025llm}. It incorporates XD-RoPE, which deconstructs the conventional RoPE~\cite{su2024roformer} into four independent subspaces: text, height, width, and time. This design establishes a native alignment mechanism that bridges 1D text sequences, 2D page layouts, and 3D spatiotemporal information, enabling the model to handle both complex layout parsing (e.g., multi-column recognition) and cross-page document analysis with logical reasoning.

\textbf{End-to-End Optimization.} In contrast to other specialized vision-language OCR models, HunyuanOCR employs a fully end-to-end paradigm for both training and inference. By scaling high-quality, application-oriented data and leveraging reinforcement learning optimization, the system eliminates the need for post-processing and the associated error accumulation typical of pipeline-based architectures. It demonstrates superior robustness in challenging scenarios such as mixed-layout document understanding.

%% file: sec/data_construction.tex
\section{Data Construction}
\subsection{Task Design}
Capitalizing on the architectural advantages of vision-language models, HunyuanOCR integrates various OCR tasks into a single, unified paradigm. This enables one model to address multiple high-frequency tasks across the OCR domain.

\subsubsection{Spotting}
As a fundamental OCR capability, text spotting requires precise localization and recognition of text within images. HunyuanOCR adopts a standardized instruction template for this task, using the fixed prompt: ``Detect and recognize text in the image, and output the text coordinates in a formatted manner.” This instruction guides the model to output both line-level text content and corresponding coordinate information. To ensure machine-parsable responses, a structured output format is defined: \texttt{<ref>text</ref><quad>(x1,y1),(x2,y2)</quad>}. Here, text inside the \verb|<ref>| and \verb|</ref>| tags denotes the recognized content, and the coordinate sequence within \verb|<quad>| and \verb|</quad>| tags specifies the bounding box of the text region using its top-left and bottom-right vertices. All coordinates are normalized to the range [0, 1000] to maintain consistency across input images of varying resolutions.


\subsubsection{Parsing}
Document parsing constitutes a core OCR capability, whose strategic importance has been heightened by the rapid advancement of large language models (LLMs). It serves not only as a key preprocessing tool for building high-quality training datasets but also as an essential upstream component in retrieval-augmented generation (RAG) systems.

HunyuanOCR provides a comprehensive document parsing solution that supports both fine-grained element-level parsing and full end-to-end document parsing.

\textbf{Fine-Grained Element Parsing}: It supports the independent identification and extraction of specialized document elements, including mathematical formulas, chemical formulas, tables, and charts. HunyuanOCR employs standardized instruction templates to guide the parsing of different document elements:

\begin{itemize}[leftmargin=*, itemsep=0pt, topsep=0pt]
    \item  Formula Parsing: Using the prompt ``Identify the formula in the image and represent it using LaTeX format."
          , the model returns the corresponding LaTeX codes of mathematical or chemical formulas.

    \item Table Parsing: Using the prompt ``Parse the table in the image into HTML."
          , the model returns the HTML codes of the tables.

    \item Chart Parsing: Using the prompt ``Parse the chart in the image, use Mermaid format for flowcharts and Markdown for other charts."
          , the model adaptively describes the chart using either Mermaid syntax or Markdown based on its type.
\end{itemize}

\textbf{End-to-End Document Parsing}: HunyuanOCR enables integrated, full-page parsing of documents containing multiple and complex element types. We use the prompt: ``Extract all information from the main body of the document image and represent it in markdown format, ignoring headers and footers. Tables should be expressed in HTML format, formulas in the document should be represented using LaTeX format, and the parsing should be organized according to the reading order.”
This instruction guides the model to perform an integrated analysis of the document image, outputting all textual content in its natural reading sequence while intelligently converting any identified tables and formulas into HTML and LaTeX formats, respectively, and outputting the spatial positions of figures or charts in the image with corresponding titles. Additionally, we introduce a generalized prompt called “Extract the text in the image”. It is designed for diverse real-world scenarios and guides the model to read any image, such as posters, street views, product packaging, or UI screens, in natural reading order. Detected tables are converted into Markdown format and formulas into LaTeX, producing clean and structured output for broad downstream use.

\subsubsection{IE \& VQA}

HunyuanOCR delivers comprehensive document understanding through robust IE and advanced VQA capabilities.

\textbf{IE}: As a core OCR function, IE precise perceptual localization and deep semantic association. HunyuanOCR provides robust structured extraction with strengths in two primary dimensions:

\begin{itemize}[leftmargin=*, itemsep=0pt, topsep=0pt]

    \item \textbf{Domain Adaptability}: HunyuanOCR is designed for open-world extraction of arbitrary fields, exhibiting strong domain adaptability while being precisely optimized for over 30 common document types. These include 30 types of cards and receipts, the detailed categories are listed in Table~\ref{tab:ie_common_categories}.

    \item \textbf{Instruction-Driven Control}: HunyuanOCR allows granular control via natural language instructions. It supports both targeted single-field extraction (e.g., ``Please output the value of $<\mathrm{Key}>$'') and parallel multi-field extraction into structured JSON based on user-provided key lists (e.g.,``Extract ['key1','key2',\dots] and return in JSON format''), enabling seamless adaptation to diverse application scenarios.

    \item \textbf{Video Subtitle Extraction}: In response to the instruction ``Extract the subtitles from the image," HunyuanOCR performs subtitle extraction from standard video screenshots, enabling robust handling of text across diverse resolutions, aspect ratios (landscape/portrait), and on-screen positions in both horizontal and vertical orientations.

\end{itemize}

\textbf{VQA}: HunyuanOCR demonstrates strong open-domain document QA performance, effectively processing open-ended queries about imaged text and generating accurate predictions. Its key capabilities include:

\begin{itemize}[leftmargin=*, itemsep=0pt, topsep=0pt]

    \item \textbf{Multi-Format Input Support}: The model processes diverse inputs including cropped text lines, mathematical formulas, documents, charts, and street-view imagery for perception and understanding.

    \item \textbf{Advanced Reasoning}: Beyond basic recognition, it performs complex tasks such as spatial and attribute understanding, logical reasoning, and numerical computation based on visual and textual content.
\end{itemize}

\subsubsection{Text Image Translation}

HunyuanOCR incorporates a comprehensive end-to-end image-to-text translation module that supports over 14 source languages—including French, German, Japanese, Korean, and many other widely used or regionally important languages—translating them into either Chinese or English. In addition, the system enables direct bidirectional translation between Chinese and English, covering both general-purpose translation scenarios and document-centric translation tasks with complex layouts.

Beyond language coverage, HunyuanOCR is designed for multi-scenario robustness, handling both document-oriented inputs—such as scanned pages, structured layouts, tables, forms, and dense paragraphs—and general scenes containing natural images with embedded text, signage, posters, captions, and other visually diverse contexts. This allows the model to perform reliable translation under variations in layout complexity, image quality, lighting, distortion, and multilingual content distribution.

To fully activate the model’s translation capabilities across different use cases, we design two complementary prompting paradigms:

\begin{itemize}[leftmargin=*, itemsep=0pt, topsep=0pt]

    \item \textbf{General-purpose Translation Prompt}
          “Extract all text from the image and translate it into Chinese/English.”
          This prompt targets general scene-text translation without assuming any document structure.

    \item \textbf{Document-oriented Translation Prompt}
          “First parse the document, then translate its content into Chinese. Ignore headers and footers; represent equations in \LaTeX{}; and render tables in HTML format.”
          This prompt is tailored for English-to-Chinese document image translation requiring structured parsing.
\end{itemize}



\subsection{Data Pipelines}
\begin{figure}[t]
    \centering
    \includegraphics[width=0.95\linewidth]{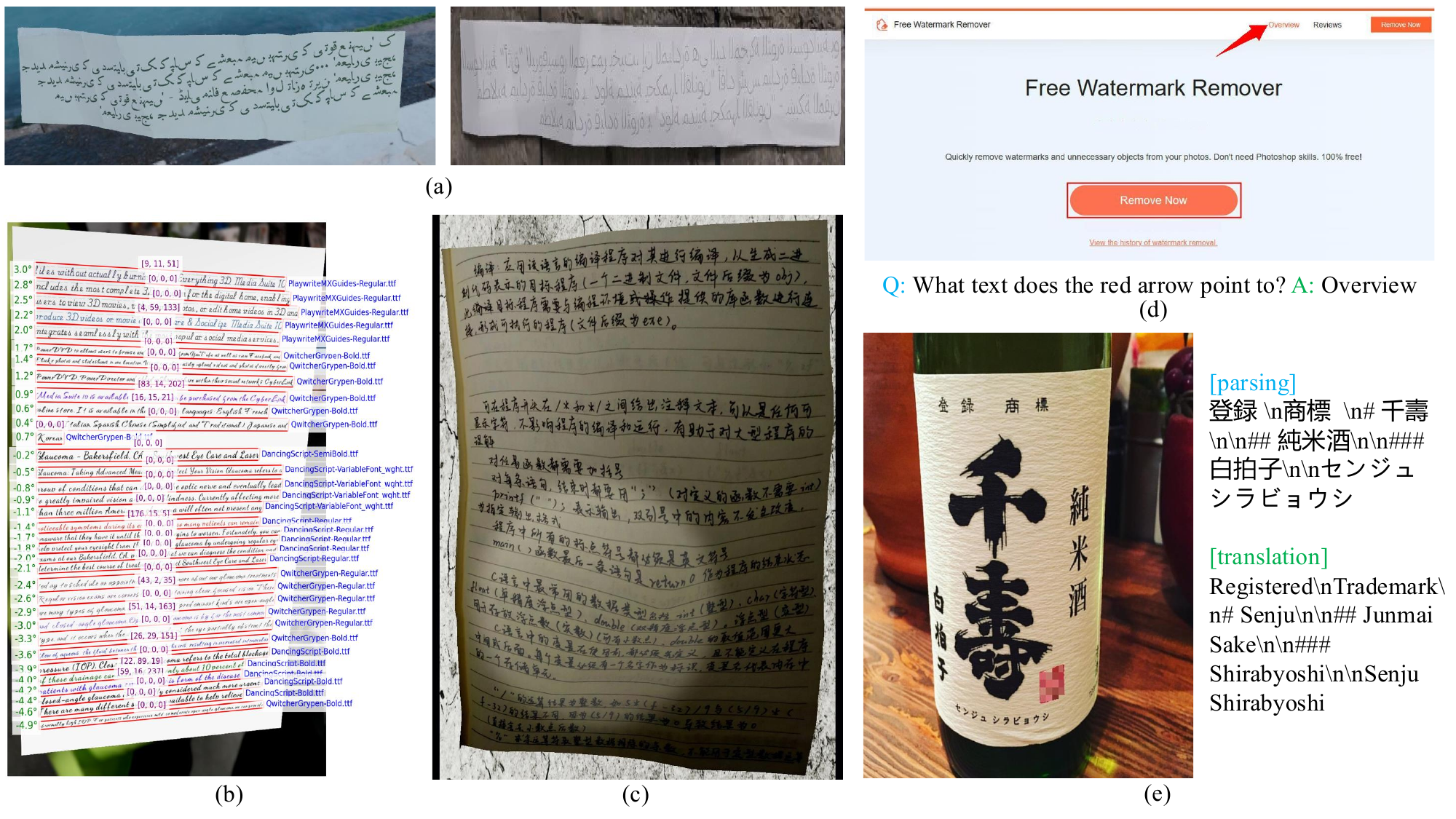}
    \vspace{-2mm}
    \caption{Illustration of image data synthesis and data augmentation results for the HunyuanOCR data pipeline.
        (a) Multilingual synthetic data with right-to-left (RTL) reading order.
        (b) Long-document, paragraph-level synthesis with controllable line-level font, language, rotation, and RGB values.
        (c) Document image warping with realistic defects, including perspective distortion, blur, and local lighting variation.
        (d) Cross-task data reuse: from spotting data to automated QA generation.
        (e) Cross-task data reuse: from multilingual parsing data to real-world text translation.}
    \label{fig:hunyuanocr_pipeline}
\end{figure}

To systematically enhance HunyuanOCR's perceptual and comprehension capabilities across diverse scenarios, languages, and layouts, we constructed large-scale, high-quality training data aligned with the core tasks described above. Beyond aggregating public benchmarks, we collected extensive real-world data through web crawling and generated high-quality synthetic samples using proprietary synthesis tools. Via a complete data production and cleaning pipeline (Fig.~\ref{fig:hunyuanocr_pipeline}), we built a corpus of over 200 million image-text pairs spanning nine major real-world scenarios—street views, documents, advertisements, handwritten text, screenshots, cards/certificates/invoices, game interfaces, video frames, and artistic typography—and covering more than 130 languages, forming a high-quality multimodal training resource.

\subsubsection{Image data synthesis}

Building upon the SynthDog framework, we have extended its capabilities to generate high-quality synthetic data for long-document parsing and translation tasks. The system supports paragraph-level rendering in over 130 languages and comprehensively handles bidirectional text layouts (LTR/RTL) as well as complex cursive scripts (Fig.~\ref{fig:hunyuanocr_pipeline}(a)–(b)).

The proposed synthesis pipeline exhibits the following core characteristics. First, it enables fine-grained control over text attributes—such as font, color, and orientation—as well as image perturbations, including lighting and shadows, during the rendering process. Second, it accurately simulates complex typographical features, such as handwritten-style fonts and mixed-font typesetting. Furthermore, the system significantly enhances support for low-resource languages, effectively improving cross-lingual generalization in OCR and machine translation. Finally, through a unified architecture, it generates image-text aligned data suitable for a variety of tasks, including spotting, long-document parsing, and cross-lingual translation.

\subsubsection{Image data augmentation}

We employ an in-house Warping Synthesis Pipeline to simulate realistic imaging defects in photographed and natural-scene documents, thereby enhancing model robustness (Fig.~\ref{fig:hunyuanocr_pipeline}(c)). The pipeline incorporates three key functions: geometric deformation via control-point manipulation to emulate folds, curves, and perspective distortions; imaging degradation with motion blur, Gaussian noise, and compression artifacts; and illumination perturbations that model global/local lighting variations, shadows, and reflections. This pipeline substantially improves the robustness of core OCR tasks, such as text spotting, document parsing, and visual question answering.

\subsubsection{Question–Answer Pair Generation}


We have developed an automated pipeline that integrates Hard Sample Retrieval, QA Generation, and Consistency Verification to produce high-quality VQA data while maximizing cross-task sample reuse. Based on a ``single source, multiple uses" principle, the pipeline jointly manages spotting, parsing outputs, and VQA annotations for each image, enabling unified training across text spotting, document parsing, and text-centric VQA tasks.

\textbf{Hard Sample Retrieval}. We employ an automated image and label-based filtering strategy to identify challenging cases from large-scale datasets. Priority is given to samples with low clarity, complex tables or formulas, code snippets, and low-resource language text. This approach ensures that extensive training effectively enhances model performance on these challenging scenarios.

\textbf{Instructional QA Generation.} We designed unified instruction templates to automatically generate question-answer (QA) pairs for multiple types of tasks using a high-performance visual language model (VLM). For instance, the system can produce parsing tasks encompassing the recognition and conversion of elements such as code snippets, formulas, tables, and charts into structured formats, including Markdown, HTML, and JSON. Furthermore, by leveraging textual content, chart attributes, semantic information, and numerical data present in the image, the method generates diverse QA pairs covering information extraction, numerical computation, content summarization, and other reasoning tasks.


\textbf{Consistency Verification and Data Refinement.} We employ a multi-model cross-validation mechanism to evaluate the confidence of generated question-answer (QA) pairs. Data that pass the validation are directly incorporated into the training set to ensure quality, while a subset of the failing cases undergoes manual verification to supplement challenging samples that are difficult for the models to process, thereby enhancing the diversity and coverage of the dataset.



%% file: sec/training_recipe.tex
\section{Training Recipe}

\subsection{Pre-Training}

We employ a four-stage training strategy for HunyuanOCR pre-training, as outlined in Table~\ref{tab:hyocr_pretrain}. The process begins with Stage 1, which warms up the vision–language bridge. In Stage 2, all model parameters are unlocked for end-to-end multimodal learning. Stage 3 extends the context window to 32k tokens to support long-document parsing and understanding. Finally, Stage 4 conducts application-oriented tuning using standardized instructions and normalized outputs, establishing a solid foundation for subsequent reinforcement learning.

\begin{itemize}[leftmargin=*, itemsep=0pt, topsep=0pt]
\item ~\textbf{Stage-1:} In the first stage, we train only the visual encoder (ViT) and a learnable MLP adapter while keeping the language model frozen, aligning visual features with the textual semantic space. The training corpus consists primarily of general image captioning data and synthetic OCR data focused on parsing and recognition tasks, supplemented with a small proportion of plain text (\(\leq 10\%\)) to preserve the core linguistic capabilities of the language model. This stage emphasizes text parsing and recognition to enhance the model's perception and structured understanding of textual content in images. Training uses approximately 50B tokens, with the learning rate warmed up from \(3\times10^{-4}\) to a peak value before decaying to \(3\times10^{-5}\).


\item ~\textbf{Stage-2:} In the second stage, all model parameters are unfrozen for end-to-end vision-language joint learning, with a focus on enhancing the model's capability for deep understanding and cognitive reasoning of structured content such as documents, tables, and charts. The training data mixture increases the proportion of synthetic samples covering multiple tasks, including text parsing, spotting, translation, and VQA, while retaining approximately (\(\leq 10\%\)) plain text to maintain instruction-following and linguistic generalization capabilities. The training utilizes approximately 300B tokens with a warmup-cosine learning rate schedule, decaying from \(2\times10^{-4}\) to \(5\times10^{-5}\).


\item ~\textbf{Stage-3:} In the third stage, we extend the model's context window to 32K by incorporating long-context parsing tasks and lengthy plain text data. This stage uses approximately 80B tokens, decaying the learning rate from \(8\times10^{-5}\) to \(5\times10^{-6}\).


\item ~\textbf{Stage-4:} We conduct annealing training using carefully curated, human-annotated real-world data supplemented with a small proportion of high-quality synthetic samples, while maintaining a 32K context window to enhance perceptual robustness in complex scenarios. By employing unified instruction templates and standardized output formats across different tasks, we ensure consistency in response patterns throughout the training data. This design not only reduces the model's learning difficulty but also facilitates the design of reward models in subsequent post-training stages. The training utilizes 24B tokens in this stage, with the learning rate linearly decaying from \(2\times10^{-5}\) to \(1\times10^{-6}\).

\end{itemize}

\begin{table}[t]
    \centering
    \caption{Overview of the four-stage pre-training recipe for HunyuanOCR pre-training.}
    \vspace{-2mm}
    \label{tab:hyocr_pretrain}
    \renewcommand{\arraystretch}{1.5}
    \setlength{\tabcolsep}{3.0pt}
    \scriptsize
    \begin{tabular}{p{0.13 \linewidth} | p{0.19\linewidth} p{0.19\linewidth} p{0.2\linewidth} p{0.2\linewidth}}
        \toprule
        \textbf{Stages}           & \textbf{Stage-1}                                                              & \textbf{Stage-2}                                                 & \textbf{Stage-3}                                                                                        & \textbf{Stage-4}                                                         \\
        \midrule
        \textbf{Purpose}          & Vision-Language Alignment                                                     & Multimodal Pre-traning                                           & Long-context Pre-training                                                                               & Application-oriented SFT                                                 \\
        \textbf{Trainable Parts}  & ViT \& Adapter                                                                & All                                                              & All                                                                                                     & All                                                                      \\
        \midrule
        \textbf{Learning Rate}    & $3\mathrm{e}{-4} \rightarrow 3\mathrm{e}{-5}$                                 & $2\mathrm{e}{-4} \rightarrow 5\mathrm{e}{-5}$                    & $8\mathrm{e}{-5} \rightarrow 5\mathrm{e}{-6}$                                                           & $2\mathrm{e}{-5} \rightarrow 1\mathrm{e}{-6}$                            \\
        \textbf{Training Tokens}  & 50B                                                                           & 300B                                                             & 80B                                                                                                     & 24B                                                                      \\
        \textbf{Sequence Length}  & 8k                                                                            & 8k                                                               & 32k                                                                                                     & 32k                                                                      \\
        \textbf{Data Composition} & Pure Text, Synthetic Parsing and Recognition Data, General Image Caption Data & Pure Text, Synthetic Spotting, Parsing, Translation and VQA Data & Long Pure Text, Real-world Auto-annotated Data, Long Document Parsing Data, Information Extraction Data & Human-annotated Data, Hard-negative Data, Standardized Instruction Data. \\
        \bottomrule
    \end{tabular}
    \normalsize
\end{table}

\subsection{Reinforcement Learning (RL)}
\label{sec:rl}
Reinforcement learning (RL) algorithms have emerged as a powerful paradigm, achieving remarkable success across various domains involving large language models (LLMs) and multimodal large language models (MLLMs). Notable applications include mathematical reasoning~\cite{grpo}, image segmentation~\cite{seg_zero}, and omni-multimodal LLMs~\cite{r1_omni}. This broad success is largely attributed to RL's ability to align model outputs with verifiable metrics~\cite{wen2025reinforcement} or human preferences~\cite{omni_dpo, sentinel}.

While RL has traditionally been applied to large-scale reasoning models, we investigate its application to lightweight OCR models that prioritize efficient and accurate text understanding. Leveraging the structured nature and inherent verifiability of many OCR tasks, we adopt Reinforcement Learning with Verifiable Rewards (RLVR) for closed-form tasks such as text spotting and document parsing. For more open-ended tasks like translation and text-centric VQA, we design reward mechanisms based on an LLM-as-a-judge approach. By integrating RLVR and LLM-as-a-judge techniques, we demonstrate that even lightweight models can achieve significant performance improvements, opening new possibilities for edge and mobile applications.

\subsubsection{Data Curation}

Our data pipeline emphasizes \textbf{quality, diversity, and difficulty balance}. In terms of quality, we combine high-quality open-source and synthetic datasets, and filter them using LLM-based judging to ensure image–text alignment and the removal of tasks that are easily exploitable (e.g., multiple-choice). For diversity, we cover a broad range of OCR-related tasks mentioned above, and maintain sufficient exploration by discarding samples with low output diversity or zero reward variance. Finally, to balance task difficulty, we employ pass-rate filtering based on model samples, removing both trivial and unsolvable examples.

\subsubsection{Reward Design}

We adopt a \textbf{ability-adaptive reward design}, where each OCR-related task type has a tailored reward formulation that aligns with its output characteristics.

\begin{itemize}[leftmargin=*, itemsep=0pt, topsep=0pt]
    \item \textbf{Spotting:} For text spotting tasks, which require joint text recognition and bounding box localization, the reward is computed as follows. Each predicted bounding box is first assigned to a ground-truth box by maximizing the Intersection over Union (IoU). The reward for each matched pair is then calculated as one minus the normalized edit distance between the predicted and ground-truth text strings. Any unmatched predictions or ground-truth boxes incur a penalty by contributing a reward of zero to the average. The final reward is the mean score across all evaluated pairs, providing a balanced measure of both localization and recognition accuracy.
    \item \textbf{Document Parsing:} Document Parsing aims to convert document images into structured formats containing textual content, mathematical formulas, and tables. The evaluation emphasizes both structural integrity and content accuracy. The reward is computed based on the normalized edit distance between the model's output and the ground-truth reference.
    \item \textbf{VQA:} The reward is binary (1 or 0), based on whether the model’s answer semantically matches the reference. The scoring model evaluates only content completeness and factual correctness, tolerating minor stylistic differences while enforcing strict alignment on key content elements.
    \item \textbf{Translation:} We use a soft reward scheme where a scoring LLM compares the generated output against the reference and assigns a score in the range [0, 5]. This raw score is then debias-normalized to [0,~1]. Crucially, this normalization is designed to expand the reward granularity in the mid-range (2–4), enabling the model to better capture subtle improvements and differences in translation quality.
\end{itemize}

\subsubsection{Training Strategy}

We adopt the Group Relative Policy Optimization (GRPO) algorithm as our main reinforcement learning framework. In each training iteration, GRPO samples a group of responses (${o_1, o_2, \cdots, o_G}$) for a given query (q) from the old policy ($\pi_{\theta_{\text{old}}}$) and updates the current policy ($\pi_{\theta}$) by maximizing the objective:

\vspace{-5mm}
\begin{equation}
    \begin{split}
        \mathcal{L}_{GRPO}(\theta) =  & \mathbb{E}_{[q \sim \mathcal{D}, \{o_i\}_{i=1}^G \sim \pi_{\theta_{\text{old}}}(\cdot|q)]} \\
        & \frac{1}{G}\sum_{i=1}^G \left[\min\left(\frac{\pi_\theta(o_i |q)}{\pi_{\theta_{\text{old}}}(o_i |q)} A_i, \text{clip} \left( \frac{\pi_\theta(o_i |q)}{\pi_{\theta_{\text{old}}}(o_i |q)}, 1 - \epsilon, 1 + \epsilon \right) A_i \right) - \beta \mathbb{D}_{KL}\left(\pi_{\theta} || \pi_{\text{ref}}\right)\right]
    \end{split}
    \label{eq:GRPO}
\end{equation}
\vspace{-3mm}

where $A_i$ represents the advantage computed from the group rewards, and $\mathbb{D}_{\text{KL}}$ is the KL-divergence term for regularization. The $\epsilon$ and $\beta$ control clipping and the strength of KL penalties, respectively.

To ensure stable and reliable training, we enforce length constraints and a strict format during reward computation. Specifically, any output that exceeds the maximum length is immediately assigned a reward of zero. Similarly, for structured tasks like spotting and document parsing, outputs that fail to follow the required schema are also directly penalized with zero reward. These constraints help the optimization process focus exclusively on valid, well-structured, and verifiable outputs, thereby guiding the model to learn accurate reasoning and formatting behavior under constrained conditions.

%% file: sec/evaluation.tex
\section{Evaluation}

\subsection{Spotting}
To comprehensively evaluate the model's text spotting performance across diverse scenarios, we constructed a benchmark comprising nine categories: artistic text, document images, game screenshots, handwritten text, advertisement scenes, card/certificate/invoice images, screen captures, street view text, and video frames. Each category contains 100 images, forming a 900-image evaluation set. Based on this benchmark, we compared HunyuanOCR with traditional pipeline-based open-source models, leading commercial APIs, and general Vision-Language Models (VLMs). The results shown in~\Cref{tab:spotting_performance} demonstrate that our approach achieves the best overall performance.

Specifically, as an end-to-end VLM solution, HunyuanOCR significantly outperforms traditional pipeline-based methods. Furthermore, compared to general VLMs, our method achieves superior accuracy with substantially fewer parameters, demonstrating notable advantages in both computational efficiency and performance.

\begin{table}[h!]
    \centering
    \footnotesize
    \setlength{\tabcolsep}{3pt}
    \caption{Comprehensive evaluation of spotting ability on in-house benchmark.}
    \vspace{-0.5em}
    \label{tab:spotting_performance}
    \resizebox{1\linewidth}{!}{
        \begin{tabular}{ll|c|ccccccccc}
            \toprule
            \textbf{Model Type}                           &
            \textbf{Model}                                &
            \textbf{Overall}                              &
            \textbf{Art}                                  &
            \textbf{Doc}                                  &
            \textbf{Game}                                 &
            \textbf{Hand}                                 &
            \textbf{Ads}                                  &
            \textbf{Receipt}                              &
            \textbf{Screen}                               &
            \textbf{Scene}                                &
            \textbf{Video}
            \\
            \midrule
            \multirow{2}{*}{\textbf{Traditional methods}} & PaddleOCR~\cite{cui2025paddleocr}           & $53.38$        & $32.83$        & $70.23$        & $51.59$        & $56.39$        & $57.38$        & $50.59$        & $63.38$        & $44.68$        & $53.35$        \\
                                                          & BaiduOCR~\cite{baiduocrapi}                 & $61.90$        & $38.5$         & \textbf{78.95} & $59.24$        & $59.06$        & $66.70$        & \textbf{63.66} & $68.18$        & $55.53$        & $67.38$        \\
            \midrule
            \multirow{4}{*}{\textbf{General VLMs}}        & Gemini-2.5-Pro~\cite{comanici2025gemini}        & $23.44$        & $21.79$        & $35.16$        & $10.02$        & $38.49$        & $29.89$        & $20.80$        & $17.59$        & $18.33$        & $18.90$        \\
            &Qwen3-VL-2B-Instruct~\cite{qwen3-vl}        & $29.68$        & $29.43$        & $19.37$        & $20.85$        & $50.57$        & $35.14$        & $24.42$        & $12.13$        & $34.90$        & $40.10$        \\
                                                          & Qwen3-VL-235B-A22B-Instruct~\cite{qwen3-vl} & $53.62$        & $46.15$        & $43.78$        & $48.00$        & $68.90$        & $64.01$        & $47.53$        & $45.91$        & $54.56$        & $63.79$        \\
                                                          & Seed-1.6-Vision~\cite{seed1.6}              & $59.23$        & $45.36$        & $55.04$        & $59.68$        & $67.46$        & $65.99$        & $55.68$        & $59.85$        & $53.66$        & $70.33$        \\
            \midrule
            \textbf{\makecell[l]{OCR-Specific VLMs}}      & \textbf{HunyuanOCR}               & \textbf{70.92} & \textbf{56.76} & 73.63          & \textbf{73.54} & \textbf{77.10} & \textbf{75.34} & 63.51          & \textbf{76.58} & \textbf{64.56} & \textbf{77.31} \\
            \bottomrule
        \end{tabular}
    }
\end{table}

\subsection{Parsing}
We systematically evaluated the model’s performance on document parsing using three benchmark datasets. First, we conducted experiments on OmniDocBench~\cite{ominibench}, a publicly available and comprehensive document parsing benchmark that includes a diverse set of digital and scanned documents covering formulas, tables, paragraphs, and various structural elements. Second, to further assess the model’s robustness in real-world captured scenarios, we created a Wild version of OmniDocBench\footnote{The Wild version of OmniDocBench will be publicly released in a future update.} by printing the original documents and re-capturing them under challenging conditions—such as manual folding, bending, and varying illumination—to simulate realistic distortions encountered in everyday document photography. Finally, we evaluated the model on DocML\footnote{The DocML multilingual parsing dataset will also be open-sourced in a future release. We invite interested parties to reach out to us for access or evaluation prior to its public release.}, our internally curated multilingual parsing dataset designed to assess robustness across multiple languages and acquisition settings. DocML spans both digital/scanned and real-world captured documents across 14 high-frequency non-Chinese/English languages, including German, Spanish, Turkish, Vietnamese, Korean, Malay, Portuguese, Russian, French, Indonesian, Thai, Italian, and Japanese.

For both OmniDocBench and its Wild variant, we followed the official evaluation protocol described in \cite{ominibench} and report results for HunyuanOCR alongside other leading document parsing models. As shown in Table~\ref{table3}, HunyuanOCR achieves the highest overall performance on both the digital/scanned and real-world captured settings, demonstrating strong generalization across diverse document formats and acquisition conditions. Notably, despite its relatively compact 1B parameter size, HunyuanOCR outperforms larger specialized OCR or VLM-based parsing models. On DocML, we adopt an overall edit-distance–based score as the evaluation metric to comprehensively measure the accuracy and robustness of parsed outputs across multilingual settings. Under this metric, HunyuanOCR demonstrates excellent multilingual parsing performance, achieving state-of-the-art results across all 14 languages. These findings collectively show that HunyuanOCR delivers robust and accurate document parsing across multilingual, multi-scene, and real-world conditions.

\begin{table}[!t]
    \small
    \centering
    \caption{Parsing performance evaluated across multilingual settings and diverse document scenarios.}
    \setlength{\abovecaptionskip}{0.2cm}
    \setlength{\tabcolsep}{0.5mm}{
        \begin{tabular}{lcc|cccc|cccc|c}
            \toprule
            \multirow{2}{*}{\raisebox{-0.7ex}{\textbf{Model Type}}} &
            \multirow{2}{*}{\raisebox{-0.7ex}{\textbf{Model}}}      &
            \multirow{2}{*}{\raisebox{-0.7ex}{\textbf{Size}}}       &
            \multicolumn{4}{c}{\textbf{OmniDocBench}}               &
            \multicolumn{4}{c}{\textbf{Wild-OmniDocBench}}          &
            \multirow{2}{*}{\raisebox{-0.7ex}{\textbf{DocML}}}
            \\
            \cmidrule(rl){4-7}
            \cmidrule(rl){8-11}
                                                                    &
                                                                    &
                                                                    &
            overall$\uparrow$                                       &
            text$\downarrow$                                        &
            formula$\uparrow$                                       &
            table$\uparrow$                                         &
            overall$\uparrow$                                       &
            text$\downarrow$                                        &
            formula$\uparrow$                                       &
            table$\uparrow$
            \\
            \midrule
            \multirow{2}{*}{\parbox{1.6cm}{\centering \textbf{General
            \\VLMs}}}
                                                                    & Gemni-2.5-pro~\citeyear{comanici2025gemini} & -    & 88.03          & 0.075 & 85.92          & 85.71          &      80.59      &        0.118        &         75.03       &    78.56     & 82.64          \\
                                                                    & Qwen3-VL-235B~\citeyear{qwen3-vl}           & 235B & 89.15          & 0.069 & 88.14          & 86.21          & 79.69          & 0.09           & 80.67          & 68.31          & 81.40          \\
            \midrule
            \multirow{3}{*}{\parbox{1.6cm}{\centering \textbf{Specialized
            \\VLMs\\(Modular)}}}
                                                                    & MonkeyOCR-pro~\citeyear{li2025monkeyocr}    & 3B   & 88.85          & 0.075 & 87.5           & 86.78          & 70.00          & 0.211          & 63.27          & 67.83          & 56.50          \\
                                                                    & MinerU2.5~\citeyear{niu2025mineru25}        & 1.2B & 90.67          & 0.047 & 88.46          & 88.22          & 70.91          & 0.218          & 64.37          & 70.15          & 52.05          \\
                                                                    & PaddleOCR-VL~\citeyear{cui2025paddleocrvl}  & 0.9B & 92.86          & 0.035 & 91.22          & 90.89          & 72.19          & 0.232          & 65.54          & 74.24          & 57.42          \\
            \midrule
            \multirow{4}{*}{\parbox{1.6cm}{\centering \textbf{Specialized
            \\VLMs\\(End2End)}}}
                                                                    & Mistral-OCR~\citeyear{mistralocr}           & -    & 78.83          & 0.164 & 82.84          & 70.03          & -              & -              & -              & -              & 64.71          \\
                                                                    & Deepseek-OCR~\citeyear{wei2025deepseekocr}  & 3B   & 87.01          & 0.073 & 83.37          & 84.97          & 74.23          & 0.178          & 70.07          & 70.41          & 57.22          \\
                                                                    & dots.ocr~\citeyear{dot2024ocr}              & 3B   & 88.41          & 0.048 & 83.22          & 86.78          & 78.01          & 0.121          & 74.23          & 71.89          & 77.50          \\
                                                                    &   \textbf{HunyuanOCR}     & 1B   & \textbf{94.10} & 0.042 & \textbf{94.73} & \textbf{91.81} & \textbf{85.21} & \textbf{0.081} & \textbf{82.09} & \textbf{81.64} & \textbf{91.03} \\
            \bottomrule
        \end{tabular}}
    \label{table3}
\end{table}

\subsection{IE \& VQA}
We systematically evaluate the model’s performance on information extraction and open-ended visual question answering tasks using three benchmark datasets. First, to assess the model’s capability on high-frequency card and document types, we constructed a test set comprising 768 samples across 30 common categories (\Cref{tab:ie_common_categories}), such as identification cards, passports, and invoices. Second, to evaluate text extraction performance in complex scenarios, we built a video subtitle dataset containing 1,000 samples covering diverse video contexts and subtitle styles. Additionally, the model was comprehensively evaluated on OCRBench~\cite{liu2024ocrbench}, a publicly available benchmark that includes 1,000 test samples and spans multiple competencies, including scene text recognition, handwritten text and formula recognition, information extraction, and open-ended question answering on documents and charts.

We evaluated the first two benchmarks using exact-match accuracy under a unified prompting protocol for multi-field JSON outputs, while adopting the official standard evaluation protocol for OCRBench. HunyuanOCR was compared against leading SOTA VLMs, including Qwen3VL-235B-Instruct, Seed1.6-VL-Instruct, and Gemini-2.5-Pro, using identical prompts and post-processing procedures such as JSON format parsing. As summarized in Table~\ref{tab:ie_performance}, HunyuanOCR achieves the highest overall accuracy across all 30 document categories in card/receipts information extraction and subtitle extraction tasks, despite having only around 1B parameters, significantly outperforming considerably larger VLMs such as Qwen3VL-235B-Instruct, Seed1.6-VL, and Gemini-2.5-Pro. On OCRBench, HunyuanOCR also demonstrates substantially better performance than DeepseekOCR at a similar scale and comparable with the larger Qwen3VL-2B-Instruct model.


\begin{table}[!t]
    \small
    \centering
    \caption{Evaluation of information extraction and visual question answering tasks.}
    \setlength{\tabcolsep}{3.5mm}
    \begin{tabular}{lccc|c}
        \toprule
        \textbf{Model}                                  & \textbf{Cards} & \textbf{Receipts} & \textbf{Video Subtitles} & \textbf{OCRBench} \\
        \midrule
        DeepSeek-OCR~\cite{wei2025deepseekocr}          & 10.04          & 40.54             & 5.41                     & 430               \\
        PP-ChatOCR~\cite{ppocr}                         & 57.02          & 50.26             & 3.1                      & -                 \\
        Qwen3-VL-2B-Instruct~\citeyear{qwen3-vl}        & 67.62          & 64.62             & 3.75                     & 858               \\
        Seed-1.6-Vision~\cite{seed1.6}                  & 70.12          & 67.5              & 60.45                    & 881               \\
        Qwen3-VL-235B-A22B-Instruct~\citeyear{qwen3-vl} & 75.59          & 78.4              & 50.74                    & \textbf{920}      \\
        Gemini-2.5-Pro~\citeyear{comanici2025gemini}    & 80.59          & 80.66             & 53.65                    & 872               \\
        \midrule
        \textbf{HunyuanOCR}                       & \textbf{92.29} & \textbf{92.53}    & \textbf{92.87}           & 860               \\
        \bottomrule
    \end{tabular}
    \label{tab:ie_performance}
\end{table}


\subsection{Text Image Translation}

We systematically evaluated the model’s text image translation capability using two benchmark datasets. For public benchmarking, we selected DoTA~\cite{Dota}, a document translation dataset designed for complex and diverse English-layout document scenarios, and used it to assess the model’s English-to-Chinese translation performance under realistic document conditions. In addition, we constructed an in-house evaluation benchmark based on DocML, where each sample is annotated with both English and Chinese translations. This internal benchmark enables a comprehensive assessment of translation robustness across multiple languages and a broad range of document types, including both digital/scanned and real-world captured scenes.

To evaluate translation quality, we adopt the COMET~\cite{comet} metric, a widely used neural-based evaluation standard for machine translation. As summarized in Table~\ref{table_trans}, HunyuanOCR surpasses VLMs with over 8B parameters on DoTA, demonstrating strong translation performance in complex document layouts despite its compact 1B scale.
Furthermore, we achieved first place in the Track 2.2 OCR-free Small Model of the ICDAR 2025 Competition on End-to-End Document Image Machine Translation Towards Complex Layouts~\cite{zhang2025icdar}, validating the effectiveness and generality of our approach.
On the DocML evaluation set, HunyuanOCR again outperforms several larger VLMs exceeding 4B parameters, highlighting its robust multilingual translation capability across diverse layouts, languages, and acquisition conditions. These findings collectively demonstrate that HunyuanOCR provides a highly efficient yet powerful solution for text image translation in both public benchmarks and real-world multilingual scenarios.

However, due to its relatively small language model, HunyuanOCR’s translation capability lags behind its strong text detection, recognition, and document parsing performance. For applications requiring higher translation accuracy, developers can cascade our multilingual parsing module with Hunyuan-MT-7B\footnote{https://huggingface.co/tencent/Hunyuan-MT-7B} or await our upcoming general vision-language models to further boost overall translation quality.

\begin{table}[!t]
    \small
    \centering
    \caption{Evaluation of photo translation. We additionally manually annotated DocML with high-quality English and Chinese reference translations to serve as ground-truth labels for evaluating text translation performance.}
    \setlength{\abovecaptionskip}{0.2cm}
    \setlength{\tabcolsep}{3.5mm}{
        \begin{tabular}{lcccc}
            \toprule
            \multirow{2}{*}{\raisebox{-0.7ex}{\textbf{Model}}} &
            \multirow{2}{*}{\raisebox{-0.7ex}{\textbf{Size}}}  &
            \multicolumn{2}{c}{\textbf{DocML}}    &
            \textbf{DoTA}                                                                           \\
            \cmidrule(rl){3-4}
            \cmidrule(rl){5-5}
                                                               &      & other2en & other2zh & en2zh \\
            \midrule
            Gemni-2.5-Flash~\cite{comanici2025gemini}          & -    & \textbf{79.26}    & \textbf{80.06}    & \textbf{85.60} \\
            Qwen3-VL-235B-Instruct~\cite{qwen3-vl}             & 235B & 73.67    & 77.20    & 80.01 \\
            Qwen3-VL-8B-Instruct~\cite{qwen3-vl}               & 8B   & 75.09    & 75.63    & 79.86 \\
            Qwen3-VL-4B-Instruct~\cite{qwen3-vl}               & 4B   & 70.38    & 70.29    & 78.45 \\
            Qwen3-VL-2B-Instruct~\cite{qwen3-vl}               & 2B   & 66.30    & 66.77    & 73.49 \\
            PP-DocTranslation                                  &  -   & 52.63    & 52.43    & 82.09\\
            \textbf{HunyuanOCR}              & 1B   & 73.38    & 73.62    & 83.48 \\
            \bottomrule
        \end{tabular}
    }
    \label{table_trans}
\end{table}

%% file: sec/conclusion.tex
\section{Conclusion}

In this paper, we present HunyuanOCR, an open-source expert vision-language model that unifies diverse OCR tasks within a lightweight, end-to-end architecture. Our work demonstrates that a compact model with only 1B parameters can achieve competitive performance against larger general-purpose VLMs and traditional pipeline systems, validating the effectiveness of our data-centric training strategy and targeted reinforcement learning approach.
HunyuanOCR achieves state-of-the-art results in text spotting, document parsing, and information extraction, while significantly simplifying deployment pipelines. These advancements align with our original goal of balancing versatility with efficiency, as outlined in the abstract.
Looking ahead, we will continue to optimize inference efficiency through token compression and architectural improvements, while expanding the model's capability to handle higher-resolution and multi-page documents. Our long-term goal remains to adapt HunyuanOCR for edge-device deployment, further democratizing robust OCR intelligence for real-world applications.

\section*{Contributors}
\begin{itemize} [leftmargin=*]
    \item Project Sponsors: Jie Jiang, Linus
    \item Project Supervisor: Han Hu
    \item Project Leader: Chengquan Zhang
    \item Core Contributors: Pengyuan Lyu, Xingyu Wan, Gengluo Li, Shangpin Peng
    \item Contributors: Weinong Wang, Liang Wu, Huawen Shen, Yu Zhou, Canhui Tang, Qi Yang, Qiming Peng, Bin Luo, Hower Yang, Xinsong Zhang, Jinnian Zhang, Houwen Peng, Hongming Yang, Senhao Xie, Longsha Zhou, Ge Pei, Binghong Wu, Rui Yan, Kan Wu, Jieneng Yang, Bochao Wang, Kai Liu, Jianchen Zhu
\end{itemize}

%% file: sec/appendix.tex
\appendix
{
    \newpage
    \par\nolinenumbers
    \centering
    \Large
    \textbf{HunyuanOCR Technical Report}\\
    \vspace{0.5em}
    Supplementary Material \\
    \vspace{0.5em}
    \linenumbers
}

This material provides supplementary details to the main paper, including the following sections:

\vspace{-0.5em}
\begin{itemize}[itemsep=0pt, parsep=2pt, leftmargin=*]
    \item (\ref{sec:recommend_instruction}) \textbf{Recommended Instruction}
    \item (\ref{sec:common_ie_categories}) \textbf{Common Supported IE Categories}
    \item (\ref{sec:reinforcement_learning_details}) \textbf{Reinforcement Learning Details}
          \begin{itemize}
              \item (\ref{subsec:appendix_rl_setup}) Training Configuration
              \item (\ref{subsec:appendix_rl_dynamics}) Training Dynamics
              \item (\ref{subsec:appendix_rl_improvements}) Task-wise Performance Improvements
          \end{itemize}
    \item (\ref{sec:Parsing_Omni_element}) \textbf{Element-level Evaluation}
    \item (\ref{sec:qualitive_examples}) \textbf{Qualitative examples}
\end{itemize}

\section{Recommended Instruction}
\label{sec:recommend_instruction}

\input{sec/appendix/instructions}

\section{Common Supported IE Categories}
\label{sec:common_ie_categories}
\input{sec/appendix/ie_list}

\section{Reinforcement Learning Details}
\label{sec:reinforcement_learning_details}
\input{sec/appendix/RL}

\section{Element-level Evaluation}
\label{sec:Parsing_Omni_element}
\input{sec/appendix/Parsing}

\newpage
\section{Qualitative examples}
\label{sec:qualitive_examples}

\input{sec/appendix/spotting}
\input{sec/appendix/parsing_fig}
\input{sec/appendix/parsing_table}
\input{sec/appendix/translation}

\input{sec/appendix/ie_vqa}

%% file: sec/appendix/instructions.tex
\Cref{tab:task_descriptions} summarizes the recommended instructions for each task supported by HunyuanOCR, and provides a bilingual (Chinese–English) reference. We recommend using the Chinese instructions to ensure the stability and reproducibility of benchmarking results.

\begin{table}[htbp]
    \centering
    \caption{Bilingual (Chinese–English) Instructions Recommended for Different Task Types.}
    \vspace{-0.5em}
    \label{tab:task_descriptions}
    \begin{tabular}{>{\centering\arraybackslash}m{2cm} m{5.5cm} m{6.5cm}}
        \toprule
        \textbf{Task}    &
        \textbf{Chinese} &
        \textbf{English}
        \\
        \midrule
        \textbf{Spotting}
                         & 检测并识别图片中的文字，将文本坐标格式化输出。
                         & Detect and recognize text in the image, and output the text coordinates in a formatted manner.
        \\
        \midrule

        \multirow{14}{*}{\textbf{Parsing}}
                         & 识别图片中的公式，用 \LaTeX 格式表示。
                         & Identify the formula in the image and represent it using \LaTeX format.
        \\
        \cmidrule(lr){2-3}
                         & 把图中的表格解析为 HTML。
                         & Parse the table in the image into HTML.
        \\
        \cmidrule(lr){2-3}
                         & 解析图中的图表，对于流程图使用 Mermaid 格式表示，其他图表使用 Markdown 格式表示。
                         & Parse the chart in the image; use Mermaid format for flowcharts and Markdown for other charts.
        \\
        \cmidrule(lr){2-3}
                         & 提取文档图片中正文的所有信息用 markdown 格式表示，其中页眉、页脚部分忽略，表格用 html 格式表达，文档中公式用 \LaTeX 格式表示，按照阅读顺序组织进行解析。
                         & Extract all information from the main body of the document image and represent it in markdown format, ignoring headers and footers. Tables should be expressed in HTML format, formulas in the document should be represented using \LaTeX format, and the parsing should be organized according to the reading order.
        \\
        \cmidrule(lr){2-3}
                    & 提取图中的文字。
                    & Extract the text in the image.
        \\
        \midrule

        \multirow{6}{*}{\textbf{\makecell{Information
        \\Extraction}}}
                         & 输出 $\mathrm{Key}$ 的值。
                         & Output the value of $\mathrm{Key}$.
        \\
        \cmidrule(lr){2-3}
                         & 提取图片中的: [`key1',`key2', \dots] 的字段内容，并按照 JSON 格式返回。
                         & Extract the content of the fields: [`key1',`key2', \dots] from the image and return it in JSON format.
        \\
        \cmidrule(lr){2-3}
                         & 提取图片中的字幕。
                         & Extract the subtitles from the image.
        \\
        \midrule
        \multirow{3}{*}{\textbf{Translation}}
                         & 先解析文档，再将文档内容翻译为中文，其中页眉、页脚忽略，公式用 \LaTeX 格式表示，表格用html格式表示。
                         & First parse the document, then translate its content into Chinese. Ignore headers and footers; represent equations in \LaTeX; and render tables in HTML format.
        \\
        \cmidrule(lr){2-3}
                         & 提取图中文字，并将其翻译成中文/英文。
                         & Extract all text from the image and translate it into Chinese/English.
        \\
        \bottomrule
    \end{tabular}
\end{table}

%% file: sec/appendix/ie_list.tex
\Cref{tab:ie_common_categories} summarizes the common card and receipt types covered by the 30 IE tasks. The card side includes more than ten categories, such as ID cards, band cards, passports, social security cards, business licenses, driver’s licenses, vehicle licenses, etc.. 
The receipt side also spans over ten types, including shopping receipts, taxi receipts, VAT invoices, train tickets, bus tickets, itineraries, bank slips, etc..

\begin{table}[htbp]
    \centering
    \caption{Common document categories for IE tasks, grouped into Cards \& Certificates and Receipts.}
    \vspace{-0.5em}
    \label{tab:ie_common_categories}
    \begin{tabular}{p{6.5cm} p{7.5cm}}
        \toprule
        \textbf{Cards \& Certificates}          & \textbf{Receipts}         \\
        \midrule
        ID cards                                 & Shopping receipts          \\
        Bank cards                               & Taxi receipts              \\
        Social security cards                    & Ferry tickets             \\
        Passports                                & Train tickets              \\
        Household registers                      & Bus tickets                \\
        Mainland travel permits                  & Itineraries                 \\
        Business licenses                       & Express waybills           \\
        Driver's licenses                        & VAT invoices               \\
        Public institution certificates          & Bank slips                 \\
        Vehicle licenses                         & Medical inspection reports \\
        Professional qualification certificates & Prescriptions              \\
        Tax registration certificates            & Medical records            \\
        Medical insurance vouchers               & Checks                     \\
        Road transport certificates              & Ride-hailing itineraries    \\
        Vehicle certificates of conformity      & Takeout orders            \\
        \bottomrule
    \end{tabular}
\end{table}

%% file: sec/appendix/RL.tex
In this section, we provide additional details on the reinforcement learning (RL) stage of HunyuanOCR beyond the description in the main text~\cref{sec:rl}. We first summarize the RL training configuration (\cref{subsec:appendix_rl_setup}), then present the training dynamics (\cref{subsec:appendix_rl_dynamics}), and finally analyze the performance improvements brought by RL training (\cref{subsec:appendix_rl_improvements}).

\subsection{Training Configuration}
\label{subsec:appendix_rl_setup}

The detailed training setup for RL is listed in~\cref{tab:rl_training_setup}. We adopt a constant learning rate schedule with Adam optimizer, a large global batch size, and long-context settings to fully exploit the long-document understanding capability of HunyuanOCR. No explicit KL penalty is applied during RL, allowing the policy to adjust more freely under the guidance of task-specific rewards. For rollout generation, we use a low temperature of 0.85 and sample $N=8$ responses per prompt to obtain a diverse set of candidates for reward evaluation and policy updates.

\subsection{Training Dynamics}
\label{subsec:appendix_rl_dynamics}

\begin{figure}[t]
    \centering
    \includegraphics[width=0.95\linewidth]{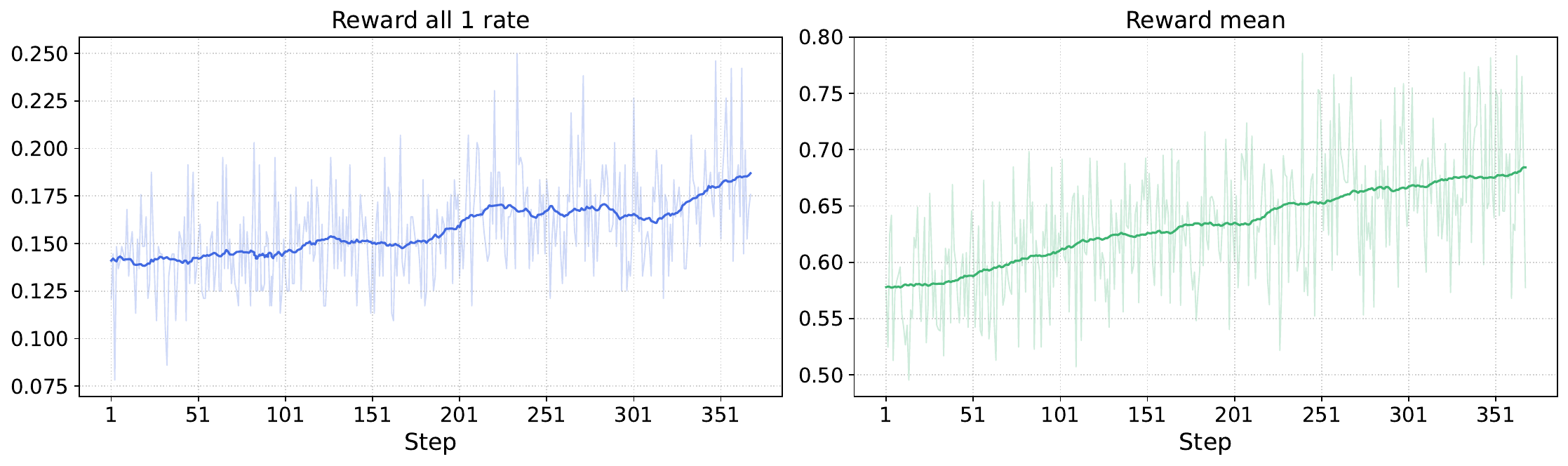}
    \vspace{-2mm}
    \caption{Training dynamics of RL. We show the proportions of all-one rewards and the mean reward value, which increases steadily over the course of training.}
    \label{fig:RL_training_dynamics}
\end{figure}

The training dynamics of the RL stage are visualized in \cref{fig:RL_training_dynamics}, where we track two key statistics: the proportion of samples receiving reward 1 at each step, and the mean reward value. As training progresses, the mean reward increases steadily. This consistent upward trend indicates that the policy gradually learns to produce outputs that better satisfy the task-specific reward criteria, validating the effectiveness and stability of the RL process.

\subsection{Task-wise Performance Improvements}
\label{subsec:appendix_rl_improvements}

After RL training, we observe substantial gains across multiple OCR-related tasks.

\textbf{Spotting.}
The spotting ability of HunyuanOCR improves significantly, especially on \emph{Art} and \emph{Screen} scenarios, where the scores increase by more than 2 points. We attribute these gains to the rule-based reward design for the spotting task, which can assess the discrepancy between the predicted outputs and ground-truth annotations at a fine-grained level. This encourages the model to simultaneously improve both the accuracy of the predicted bounding boxes and the correctness of the recognized text.

\textbf{Parsing.}
For the parsing task, the score on OmniDocBench increases from 92.5 to 94.1 after RL training. This improvement further demonstrates the effectiveness of the rule-based reward design, which precisely measures content consistency between the model’s outputs and the reference text.

\textbf{Information Extraction, VQA, and Translation.}
In addition, the information extraction (IE) task improves by about 2 points, the average score on OCRBench increased by 3.3, and the text image translation task also shows noticeable gains. These results indicate that the LLM-as-a-judge–based reward design can effectively guide the model to produce more faithful and semantically accurate outputs in higher-level understanding tasks.

\textbf{Discussion.}
In summary, we attribute the effectiveness of RL in HunyuanOCR primarily to two factors:

\begin{itemize}[leftmargin=*]
    \item \textbf{High-quality training data.} Carefully curated and diverse RL training data provide a solid foundation for the model to learn robust behaviors across spotting, parsing, IE, and translation scenarios.
    \item \textbf{Fine-grained reward design.} Task-specific, fine-grained reward functions (both rule-based and LLM-as-a-judge–based) allow the model to receive precise feedback on multiple aspects of its outputs, leading to balanced improvements in recognition accuracy, structural parsing, and semantic understanding.
\end{itemize}

These elements work together to make the RL stage an effective complement to supervised training, yielding a more capable and reliable HunyuanOCR model.



\begin{table}[t!]
    \centering
    \caption{
        Training setup for reinforcement learning.
    }
    \vspace{-2mm}
    \resizebox{0.4\columnwidth}{!} {%
        \begin{tabular}{l|c}
            \toprule
            \textbf{Setting}         & \textbf{Value}                \\
            \midrule
            \rowcolor{gray!20}  \multicolumn{2}{l}{\textbf{Actor}}   \\
            Learning rate            & 8e-7                          \\
            Micro batch size per GPU & 1                             \\
            Optimizer                & Adam                          \\
            Lr schedule              & Constant                      \\
            zero-stage               & 3                             \\
            Global batch size        & 512                           \\
            Max prompt length        & 6144                          \\
            Max response length      & 16384                         \\
            KL loss coefficient      & 0                             \\
            \rowcolor{gray!20}  \multicolumn{2}{l}{\textbf{Rollout}} \\
            Temperature              & 0.85                          \\
            N                        & 8                             \\
            Top-p                    & 0.95                          \\
            Tok-k                    & 50                            \\
            \bottomrule
        \end{tabular}
    }
    \vspace{-4mm}
    \label{tab:rl_training_setup}
\end{table}

%% file: sec/appendix/Parsing.tex
To conduct a focused evaluation of formula recognition and table recognition, we apply targeted cropping to formula regions and table  within the OmniDocBench1.5 benchmark. This preprocessing step isolates the relevant structural components, enabling a more precise assessment of the corresponding capabilities of HunyuanOCR.

\begin{table}[!t]
    \small
    \centering
    \caption{Comparison of OmniDocBench-Table-block and OmniDocBench-Formula-block Performance.}
    \setlength{\abovecaptionskip}{0.2cm}
    \setlength{\tabcolsep}{1mm}{
        \begin{tabular}{l|cc|ccc}
            \toprule
            \multirow{2}{*}{\textbf{Model}}      &
            \multicolumn{2}{c}{\textbf{Table}}               &
            \multicolumn{3}{c}{\textbf{Formula}}          
            \\
            \cmidrule(rl){2-3}
            \cmidrule(rl){4-6}
                                                                    &
            Overall TEDS$\uparrow$                                            &
            Structural TEDS$\uparrow$                                         &
            Overall CDM$\uparrow$                                       &
            EN CDM$\uparrow$                                        &
            ZH CDM$\uparrow$                                       
            \\
            \midrule
         
                                                                     MinerU2-VLM\cite{wang2024mineruopensourcesolutionprecise} & 0.9002    & 0.9369          & 0.8286 & 0.9616          & 0.6956                    \\
                                                                     dots.ocr & 0.8194    & 0.8442          & 0.4641 & 0.4868          & 0.4414                    \\
                                                                     MinerU2.5 & 0.9005    & 0.9539          & 0.9187 & \textbf{0.9751}          & 0.8623                    \\
                                                                     PaddleOCR-VL & 0.9195    & 0.9543          & 0.9453 & 0.9677          & 0.9228                    \\
                                                                     HunyuanOCR & \textbf{0.9574}    & \textbf{0.9771}          & \textbf{0.9695} & 0.9706          & \textbf{0.9645}                    \\

            \bottomrule
        \end{tabular}}
    \label{table_element}
\end{table}

For evaluating cropped formulas, we employ the prompt “识别图片中的公式，用 \LaTeX 格式表示。” Similarly, for assessing cropped table elements, we use the prompt “把图中的表格解析为 HTML。” The corresponding results are summarized in Table \ref{table_element}. Across both tasks, HunyuanOCR demonstrates strong capability in fine-grained element parsing. These findings suggest that, when high-precision recognition of specific sub-elements is required, users may leverage prompts tailored separately for formulas and tables to achieve optimal inference performance.

%% file: sec/appendix/spotting.tex
\renewcommand{\arraystretch}{1.5}

\begin{figure}[!h]
    \centering
    \begin{tabular}{@{}c@{\hspace{0.5cm}}c@{}}
        \begin{minipage}{0.45\textwidth}
            \centering
            Input image \\
            \includegraphics[width=0.95\linewidth]{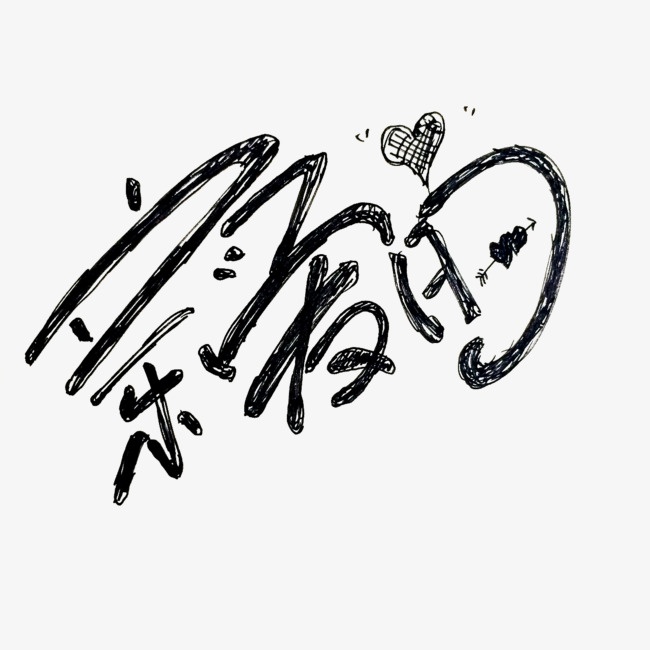}
        \end{minipage}
         &
        \begin{minipage}{0.45\textwidth}
            \centering
            Visualization of HunyuanOCR output \\
            \includegraphics[width=0.95\linewidth]{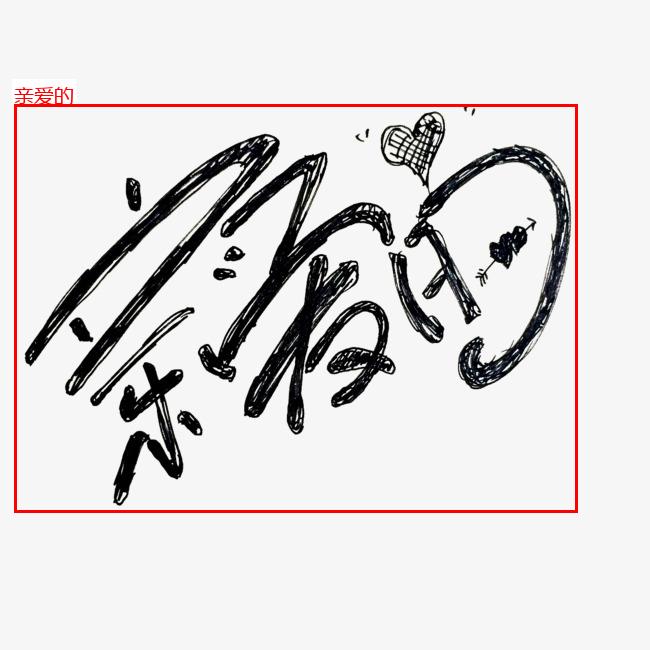}
        \end{minipage}
    \end{tabular}

    \vspace{0.5cm}
    \begin{minipage}{0.9\textwidth}
        \textbf{Prompt:} 检测并识别图片中的文字，将文本坐标格式化输出。
        \\
        \\
        \textbf{HunyuanOCR:} \texttt{<ref>亲爱的</ref><quad>(22,160),(888,788)</quad>}
    \end{minipage}
    \caption{Robust Text Spotting Results of HunyuanOCR on Artistic Font.}
    \label{fig:spotting11}
\end{figure}

\begin{figure}[!h]
    \centering
    \begin{tabular}{@{}c@{\hspace{0.2cm}}c@{}}
        \begin{minipage}{0.47\textwidth}
            \centering
            Input image \\
            \includegraphics[width=0.95\linewidth]{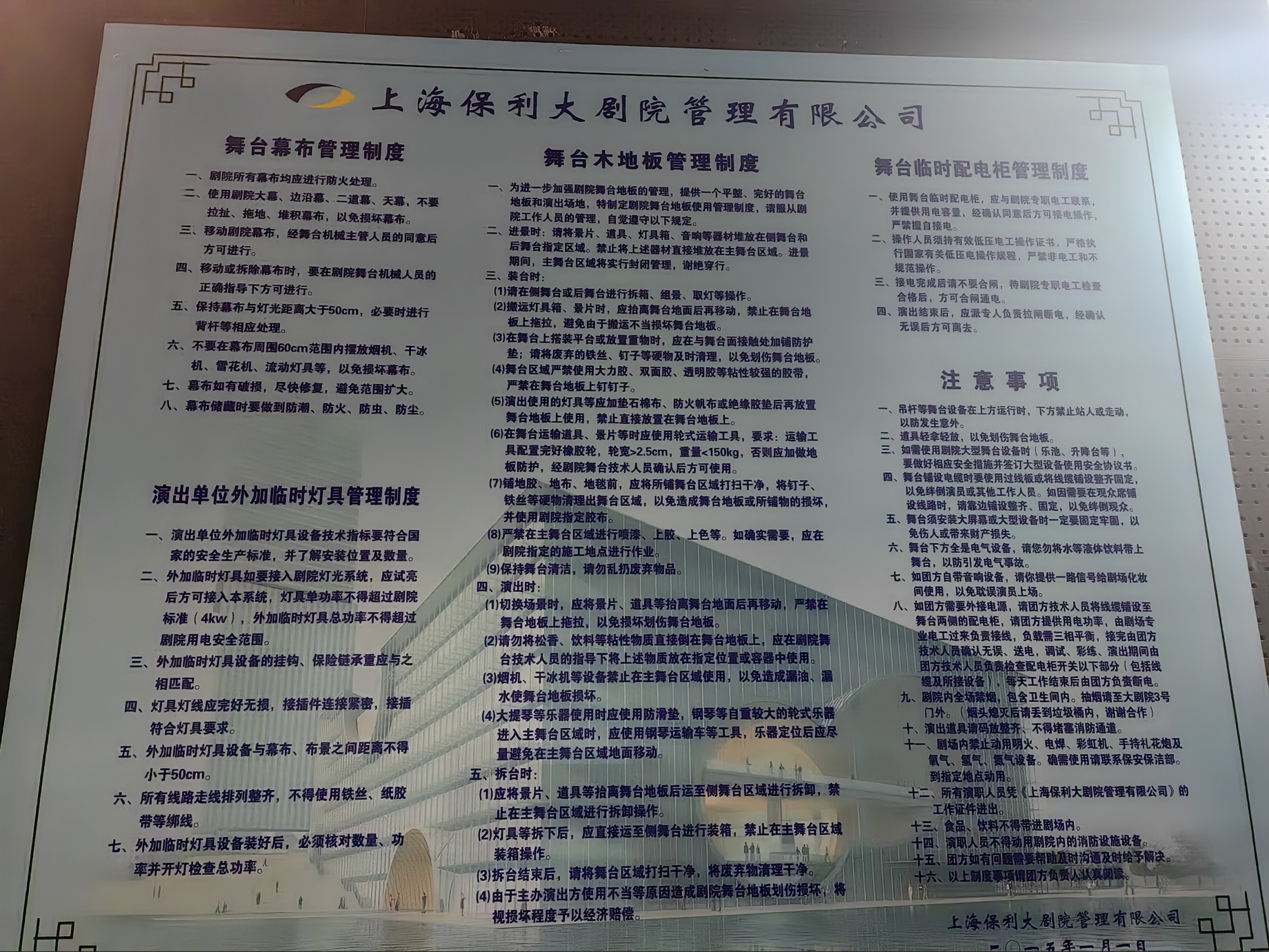}
        \end{minipage}
         &
        \begin{minipage}{0.47\textwidth}
            \centering
            Visualization of HunyuanOCR output \\
            \includegraphics[width=0.95\linewidth]{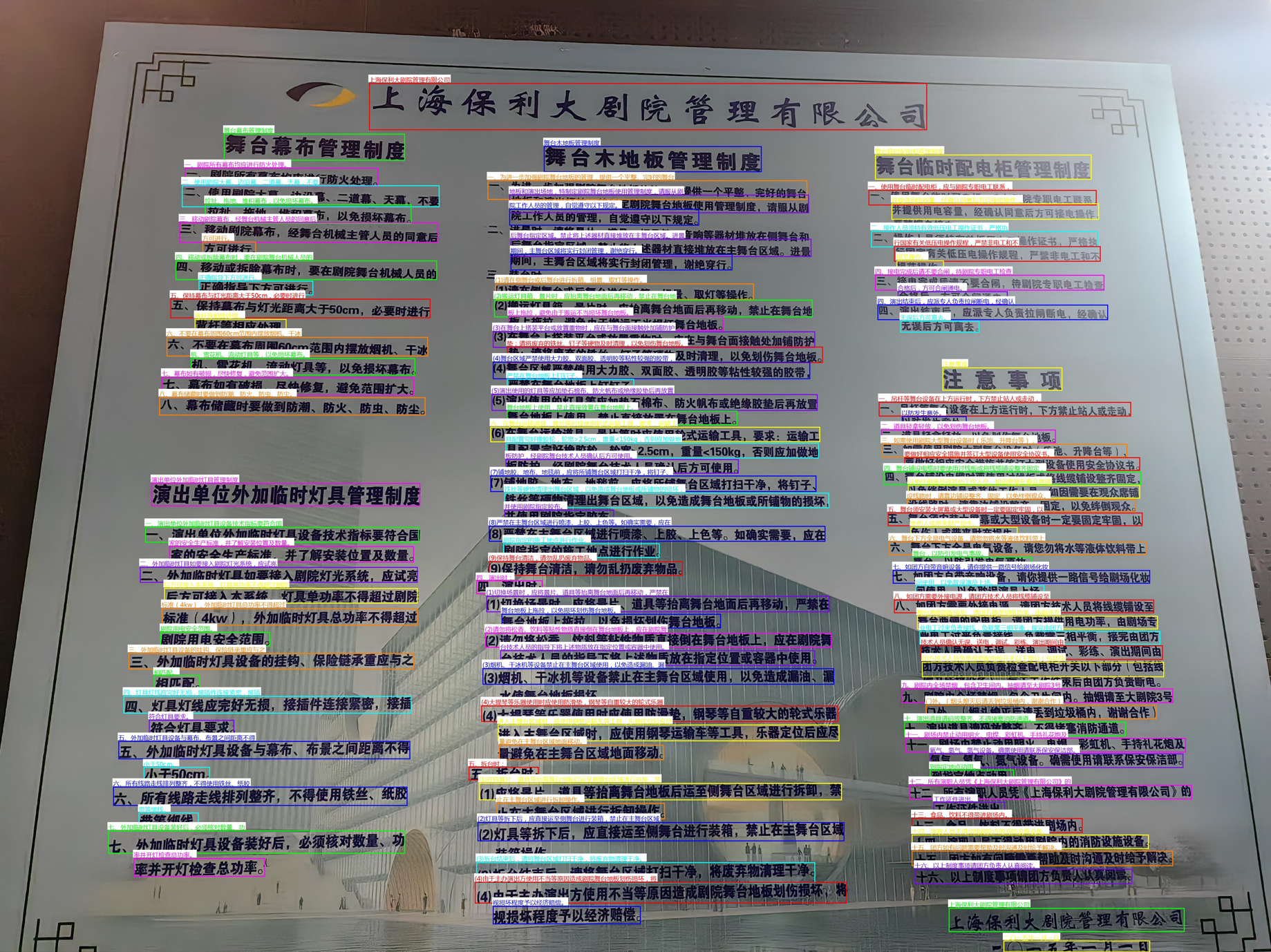}
        \end{minipage}
    \end{tabular}
    \vspace{0.2cm}
    \begin{minipage}{0.9\textwidth}
        \textbf{Prompt:} 检测并识别图片中的文字，将文本坐标格式化输出。
        \\
        \textbf{HunyuanOCR:}
        {
            \fontsize{3pt}{3pt}\selectfont 
            \textless ref\textgreater上海保利大剧院管理有限公司\textless /ref\textgreater\textless quad\textgreater(291,90),(729,139)\textless /quad\textgreater\textless ref\textgreater舞台幕布管理制度\textless /ref\textgreater\textless quad\textgreater(177,143),(319,171)\textless /quad\textgreater\textless ref\textgreater舞台木地板管理制度\textless /ref\textgreater\textless quad\textgreater(428,156),(599,183)\textless /quad\textgreater\textless ref\textgreater舞台临时配电柜管理制度\textless /ref\textgreater\textless quad\textgreater(688,165),(858,191)\textless /quad\textgreater\textless ref\textgreater一、剧院所有幕布均应进行防火处理。\textless /ref\textgreater\textless quad\textgreater(146,179),(298,198)\textless /quad\textgreater\textless ref\textgreater二、使用剧院大幕、边沿幕、二道幕、天幕，不要\textless /ref\textgreater\textless quad\textgreater(144,197),(346,220)\textless /quad\textgreater\textless ref\textgreater一、为进一步加强剧院舞台地板的管理，提供一个平整、完好的舞台\textless /ref\textgreater\textless quad\textgreater(384,192),(635,212)\textless /quad\textgreater\textless ref\textgreater地板和演出场地，特制定剧院舞台地板使用管理制度，请服从剧\textless /ref\textgreater\textless quad\textgreater(401,207),(636,226)\textless /quad\textgreater\textless ref\textgreater一、使用舞台临时配电柜，应与剧院专职电工联系，\textless /ref\textgreater\textless quad\textgreater(683,202),(862,218)\textless /quad\textgreater\textless ref\textgreater拉扯、拖地、堆积幕布，以免损坏幕布。\textless /ref\textgreater\textless quad\textgreater(162,217),(324,237)\textless /quad\textgreater\textless ref\textgreater院工作人员的管理，自觉遵守以下规定。\textless /ref\textgreater\textless quad\textgreater(401,222),(550,239)\textless /quad\textgreater\textless ref\textgreater并提供用电容量，经确认同意后方可接电操作，\textless /ref\textgreater\textless quad\textgreater(701,216),(864,233)\textless /quad\textgreater\textless ref\textgreater三、移动剧院幕布，经舞台机械主管人员的同意后\textless /ref\textgreater\textless quad\textgreater(142,236),(345,257)\textless /quad\textgreater\textless ref\textgreater二、操作人员须持有效低压电工操作证书，严格执\textless /ref\textgreater\textless quad\textgreater(685,245),(863,262)\textless /quad\textgreater\textless ref\textgreater方可进行。\textless /ref\textgreater\textless quad\textgreater(160,256),(202,271)\textless /quad\textgreater\textless ref\textgreater后舞台指定区域。禁止将上述器材直接堆放在主舞台区域。进景\textless /ref\textgreater\textless quad\textgreater(402,253),(638,272)\textless /quad\textgreater\textless ref\textgreater行国家有关低压电操作规程，严禁非电工和不\textless /ref\textgreater\textless quad\textgreater(703,261),(865,277)\textless /quad\textgreater\textless ref\textgreater四、移动或拆除幕布时，要在剧院舞台机械人员的\textless /ref\textgreater\textless quad\textgreater(139,276),(344,296)\textless /quad\textgreater\textless ref\textgreater期间，主舞台区域将实行封闭管理，谢绝穿行。\textless /ref\textgreater\textless quad\textgreater(402,269),(576,286)\textless /quad\textgreater\textless ref\textgreater规范操作。\textless /ref\textgreater\textless quad\textgreater(705,276),(742,289)\textless /quad\textgreater\textless ref\textgreater四、接电完成后请不要合闸，待剧院专职电工检查\textless /ref\textgreater\textless quad\textgreater(688,291),(868,307)\textless /quad\textgreater\textless ref\textgreater正确指导下方可进行。\textless /ref\textgreater\textless quad\textgreater(157,297),(247,312)\textless /quad\textgreater\textless ref\textgreater(1)请在侧舞台或后舞台进行拆箱、组景、取灯等操作。\textless /ref\textgreater\textless quad\textgreater(390,300),(593,317)\textless /quad\textgreater\textless ref\textgreater合格后，方可合闸通电。\textless /ref\textgreater\textless quad\textgreater(706,308),(792,322)\textless /quad\textgreater\textless ref\textgreater五、保持幕布与灯光距离大于50cm，必要时进行\textless /ref\textgreater\textless quad\textgreater(135,316),(339,336)\textless /quad\textgreater\textless ref\textgreater(2)搬运灯具箱、景片时，应抬离舞台地面后再移动，禁止在舞台地\textless /ref\textgreater\textless quad\textgreater(389,317),(639,335)\textless /quad\textgreater\textless ref\textgreater四、演出结束后，应派专人负责拉闸断电，经确认\textless /ref\textgreater\textless quad\textgreater(690,322),(871,338)\textless /quad\textgreater\textless ref\textgreater背杆等相应处理。\textless /ref\textgreater\textless quad\textgreater(154,337),(226,352)\textless /quad\textgreater\textless ref\textgreater板上拖拉，避免由于搬运不当损坏舞台地板。\textless /ref\textgreater\textless quad\textgreater(400,334),(569,349)\textless /quad\textgreater\textless ref\textgreater无误后方可离去。\textless /ref\textgreater\textless quad\textgreater(708,339),(770,351)\textless /quad\textgreater\textless ref\textgreater六、不要在幕布周围60cm范围内摆放烟机、干冰\textless /ref\textgreater\textless quad\textgreater(132,356),(337,376)\textless /quad\textgreater\textless ref\textgreater(3)在舞台上搭装平台或放置重物时，应在与舞台面接触处加铺防护\textless /ref\textgreater\textless quad\textgreater(388,350),(641,367)\textless /quad\textgreater\textless ref\textgreater垫；请将废弃的铁丝、钉子等硬物及时清理，以免划伤舞台地板。\textless /ref\textgreater\textless quad\textgreater(399,366),(647,383)\textless /quad\textgreater\textless ref\textgreater机、雪花机、流动灯具等，以免损坏幕布。\textless /ref\textgreater\textless quad\textgreater(151,378),(327,396)\textless /quad\textgreater\textless ref\textgreater(4)舞台区域严禁使用大力胶、双面胶、透明胶等粘性较强的胶带，\textless /ref\textgreater\textless quad\textgreater(388,382),(637,400)\textless /quad\textgreater\textless ref\textgreater注意事项\textless /ref\textgreater\textless quad\textgreater(741,388),(835,412)\textless /quad\textgreater\textless ref\textgreater七、幕布如有破损，尽快修复，避免范围扩大。\textless /ref\textgreater\textless quad\textgreater(128,398),(326,417)\textless /quad\textgreater\textless ref\textgreater严禁在舞台地板上钉钉子。\textless /ref\textgreater\textless quad\textgreater(399,400),(499,414)\textless /quad\textgreater\textless ref\textgreater八、幕布储藏时要做到防潮、防火、防虫、防尘。\textless /ref\textgreater\textless quad\textgreater(126,419),(335,438)\textless /quad\textgreater\textless ref\textgreater(5)演出使用的灯具等应加垫石棉布、防火帆布或绝缘胶垫后再放置\textless /ref\textgreater\textless quad\textgreater(387,416),(643,433)\textless /quad\textgreater\textless ref\textgreater一、吊杆等舞台设备在上方运行时，下方禁止站人或走动，\textless /ref\textgreater\textless quad\textgreater(691,424),(889,439)\textless /quad\textgreater\textless ref\textgreater舞台地板上使用，禁止直接放置在舞台地板上。\textless /ref\textgreater\textless quad\textgreater(399,433),(580,448)\textless /quad\textgreater\textless ref\textgreater以防发生意外。\textless /ref\textgreater\textless quad\textgreater(709,439),(760,451)\textless /quad\textgreater\textless ref\textgreater六、在舞台运输道具、景片等时应使用轮式运输工具，要求：运输工\textless /ref\textgreater\textless quad\textgreater(386,450),(645,466)\textless /quad\textgreater\textless ref\textgreater二、道具轻拿轻放，以免划伤舞台地板。\textless /ref\textgreater\textless quad\textgreater(693,453),(830,467)\textless /quad\textgreater\textless ref\textgreater具配置完好橡胶轮，轮宽\textgreater2.5cm，重量\textless 150kg，否则应加做地\textless /ref\textgreater\textless quad\textgreater(398,467),(644,483)\textless /quad\textgreater\textless ref\textgreater三、如需使用剧院大型舞台设备时（乐池、升降台等），\textless /ref\textgreater\textless quad\textgreater(693,468),(886,482)\textless /quad\textgreater\textless ref\textgreater板防护，经剧院舞台技术人员确认后方可使用。\textless /ref\textgreater\textless quad\textgreater(398,484),(581,499)\textless /quad\textgreater\textless ref\textgreater要做好相应安全措施并签订大型设备使用安全协议书。\textless /ref\textgreater\textless quad\textgreater(711,482),(896,496)\textless /quad\textgreater\textless ref\textgreater四、舞台铺设电缆时要使用过线板或将线缆铺设整齐固定，\textless /ref\textgreater\textless quad\textgreater(695,497),(897,511)\textless /quad\textgreater\textless ref\textgreater(7)铺地胶、地布、地毯前，应将所铺舞台区域打扫干净，将钉子、\textless /ref\textgreater\textless quad\textgreater(386,501),(642,518)\textless /quad\textgreater\textless ref\textgreater以免绊倒演员或其他工作人员。如因需要在观众席铺\textless /ref\textgreater\textless quad\textgreater(714,511),(896,525)\textless /quad\textgreater\textless ref\textgreater演出单位外加临时灯具管理制度\textless /ref\textgreater\textless quad\textgreater(120,509),(331,533)\textless /quad\textgreater\textless ref\textgreater铁丝等硬物清理出舞台区域，以免造成舞台地板或所铺物的损坏，\textless /ref\textgreater\textless quad\textgreater(397,519),(652,535)\textless /quad\textgreater\textless ref\textgreater设线路时，请靠边铺设整齐、固定，以免绊倒观众。\textless /ref\textgreater\textless quad\textgreater(713,526),(893,539)\textless /quad\textgreater\textless ref\textgreater并使用剧院指定胶布。\textless /ref\textgreater\textless quad\textgreater(397,537),(483,552)\textless /quad\textgreater\textless ref\textgreater五、舞台须安装大屏幕或大型设备时一定要固定牢固，以\textless /ref\textgreater\textless quad\textgreater(698,540),(898,554)\textless /quad\textgreater\textless ref\textgreater一、演出单位外加临时灯具设备技术指标要符合国\textless /ref\textgreater\textless quad\textgreater(115,555),(330,571)\textless /quad\textgreater\textless ref\textgreater(8)严禁在主舞台区域进行喷漆、上胶、上色等。如确实需要，应在\textless /ref\textgreater\textless quad\textgreater(385,554),(649,570)\textless /quad\textgreater\textless ref\textgreater免伤人或带来财产损失。\textless /ref\textgreater\textless quad\textgreater(716,555),(800,568)\textless /quad\textgreater\textless ref\textgreater家的安全生产标准，并了解安装位置及数量。\textless /ref\textgreater\textless quad\textgreater(134,575),(326,592)\textless /quad\textgreater\textless ref\textgreater剧院指定的施工地点进行作业。\textless /ref\textgreater\textless quad\textgreater(396,572),(519,587)\textless /quad\textgreater\textless ref\textgreater六、舞台下方全是电气设备，请您勿将水等液体饮料带上\textless /ref\textgreater\textless quad\textgreater(699,570),(901,584)\textless /quad\textgreater\textless ref\textgreater二、外加临时灯具如要接入剧院灯光系统，应试亮\textless /ref\textgreater\textless quad\textgreater(111,597),(329,613)\textless /quad\textgreater\textless ref\textgreater(9)保持舞台清洁，请勿乱扔废弃物品。\textless /ref\textgreater\textless quad\textgreater(385,591),(537,606)\textless /quad\textgreater\textless ref\textgreater舞台，以防引发电气事故。\textless /ref\textgreater\textless quad\textgreater(718,586),(811,598)\textless /quad\textgreater\textless ref\textgreater七、如团方自带音响设备，请你提供一路信号给剧场化妆\textless /ref\textgreater\textless quad\textgreater(702,600),(904,614)\textless /quad\textgreater\textless ref\textgreater后方可接入本系统，灯具单功率不得超过剧院\textless /ref\textgreater\textless quad\textgreater(130,619),(329,635)\textless /quad\textgreater\textless ref\textgreater四、演出时：\textless /ref\textgreater\textless quad\textgreater(375,611),(427,625)\textless /quad\textgreater\textless ref\textgreater间使用，以免耽误演员上场。\textless /ref\textgreater\textless quad\textgreater(720,616),(821,628)\textless /quad\textgreater\textless ref\textgreater标准（4kw），外加临时灯具总功率不得超过\textless /ref\textgreater\textless quad\textgreater(128,640),(329,658)\textless /quad\textgreater\textless ref\textgreater(1)切换场景时，应将景片、道具等抬离舞台地面后再移动，严禁在\textless /ref\textgreater\textless quad\textgreater(383,627),(652,644)\textless /quad\textgreater\textless ref\textgreater八、如团方需要外接电源，请团方技术人员将线缆铺设至\textless /ref\textgreater\textless quad\textgreater(703,631),(907,645)\textless /quad\textgreater\textless ref\textgreater剧院用电安全范围。\textless /ref\textgreater\textless quad\textgreater(127,665),(213,680)\textless /quad\textgreater\textless ref\textgreater舞台地板上拖拉，以免损坏划伤舞台地板。\textless /ref\textgreater\textless quad\textgreater(395,646),(567,661)\textless /quad\textgreater\textless ref\textgreater舞台两侧的配电柜，请团方提供用电功率，由剧场专\textless /ref\textgreater\textless quad\textgreater(721,648),(910,662)\textless /quad\textgreater\textless ref\textgreater(2)请勿将松香、饮料等粘性物质直接倒在舞台地板上，应在剧院舞\textless /ref\textgreater\textless quad\textgreater(382,666),(654,681)\textless /quad\textgreater\textless ref\textgreater业电工过来负责接线，负载需三相平衡，接完由团方\textless /ref\textgreater\textless quad\textgreater(723,664),(912,678)\textless /quad\textgreater\textless ref\textgreater三、外加临时灯具设备的挂钩、保险链承重应与之\textless /ref\textgreater\textless quad\textgreater(102,687),(326,704)\textless /quad\textgreater\textless ref\textgreater台技术人员的指导下将上述物质放在指定位置或容器中使用。\textless /ref\textgreater\textless quad\textgreater(394,684),(641,700)\textless /quad\textgreater\textless ref\textgreater技术人员确认无误、送电，调试、彩练、演出期间由\textless /ref\textgreater\textless quad\textgreater(724,679),(914,694)\textless /quad\textgreater\textless ref\textgreater相匹配。\textless /ref\textgreater\textless quad\textgreater(122,710),(158,726)\textless /quad\textgreater\textless ref\textgreater(3)烟机、干冰机等设备禁止在主舞台区域使用，以免造成漏油、漏\textless /ref\textgreater\textless quad\textgreater(380,703),(656,720)\textless /quad\textgreater\textless ref\textgreater团方技术人员负责检查配电柜开关以下部分（包括线\textless /ref\textgreater\textless quad\textgreater(725,696),(915,712)\textless /quad\textgreater\textless ref\textgreater九、剧院内全场禁烟，包含卫生间内。抽烟请至大剧院3号\textless /ref\textgreater\textless quad\textgreater(709,724),(922,739)\textless /quad\textgreater\textless ref\textgreater四、灯具灯线应完好无损，接插件连接紧密，接插\textless /ref\textgreater\textless quad\textgreater(98,732),(324,750)\textless /quad\textgreater\textless ref\textgreater门外。（烟头熄灭后请丢到垃圾桶内，谢谢合作）\textless /ref\textgreater\textless quad\textgreater(728,741),(909,756)\textless /quad\textgreater\textless ref\textgreater符合灯具要求。\textless /ref\textgreater\textless quad\textgreater(118,757),(185,773)\textless /quad\textgreater\textless ref\textgreater(4)大提琴等乐器使用时应使用防滑垫，钢琴等自重较大的轮式乐器\textless /ref\textgreater\textless quad\textgreater(379,742),(659,759)\textless /quad\textgreater\textless ref\textgreater十、演出道具请码放整齐、不得堵塞消防通道。\textless /ref\textgreater\textless quad\textgreater(711,759),(885,774)\textless /quad\textgreater\textless ref\textgreater五、外加临时灯具设备与幕布、布景之间距离不得\textless /ref\textgreater\textless quad\textgreater(94,779),(323,798)\textless /quad\textgreater\textless ref\textgreater进入主舞台区域时，应使用钢琴运输车等工具，乐器定位后应尽\textless /ref\textgreater\textless quad\textgreater(392,762),(660,779)\textless /quad\textgreater\textless ref\textgreater十一、剧场内禁止动用明火、电焊、彩虹机、手持礼花炮及\textless /ref\textgreater\textless quad\textgreater(712,776),(932,791)\textless /quad\textgreater\textless ref\textgreater小于50cm。\textless /ref\textgreater\textless quad\textgreater(114,806),(166,821)\textless /quad\textgreater\textless ref\textgreater量避免在主舞台区域地面移动。\textless /ref\textgreater\textless quad\textgreater(393,783),(522,799)\textless /quad\textgreater\textless ref\textgreater氧气、氢气、氮气设备。确需使用请联系保安保洁部。\textless /ref\textgreater\textless quad\textgreater(731,792),(930,807)\textless /quad\textgreater\textless ref\textgreater五、拆台时：\textless /ref\textgreater\textless quad\textgreater(369,806),(424,820)\textless /quad\textgreater\textless ref\textgreater到指定地点动用。\textless /ref\textgreater\textless quad\textgreater(731,809),(797,823)\textless /quad\textgreater\textless ref\textgreater六、所有线路走线排列整齐，不得使用铁丝、纸胶\textless /ref\textgreater\textless quad\textgreater(90,827),(321,847)\textless /quad\textgreater\textless ref\textgreater(1)应将景片、道具等抬离舞台地板后运至侧舞台区域进行拆卸，禁\textless /ref\textgreater\textless quad\textgreater(377,823),(662,841)\textless /quad\textgreater\textless ref\textgreater十二、所有演职人员凭《上海保利大剧院管理有限公司》的\textless /ref\textgreater\textless quad\textgreater(715,825),(937,840)\textless /quad\textgreater\textless ref\textgreater带等绑线。\textless /ref\textgreater\textless quad\textgreater(110,854),(157,870)\textless /quad\textgreater\textless ref\textgreater止在主舞台区域进行拆卸操作。\textless /ref\textgreater\textless quad\textgreater(391,844),(523,861)\textless /quad\textgreater\textless ref\textgreater工作证件进出。\textless /ref\textgreater\textless quad\textgreater(734,843),(790,857)\textless /quad\textgreater\textless ref\textgreater十三、食品、饮料不得带进剧场内。\textless /ref\textgreater\textless quad\textgreater(717,860),(851,875)\textless /quad\textgreater\textless ref\textgreater七、外加临时灯具设备装好后，必须核对数量、功\textless /ref\textgreater\textless quad\textgreater(86,873),(319,897)\textless /quad\textgreater\textless ref\textgreater(2)灯具等拆下后，应直接运至侧舞台进行装箱，禁止在主舞台区域\textless /ref\textgreater\textless quad\textgreater(376,864),(664,884)\textless /quad\textgreater\textless ref\textgreater十四、演职人员不得动用剧院内的消防设施设备。\textless /ref\textgreater\textless quad\textgreater(717,877),(903,892)\textless /quad\textgreater\textless ref\textgreater率并开灯检查总功率。\textless /ref\textgreater\textless quad\textgreater(106,902),(209,921)\textless /quad\textgreater\textless ref\textgreater(3)拆台结束后，请将舞台区域打扫干净，将废弃物清理干净。\textless /ref\textgreater\textless quad\textgreater(375,906),(641,928)\textless /quad\textgreater\textless ref\textgreater十五、团方如有问题需要帮助及时沟通及时给予解决。\textless /ref\textgreater\textless quad\textgreater(717,894),(922,910)\textless /quad\textgreater\textless ref\textgreater十六、以上制度事项请团方负责人认真阅读。\textless /ref\textgreater\textless quad\textgreater(719,913),(890,929)\textless /quad\textgreater\textless ref\textgreater(4)由于主办演出方使用不当等原因造成剧院舞台地板划伤损坏，将\textless /ref\textgreater\textless quad\textgreater(374,927),(666,949)\textless /quad\textgreater\textless ref\textgreater上海保利大剧院管理有限公司\textless /ref\textgreater\textless quad\textgreater(746,954),(931,979)\textless /quad\textgreater\textless ref\textgreater视损坏程度予以经济赔偿。\textless /ref\textgreater\textless quad\textgreater(388,952),(504,971)\textless /quad\textgreater\textless ref\textgreater二〇一五年一月一日\textless /ref\textgreater\textless quad\textgreater(789,988),(904,1000)\textless /quad\textgreater
}
    \end{minipage}
    \caption{Robust Text Spotting Results of HunyuanOCR on Dense Documents.}
    \label{fig:spotting12}
\end{figure}

\newpage
\renewcommand{\arraystretch}{1.5}
\begin{figure}[!h]
    \centering
    \begin{tabular}{@{}c@{\hspace{0.5cm}}c@{}}
        \begin{minipage}[b]{0.45\textwidth}
            \centering
            Input image \\
            \includegraphics[width=0.9\linewidth]{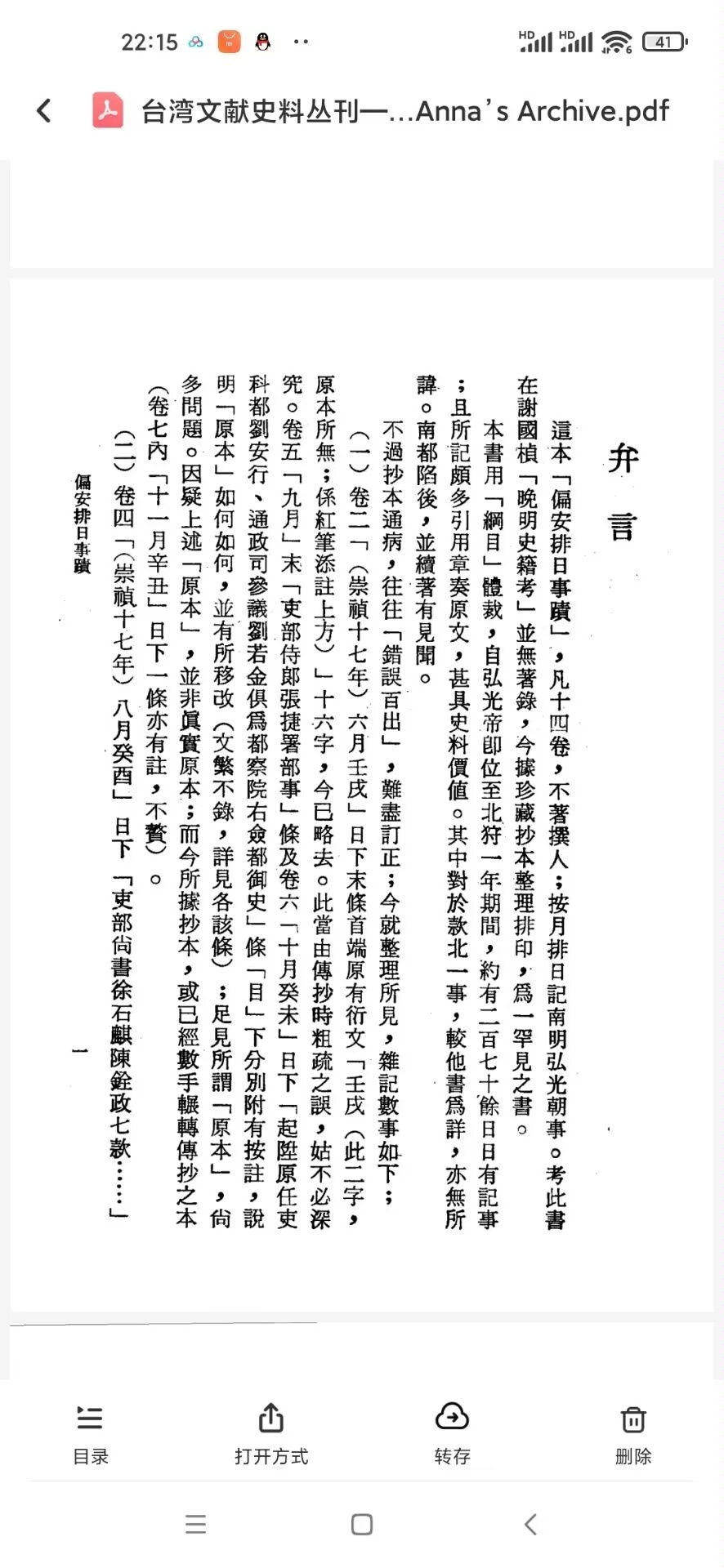}
        \end{minipage}
         &
        \begin{minipage}[b]{0.45\textwidth}
            \centering
            Visualization of HunyuanOCR output \\
            \includegraphics[width=0.9\linewidth]{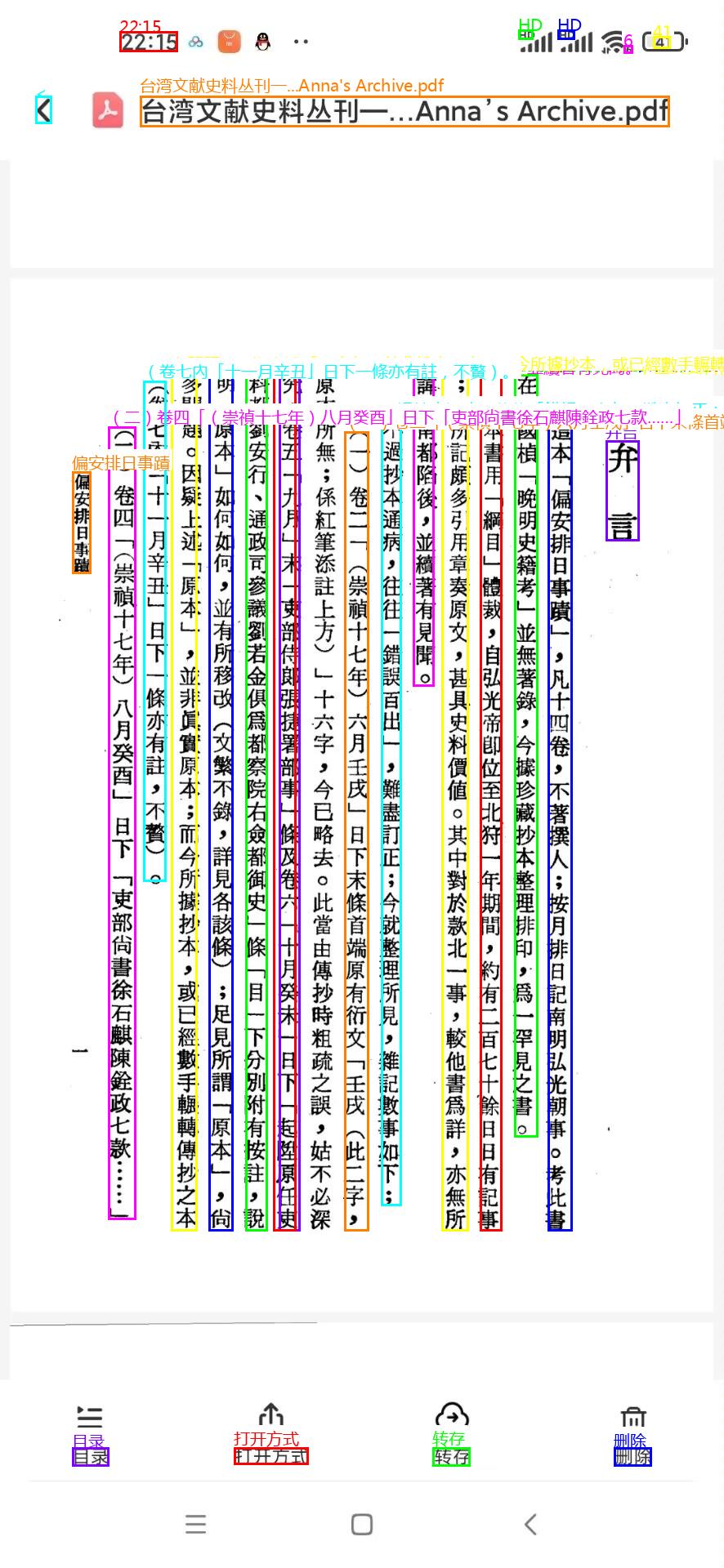}
        \end{minipage}
    \end{tabular}
    \vspace{1cm}
    \begin{minipage}{0.9\textwidth}
        \textbf{Prompt:} 检测并识别图片中的文字，将文本坐标格式化输出。
        \\
        \\
        \textbf{HunyuanOCR:}
        {
            \scriptsize
            \textless ref\textgreater 22:15\textless /ref \textgreater\textless quad \textgreater(165,20),(245,33)\textless /quad \textgreater \textless ref \textgreater HD\textless /ref \textgreater \textless quad \textgreater (716,19),(737,25)\textless /quad \textgreater \textless ref \textgreater HD\textless /ref \textgreater \textless quad \textgreater (770,19),(794,25)\textless /quad \textgreater \textless ref \textgreater 41\textless /ref \textgreater \textless quad \textgreater (902,23),(926,31)\textless /quad \textgreater \textless ref \textgreater 6\textless /ref \textgreater \textless quad \textgreater (862,29),(874,34)\textless /quad \textgreater \textless ref \textgreater \textless /ref \textgreater \textless quad \textgreater (49,61),(72,79)\textless /quad \textgreater \textless ref \textgreater 台湾文献史料丛刊一...Anna's Archive.pdf\textless /ref \textgreater \textless quad \textgreater (194,61),(925,81)\textless /quad \textgreater \textless ref \textgreater 弁言\textless /ref \textgreater \textless quad \textgreater (837,281),(883,345)\textless /quad \textgreater \textless ref \textgreater 本書用「綱目」體裁，自弘光帝即位至北狩一年期間，約有二百七十餘日日有記事\textless /ref \textgreater \textless quad \textgreater (663,239),(694,785)\textless /quad \textgreater \textless ref \textgreater 在謝國楨「晚明史籍考」並無著錄，今據珍藏抄本整理排印，爲一罕見之書。\textless /ref \textgreater \textless quad \textgreater (710,239),(743,725)\textless /quad \textgreater \textless ref \textgreater 這本「偏安排日事蹟」，凡十四卷，不著撰人；按月排日記南明弘光朝事。考此書\textless /ref \textgreater \textless quad \textgreater (757,268),(791,785)\textless /quad \textgreater \textless ref \textgreater ；且所記頗多引用章奏原文，甚具史料價值。其中對於款北一事，較他書爲詳，亦無所\textless /ref \textgreater \textless quad \textgreater (610,240),(647,785)\textless /quad \textgreater \textless ref \textgreater 諱。南都陷後，並續著有見聞。\textless /ref \textgreater \textless quad \textgreater (571,240),(600,438)\textless /quad \textgreater \textless ref \textgreater 不過抄本通病，往往「錯誤百出」，難盡訂正；今就整理所見，雜記數事如下；\textless /ref \textgreater \textless quad \textgreater (526,267),(555,769)\textless /quad \textgreater \textless ref \textgreater （一）卷二「（崇禎十七年）六月壬戌」日下末條首端原有衍文「壬戌（此二字，\textless /ref \textgreater \textless quad \textgreater (476,275),(510,785)\textless /quad \textgreater \textless ref \textgreater 原本所無；係紅筆添註上方）」十六字，今已略去。此當由傳抄時粗疏之誤，姑不必深\textless /ref \textgreater \textless quad \textgreater (385,238),(415,785)\textless /quad \textgreater \textless ref \textgreater 究。卷五「九月」末「吏部侍郎張捷署部事」條及卷六「十月癸未」日下「起陞原任吏 \textless /ref \textgreater \textless quad \textgreater (378,238),(409,785)\textless /quad \textgreater \textless ref \textgreater 科都劉安行、通政司參議劉若金俱爲都察院右僉都御史」條「日」下分别附有按註，說\textless /ref \textgreater \textless quad \textgreater (339,238),(370,785) \textless /quad \textgreater \textless ref \textgreater 明「原本」如何如何，並有所移改（文繁不錄，詳見各該條）；足見所謂「原本」，尙\textless /ref \textgreater \textless quad \textgreater (288,238),(322,785)\textless /quad \textgreater \textless ref \textgreater 多問題。因疑上述「原本」，並非眞實原本；而今所據抄本，或已經數手輾轉傳抄之本\textless /ref \textgreater \textless quad \textgreater (237,238),(273,785)\textless /quad \textgreater \textless ref \textgreater （二）卷四「（崇禎十七年）八月癸酉」日下「吏部尙書徐石麒陳銓政七款……」\textless /ref \textgreater \textless quad \textgreater (150,272),(188,778)\textless /quad \textgreater \textless ref \textgreater （卷七內「十一月辛丑」日下一條亦有註，不贅）。\textless /ref \textgreater \textless quad \textgreater (198,243),(230,562)\textless /quad \textgreater \textless ref \textgreater 偏安排日事蹟\textless /ref \textgreater \textless quad \textgreater (100,301),(126,366)\textless /quad \textgreater \textless ref \textgreater 目录\textless /ref \textgreater \textless quad \textgreater (100,923),(151,935)\textless /quad \textgreater \textless ref \textgreater 打开方式\textless /ref \textgreater \textless quad \textgreater (323,923),(426,934)\textless /quad \textgreater \textless ref \textgreater 转存\textless /ref \textgreater \textless quad \textgreater (598,923),(650,935)\textless /quad \textgreater \textless ref \textgreater 删除\textless /ref \textgreater \textless quad \textgreater (848,923),(900,935)\textless /quad \textgreater}
    \end{minipage}
    \vspace{-3em}
    \caption{Robust Text Spotting Performance of HunyuanOCR in Complex Document Scenarios.}
    \label{fig:spotting2}
\end{figure}

%% file: sec/appendix/parsing_fig.tex
\renewcommand{\arraystretch}{1.5}
\begin{figure}[!h]
    \centering
    \begin{tabular}{@{}c@{\hspace{0.5cm}}c@{}}
        \begin{minipage}{0.45\textwidth}
            \centering
            Input image \\
            \includegraphics[height=0.92\textwidth]{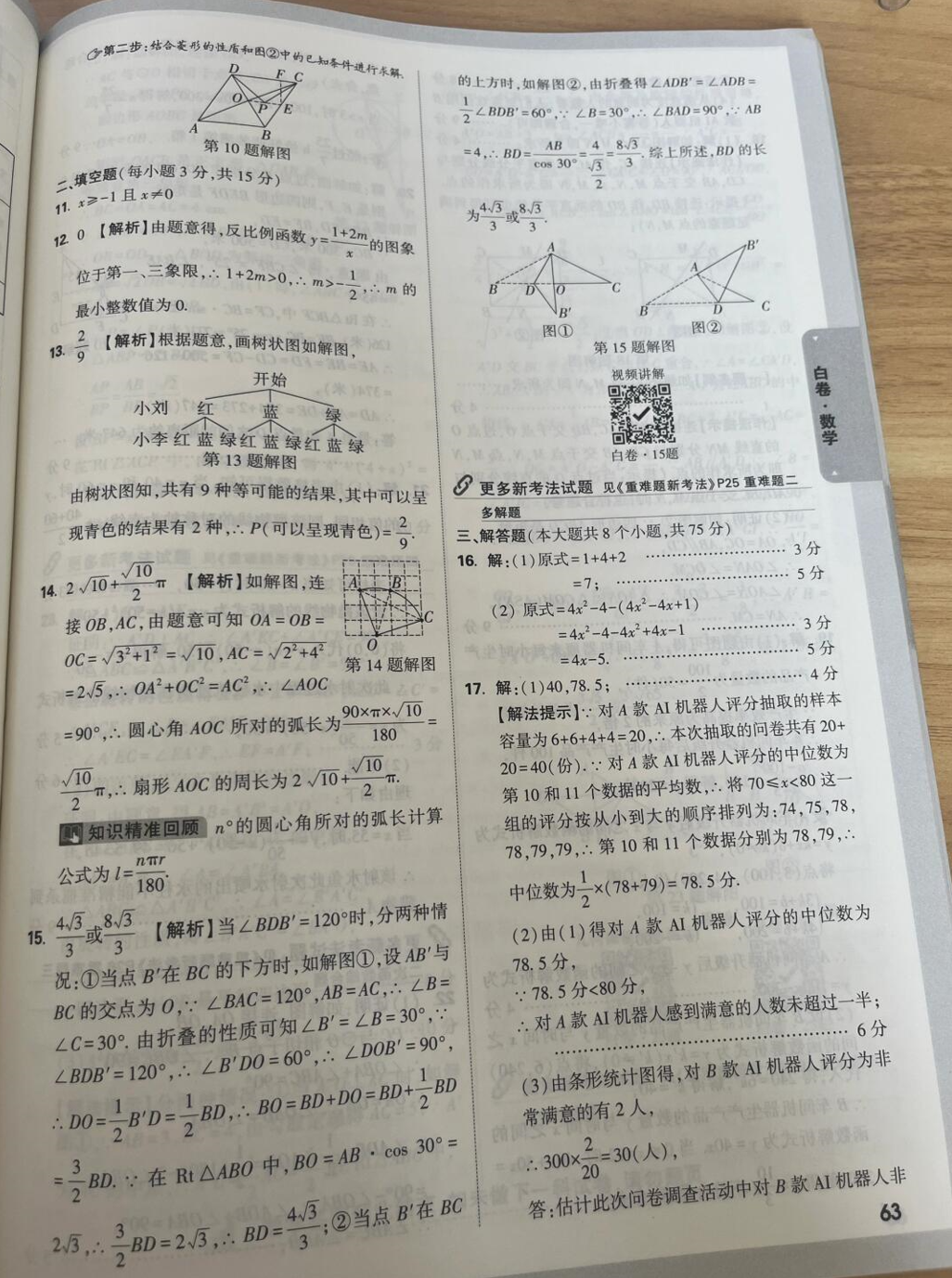}
        \end{minipage}
         &
        \begin{minipage}{0.45\textwidth}
            \centering
            Visualization of HunyuanOCR output \\
            \includegraphics[height=0.92\textwidth]{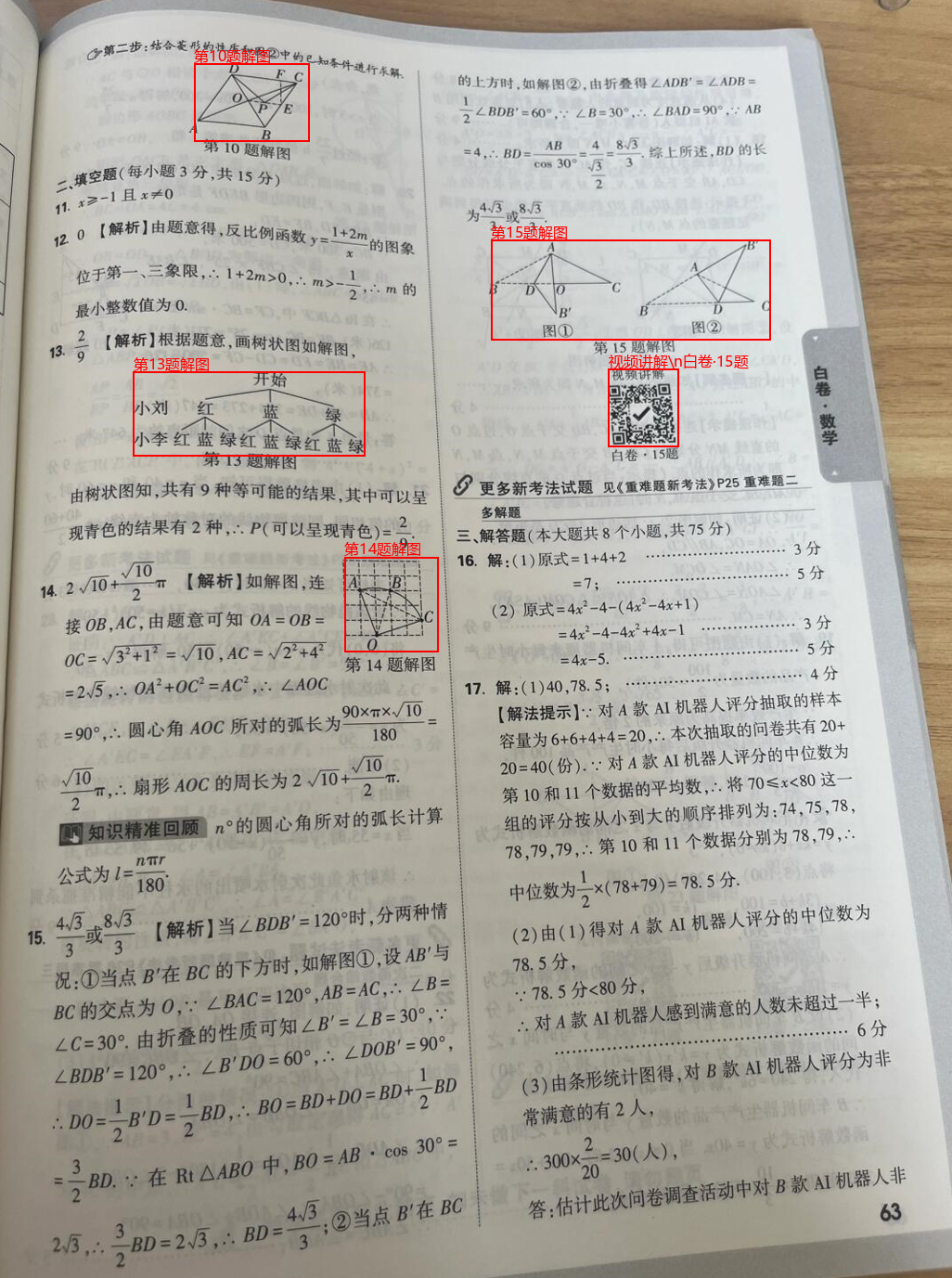}
        \end{minipage}
    \end{tabular}
    \vspace{1cm}
    \begin{minipage}{0.9\textwidth}
        \textbf{Prompt:} 提取文档图片中正文的所有信息用markdown格式表示，其中页眉、页脚部分忽略，表格用html格式表达，文档中公式用latex格式表示，按照阅读顺序组织进行解析。
        \\
        \\
        \textbf{HunyuanOCR:}
        {
        \scriptsize
        第二步:结合菱形的性质和图②中的已知条件进行求解. \\

        \textless pFig\textgreater 第10题解图\textless /pFig\textgreater \textless quad\textgreater (204,50),(325,111)\textless /quad\textgreater \\

        \#\# 二、填空题(每小题3分,共15分)

        11.$ x\geq - 1 $且$ x\ne 0 $

        12.0【解析】由题意得,反比例函数$ y=\frac{1+2m}{x} $的图象位于第一、三象限,$ \therefore 1+2m\textgreater 0,\therefore m\textgreater -\frac{1}{2},\therefore m $ 的最小整数值为0.

        13.$ \frac{2}{9} $ 【解析】根据题意,画树状图如解图,\\

        \textless pFig\textgreater 第13题解图\textless /pFig\textgreater \textless quad\textgreater (145,291),(388,357)\textless /quad\textgreater \\

        由树状图知,共有9种等可能的结果,其中可以呈现青色的结果有2种,$ \therefore P(\text{可以呈现青色})=\frac{2}{9} $.

        14.$ 2\sqrt{10}+\frac{\sqrt{10}}{2}\pi $ 【解析】如解图,连接OB,AC,由题意可知 $OA=OB=OC=\sqrt{3^{2}+1^{2}}=\sqrt{10},AC=\sqrt{2^{2}+4^{2}}=2\sqrt{5}$,
        $\therefore OA^{2}+OC^{2}=AC^{2}$,
        $\therefore\angle AOC=90^{\circ}$,
        $\therefore$ 圆心角AOC所对的弧长为 $\frac{90\times\pi\times\sqrt{10}}{180}=\frac{\sqrt{10}}{2}\pi$,
        $\therefore$扇形AOC的周长为$ 2\sqrt{10}+\frac{\sqrt{10}}{2}\pi $.\\

        \textless pFig\textgreater 第14题解图\textless /pFig\textgreater \textless quad\textgreater (362,436),(461,510)\textless /quad\textgreater \\

        知识精准回顾：$ n^{\circ} $的圆心角所对的弧长计算公式为 $l=\frac{n\pi r}{180} $.

        15.$ \frac{4\sqrt{3}}{3} $或$ \frac{8\sqrt{3}}{3}$ 【解析】当$\angle BDB^{\prime}=120^{\circ}$时,
        分两种情况：①当点B'在BC的下方时,如解图①,设AB'与BC的交点为O,$ \because\angle BAC = 120^{\circ},AB = AC,\therefore\angle B=\angle C=30^{\circ} $.
        由折叠的性质可知$ \angle B^{\prime}=\angle B=30^{\circ},\because\angle BDB^{\prime}=120^{\circ},\therefore\angle B^{\prime}DO=60^{\circ},\therefore\angle DOB^{\prime}=90^{\circ},\therefore DO=\frac{1}{2}B^{\prime}D=\frac{1}{2}BD,\therefore BO=BD+DO=BD+\frac{1}{2}BD=\frac{3}{2}BD.\because $
        在$ Rt\triangle ABO $中,$ BO=AB\cdot\cos30^{\circ}=2\sqrt{3},\therefore\frac{3}{2}BD=2\sqrt{3},\therefore BD=\frac{4\sqrt{3}}{3} $；
        ②当点B'在BC的上方时,如解图②,由折叠得$ \angle ADB^{\prime}=\angle ADB=\frac{1}{2}\angle BDB^{\prime} = 60^{\circ},\because\angle B=30^{\circ},\therefore\angle BAD=90^{\circ},\because AB = 4,\therefore BD=\frac{AB}{\cos30^{\circ}}=\frac{4}{\frac{\sqrt{3}}{2}}=\frac{8\sqrt{3}}{3} $.

                    综上所述,BD的长为$ \frac{4\sqrt{3}}{3} $或$ \frac{8\sqrt{3}}{3} $. \\

                    \textless pFig\textgreater 第15题解图\textless /pFig\textgreater \textless quad\textgreater (516,188),(809,266)\textless /quad\textgreater \\

                    \textless pFig\textgreater 视频讲解\textbackslash n白卷·15题\textless /pFig\textgreater \textless quad\textgreater (639,289),(713,350)\textless /quad\textgreater\\

                $ \underline{\text{更多新考法实体 见《重难题新考法》P25重难题二}} $

                $ \underline{\text{多解题}} $

                    \#\# 三、解答题（本大题共8个小题,共75分）

                    16.解：(1) 原式=1+4+2·····3分
                    =7;·····5分

                    (2) 原式=$ 4x^{2}-4-(4x^{2}-4x+1) $
                $ =4x^{2}-4-4x^{2}+4x-1 $·····3分
                $ =4x-5 $.·····5分

                    17.解：(1)40,78.5;·····4分

                    【解法提示】 $ \because $对A款AI机器人评分抽取的样本容量为6+6+4+4=20,$ \therefore $本次抽取的问卷共有20+20=40(份).$ \because $对A款AI机器人评分的中位数为第10和11个数据的平均数,$ \therefore $将$ 70\leq x \textless  80 $这一组的评分按从小到大的顺序排列为:74,75,78,78,79,79,$ \therefore $第10和11个数据分别为78,79,$ \therefore $中位数为$ \frac{1}{2}\times(78 + 79)=78.5 $分.

                    (2) 由(1)得对A款AI机器人评分的中位数为78.5分,
                $ \because $78.5分\textless  80分,
                $ \therefore $对A款AI机器人感到满意的人数未超过一半;·····6分

                    (3) 由条形统计图得,对B款AI机器人评分为非常满意的有2人,
                $ \therefore300\times\frac{2}{20}=30 $ (人) ,

                答：估计此次问卷调查活动中对B款AI机器人非}  \\
    \end{minipage}
    \vspace{-0.5em}
    \caption{Robust Parsing Performance of HunyuanOCR in Complex Figure Scenarios.}
    \label{fig:spotting_table}
\end{figure}

%% file: sec/appendix/parsing_table.tex
\renewcommand{\arraystretch}{1.5}
\begin{figure}[!h]
    \centering
    \begin{tabular}{@{}c@{\hspace{0.5cm}}c@{}}
        \begin{minipage}{0.45\textwidth}
            \centering
            Input image \\
            \includegraphics[height=0.92\textwidth]{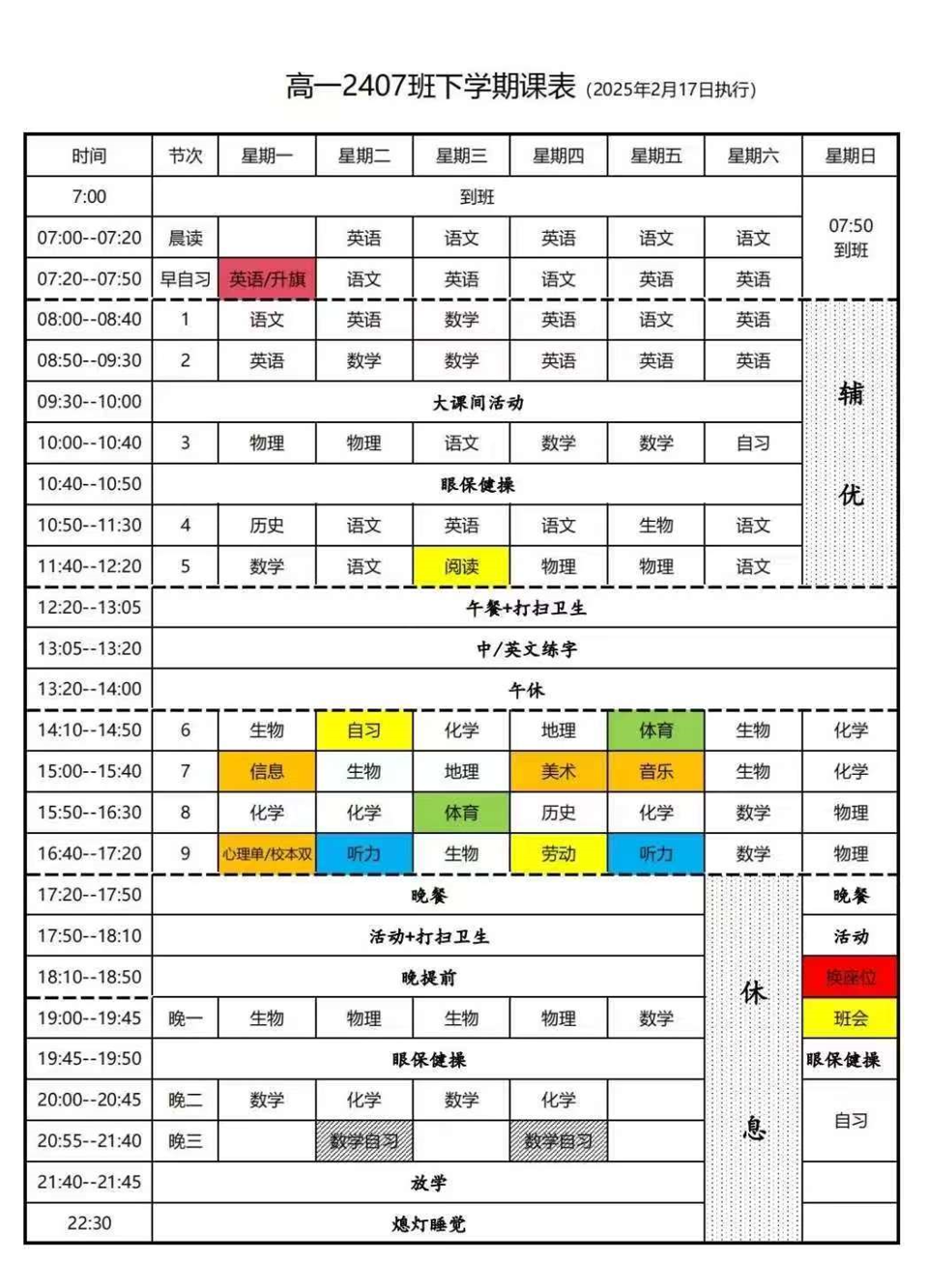}
        \end{minipage}
         &
        \begin{minipage}{0.45\textwidth}
            \centering
            Visualization of HunyuanOCR output \\
            \includegraphics[height=0.92\textwidth]{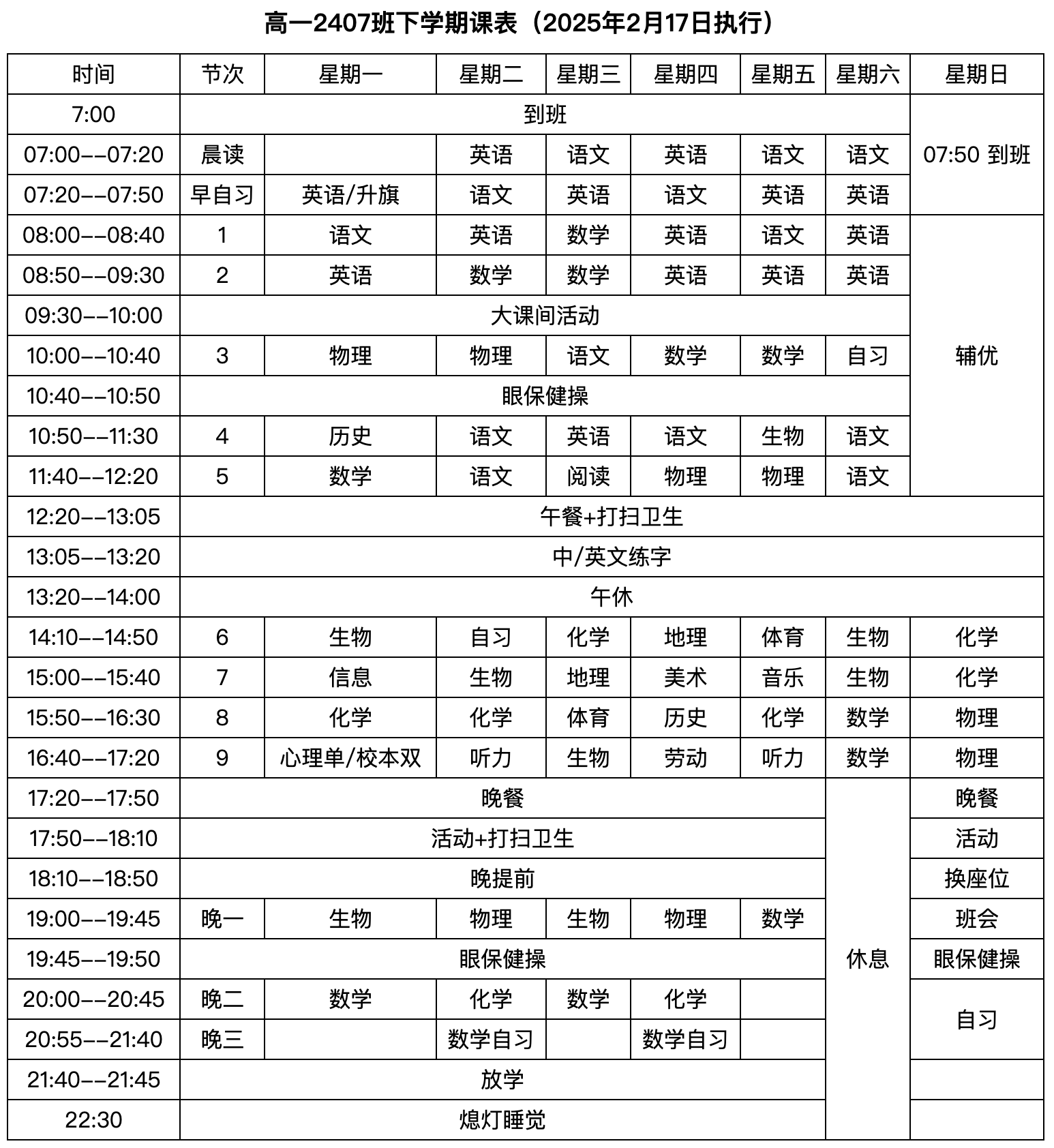}
        \end{minipage}
    \end{tabular}
    \vspace{0.5cm}
    \begin{minipage}{0.9\textwidth}
        \textbf{Prompt:} 提取文档图片中正文的所有信息用markdown格式表示，其中页眉、页脚部分忽略，表格用html格式表达，文档中公式用latex格式表示，按照阅读顺序组织进行解析。\\
        \\
        \textbf{HunyuanOCR:}
        {
            \scriptsize
            \textless table\textgreater \textless caption\textgreater 高一2407班下学期课表（2025年2月17日执行）\textless /caption\textgreater \textless tr\textgreater \textless td\textgreater 时间\textless /td\textgreater \textless td\textgreater 节次\textless /td\textgreater \textless td\textgreater 星期一\textless /td\textgreater \textless td\textgreater 星期二\textless /td\textgreater \textless td\textgreater 星期三\textless /td\textgreater \textless td\textgreater 星期四\textless /td\textgreater \textless td\textgreater 星期五\textless /td\textgreater \textless td\textgreater 星期六\textless /td\textgreater \textless td\textgreater 星期日\textless /td\textgreater \textless /tr\textgreater \textless tr\textgreater \textless td\textgreater 7:00\textless /td\textgreater \textless td colspan="7"\textgreater 到班\textless /td\textgreater \textless td rowspan="3"\textgreater 07:50 到班\textless /td\textgreater \textless /tr\textgreater \textless tr\textgreater \textless td\textgreater 07:00--07:20\textless /td\textgreater \textless td\textgreater 晨读\textless /td\textgreater \textless td\textgreater \textless /td\textgreater \textless td\textgreater 英语\textless /td\textgreater \textless td\textgreater 语文\textless /td\textgreater \textless td\textgreater 英语\textless /td\textgreater \textless td\textgreater 语文\textless /td\textgreater \textless td\textgreater 语文\textless /td\textgreater \textless /tr\textgreater \textless tr\textgreater \textless td\textgreater 07:20--07:50\textless /td\textgreater \textless td\textgreater 早自习\textless /td\textgreater \textless td\textgreater 英语/升旗\textless /td\textgreater \textless td\textgreater 语文\textless /td\textgreater \textless td\textgreater 英语\textless /td\textgreater \textless td\textgreater 语文\textless /td\textgreater \textless td\textgreater 英语\textless /td\textgreater \textless td\textgreater 英语\textless /td\textgreater \textless /tr\textgreater \textless tr\textgreater \textless td\textgreater 08:00--08:40\textless /td\textgreater \textless td\textgreater 1\textless /td\textgreater \textless td\textgreater 语文\textless /td\textgreater \textless td\textgreater 英语\textless /td\textgreater \textless td\textgreater 数学\textless /td\textgreater \textless td\textgreater 英语\textless /td\textgreater \textless td\textgreater 语文\textless /td\textgreater \textless td\textgreater 英语\textless /td\textgreater \textless td rowspan="7"\textgreater 辅优\textless /td\textgreater \textless /tr\textgreater \textless tr\textgreater \textless td\textgreater 08:50--09:30\textless /td\textgreater \textless td\textgreater 2\textless /td\textgreater \textless td\textgreater 英语\textless /td\textgreater \textless td\textgreater 数学\textless /td\textgreater \textless td\textgreater 数学\textless /td\textgreater \textless td\textgreater 英语\textless /td\textgreater \textless td\textgreater 英语\textless /td\textgreater \textless td\textgreater 英语\textless /td\textgreater \textless /tr\textgreater \textless tr\textgreater \textless td\textgreater 09:30--10:00\textless /td\textgreater \textless td colspan="7"\textgreater 大课间活动\textless /td\textgreater \textless /tr\textgreater \textless tr\textgreater \textless td\textgreater 10:00--10:40\textless /td\textgreater \textless td\textgreater 3\textless /td\textgreater \textless td\textgreater 物理\textless /td\textgreater \textless td\textgreater 物理\textless /td\textgreater \textless td\textgreater 语文\textless /td\textgreater \textless td\textgreater 数学\textless /td\textgreater \textless td\textgreater 数学\textless /td\textgreater \textless td\textgreater 自习\textless /td\textgreater \textless /tr\textgreater \textless tr\textgreater \textless td\textgreater 10:40--10:50\textless /td\textgreater \textless td colspan="7"\textgreater 眼保健操\textless /td\textgreater \textless /tr\textgreater \textless tr\textgreater \textless td\textgreater 10:50--11:30\textless /td\textgreater \textless td\textgreater 4\textless /td\textgreater \textless td\textgreater 历史\textless /td\textgreater \textless td\textgreater 语文\textless /td\textgreater \textless td\textgreater 英语\textless /td\textgreater \textless td\textgreater 语文\textless /td\textgreater \textless td\textgreater 生物\textless /td\textgreater \textless td\textgreater 语文\textless /td\textgreater \textless /tr\textgreater \textless tr\textgreater \textless td\textgreater 11:40--12:20\textless /td\textgreater \textless td\textgreater 5\textless /td\textgreater \textless td\textgreater 数学\textless /td\textgreater \textless td\textgreater 语文\textless /td\textgreater \textless td\textgreater 阅读\textless /td\textgreater \textless td\textgreater 物理\textless /td\textgreater \textless td\textgreater 物理\textless /td\textgreater \textless td\textgreater 语文\textless /td\textgreater \textless /tr\textgreater \textless tr\textgreater \textless td\textgreater 12:20--13:05\textless /td\textgreater \textless td colspan="8"\textgreater 午餐+打扫卫生\textless /td\textgreater \textless /tr\textgreater \textless tr\textgreater \textless td\textgreater 13:05--13:20\textless /td\textgreater \textless td colspan="8"\textgreater 中/英文练字\textless /td\textgreater \textless /tr\textgreater \textless tr\textgreater \textless td\textgreater 13:20--14:00\textless /td\textgreater \textless td colspan="8"\textgreater 午休\textless /td\textgreater \textless /tr\textgreater \textless tr\textgreater \textless td\textgreater 14:10--14:50\textless /td\textgreater \textless td\textgreater 6\textless /td\textgreater \textless td\textgreater 生物\textless /td\textgreater \textless td\textgreater 自习\textless /td\textgreater \textless td\textgreater 化学\textless /td\textgreater \textless td\textgreater 地理\textless /td\textgreater \textless td\textgreater 体育\textless /td\textgreater \textless td\textgreater 生物\textless /td\textgreater \textless td\textgreater 化学\textless /td\textgreater \textless /tr\textgreater \textless tr\textgreater \textless td\textgreater 15:00--15:40\textless /td\textgreater \textless td\textgreater 7\textless /td\textgreater \textless td\textgreater 信息\textless /td\textgreater \textless td\textgreater 生物\textless /td\textgreater \textless td\textgreater 地理\textless /td\textgreater \textless td\textgreater 美术\textless /td\textgreater \textless td\textgreater 音乐\textless /td\textgreater \textless td\textgreater 生物\textless /td\textgreater \textless td\textgreater 化学\textless /td\textgreater \textless /tr\textgreater \textless tr\textgreater \textless td\textgreater 15:50--16:30\textless /td\textgreater \textless td\textgreater 8\textless /td\textgreater \textless td\textgreater 化学\textless /td\textgreater \textless td\textgreater 化学\textless /td\textgreater \textless td\textgreater 体育\textless /td\textgreater \textless td\textgreater 历史\textless /td\textgreater \textless td\textgreater 化学\textless /td\textgreater \textless td\textgreater 数学\textless /td\textgreater \textless td\textgreater 物理\textless /td\textgreater \textless /tr\textgreater \textless tr\textgreater \textless td\textgreater 16:40--17:20\textless /td\textgreater \textless td\textgreater 9\textless /td\textgreater \textless td\textgreater 心理单/校本双\textless /td\textgreater \textless td\textgreater 听力\textless /td\textgreater \textless td\textgreater 生物\textless /td\textgreater \textless td\textgreater 劳动\textless /td\textgreater \textless td\textgreater 听力\textless /td\textgreater \textless td\textgreater 数学\textless /td\textgreater \textless td\textgreater 物理\textless /td\textgreater \textless /tr\textgreater \textless tr\textgreater \textless td\textgreater 17:20--17:50\textless /td\textgreater \textless td colspan="6"\textgreater 晚餐\textless /td\textgreater \textless td rowspan="9"\textgreater 休息\textless /td\textgreater \textless td\textgreater 晚餐\textless /td\textgreater \textless /tr\textgreater \textless tr\textgreater \textless td\textgreater 17:50--18:10\textless /td\textgreater \textless td colspan="6"\textgreater 活动+打扫卫生\textless /td\textgreater \textless td\textgreater 活动\textless /td\textgreater \textless /tr\textgreater \textless tr\textgreater \textless td\textgreater 18:10--18:50\textless /td\textgreater \textless td colspan="6"\textgreater 晚提前\textless /td\textgreater \textless td\textgreater 换座位\textless /td\textgreater \textless /tr\textgreater \textless tr\textgreater \textless td\textgreater 19:00--19:45\textless /td\textgreater \textless td\textgreater 晚一\textless /td\textgreater \textless td\textgreater 生物\textless /td\textgreater \textless td\textgreater 物理\textless /td\textgreater \textless td\textgreater 生物\textless /td\textgreater \textless td\textgreater 物理\textless /td\textgreater \textless td\textgreater 数学\textless /td\textgreater \textless td\textgreater 班会\textless /td\textgreater \textless /tr\textgreater \textless tr\textgreater \textless td\textgreater 19:45--19:50\textless /td\textgreater \textless td colspan="6"\textgreater 眼保健操\textless /td\textgreater \textless td\textgreater 眼保健操\textless /td\textgreater \textless /tr\textgreater \textless tr\textgreater \textless td\textgreater 20:00--20:45\textless /td\textgreater \textless td\textgreater 晚二\textless /td\textgreater \textless td\textgreater 数学\textless /td\textgreater \textless td\textgreater 化学\textless /td\textgreater \textless td\textgreater 数学\textless /td\textgreater \textless td\textgreater 化学\textless /td\textgreater \textless td\textgreater \textless /td\textgreater \textless td rowspan="2"\textgreater 自习\textless /td\textgreater \textless /tr\textgreater \textless tr\textgreater \textless td\textgreater 20:55--21:40\textless /td\textgreater \textless td\textgreater 晚三\textless /td\textgreater \textless td\textgreater \textless /td\textgreater \textless td\textgreater 数学自习\textless /td\textgreater \textless td\textgreater \textless /td\textgreater \textless td\textgreater 数学自习\textless /td\textgreater \textless td\textgreater \textless /td\textgreater \textless /tr\textgreater \textless tr\textgreater \textless td\textgreater 21:40--21:45\textless /td\textgreater \textless td colspan="6"\textgreater 放学\textless /td\textgreater \textless td\textgreater \textless /td\textgreater \textless /tr\textgreater \textless tr\textgreater \textless td\textgreater 22:30\textless /td\textgreater \textless td colspan="6"\textgreater 熄灯睡觉\textless /td\textgreater \textless td\textgreater \textless /td\textgreater \textless /tr\textgreater \textless /table\textgreater
        }  \\
    \end{minipage}

    \caption{Robust Parsing Performance of HunyuanOCR in Complex Table Scenarios.}
    \label{fig:spotting_table_3}
\end{figure}

\begin{figure}[!h]
    \centering
    \begin{tabular}{@{}c@{\hspace{0.5cm}}c@{}}
        \begin{minipage}{0.45\textwidth}
            \centering
            Input image \\
            \includegraphics[height=0.92\textwidth]{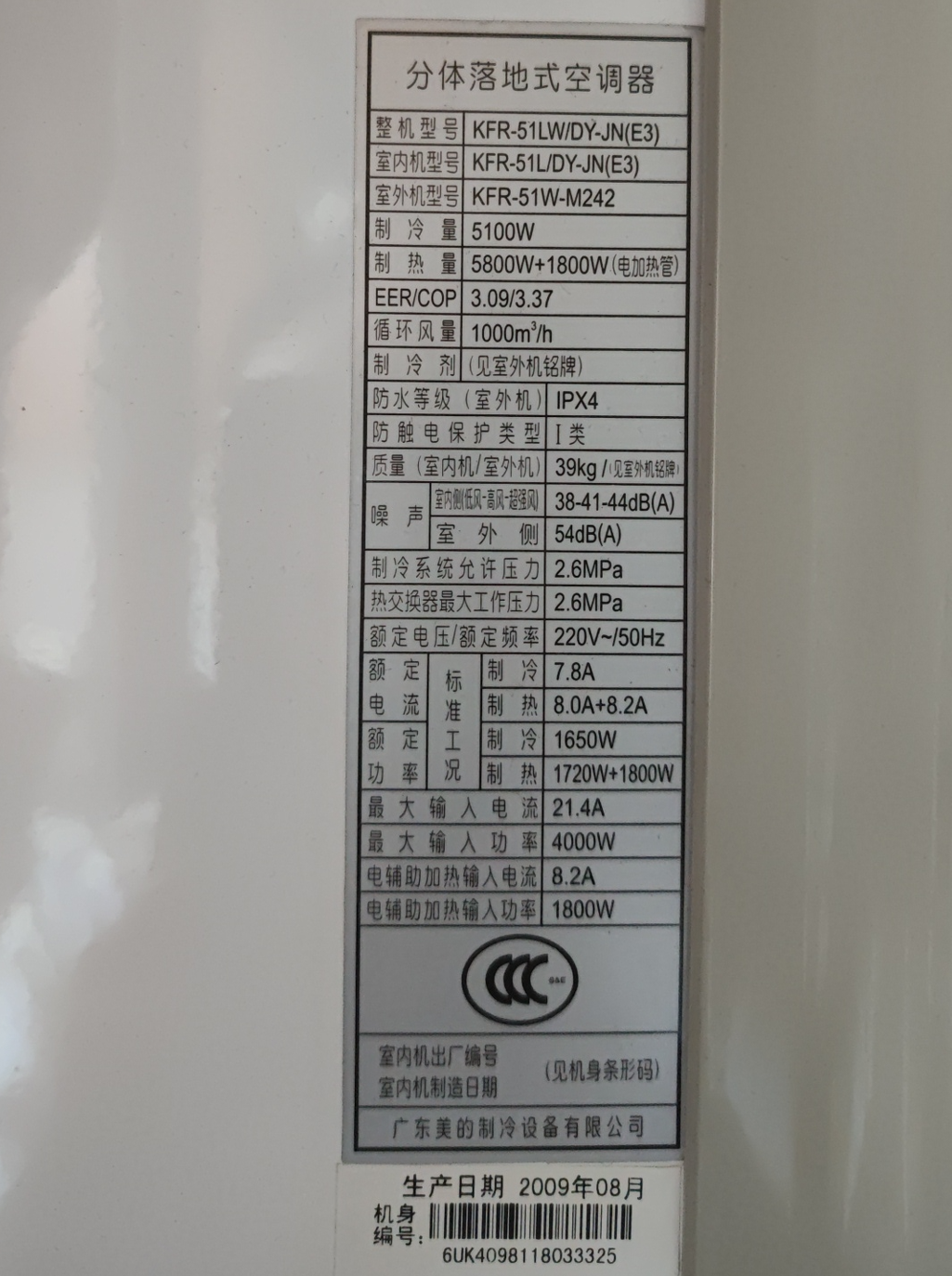}
        \end{minipage}
         &
        \begin{minipage}{0.45\textwidth}
            \centering
            Visualization of HunyuanOCR output \\
            \includegraphics[height=0.92\textwidth]{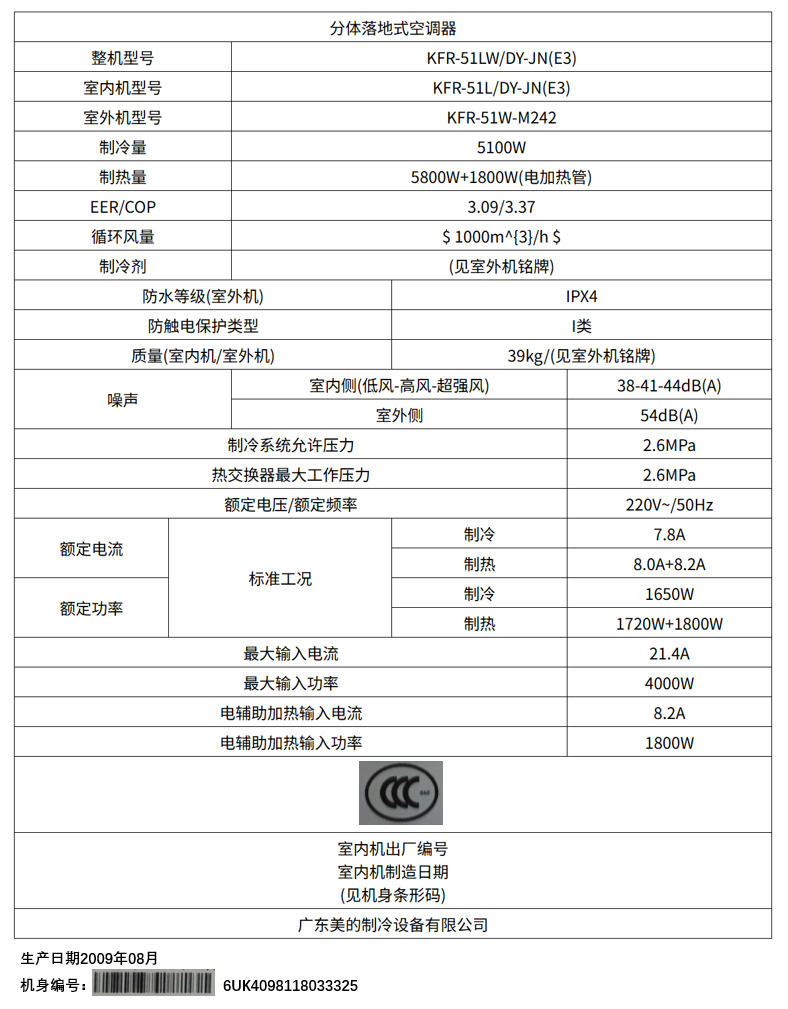}
        \end{minipage}
    \end{tabular}

    \vspace{0.5cm}
    \begin{minipage}{0.9\textwidth}
        \textbf{Prompt:} 提取文档图片中正文的所有信息用markdown格式表示，其中页眉、页脚部分忽略，表格用html格式表达，文档中公式用latex格式表示，按照阅读顺序组织进行解析。\\
        \\
        \textbf{HunyuanOCR:}
        {
            \scriptsize
            \textless table\textgreater \textless caption\textgreater \textless /caption\textgreater \textless tr\textgreater \textless td colspan="5"\textgreater 分体落地式空调器\textless /td\textgreater \textless /tr\textgreater \textless tr\textgreater \textless td  colspan="2"\textgreater 整机型号\textless /td\textgreater \textless td  colspan="3"\textgreater KFR-51LW/DY-JN(E3)\textless /td\textgreater \textless /tr\textgreater \textless tr\textgreater \textless td  colspan="2"\textgreater 室内机型号\textless /td\textgreater \textless td  colspan="3"\textgreater KFR-51L/DY-JN(E3)\textless /td\textgreater \textless /tr\textgreater \textless tr\textgreater \textless td  colspan="2"\textgreater 室外机型号\textless /td\textgreater \textless td  colspan="3"\textgreater KFR-51W-M242\textless /td\textgreater \textless /tr\textgreater \textless tr\textgreater \textless td  colspan="2"\textgreater 制冷量\textless /td\textgreater \textless td  colspan="3"\textgreater 5100W\textless /td\textgreater \textless /tr\textgreater \textless tr\textgreater \textless td  colspan="2"\textgreater 制热量\textless /td\textgreater \textless td  colspan="3"\textgreater 5800W+1800W(电加热管)\textless /td\textgreater \textless /tr\textgreater \textless tr\textgreater \textless td  colspan="2"\textgreater EER/COP\textless /td\textgreater \textless td  colspan="3"\textgreater 3.09/3.37\textless /td\textgreater \textless /tr\textgreater \textless tr\textgreater \textless td  colspan="2"\textgreater 循环风量\textless /td\textgreater \textless td  colspan="3"\textgreater $ 1000m^{3}/h $\textless /td\textgreater \textless /tr\textgreater \textless tr\textgreater \textless td  colspan="2"\textgreater 制冷剂\textless /td\textgreater \textless td  colspan="3"\textgreater (见室外机铭牌)\textless /td\textgreater \textless /tr\textgreater \textless tr\textgreater \textless td  colspan="3"\textgreater 防水等级(室外机)\textless /td\textgreater \textless td  colspan="2"\textgreater IPX4\textless /td\textgreater \textless /tr\textgreater \textless tr\textgreater \textless td  colspan="3"\textgreater 防触电保护类型\textless /td\textgreater \textless td  colspan="2"\textgreater I类\textless /td\textgreater \textless /tr\textgreater \textless tr\textgreater \textless td  colspan="3"\textgreater 质量(室内机/室外机)\textless /td\textgreater \textless td  colspan="2"\textgreater 39kg/(见室外机铭牌)\textless /td\textgreater \textless /tr\textgreater \textless tr\textgreater \textless td rowspan="2" colspan="2"\textgreater 噪声\textless /td\textgreater \textless td  colspan="2"\textgreater 室内侧(低风-高风-超强风) \textless /td\textgreater \textless td\textgreater 38-41-44dB(A)\textless /td\textgreater \textless /tr\textgreater \textless tr\textgreater \textless td  colspan="2"\textgreater 室外侧\textless /td\textgreater \textless td\textgreater 54dB(A)\textless /td\textgreater \textless /tr\textgreater \textless tr\textgreater \textless td  colspan="4"\textgreater 制冷系统允许压力\textless /td\textgreater \textless td\textgreater  2.6MPa\textless /td\textgreater \textless /tr\textgreater \textless tr\textgreater \textless td  colspan="4"\textgreater 热交换器最大工作压力\textless /td\textgreater \textless td\textgreater  2.6MPa\textless /td\textgreater \textless /tr\textgreater \textless tr\textgreater \textless td  colspan="4"\textgreater 额定电压/额定频率\textless /td\textgreater \textless td\textgreater  220V~/50Hz\textless /td\textgreater \textless /tr\textgreater \textless tr\textgreater \textless td rowspan="2"\textgreater 额定电流\textless /td\textgreater \textless td rowspan="4" colspan="2"\textgreater 标准工况\textless /td\textgreater \textless td\textgreater 制冷\textless /td\textgreater \textless td\textgreater 7.8A\textless /td\textgreater \textless /tr\textgreater \textless tr\textgreater \textless td\textgreater 制热\textless /td\textgreater \textless td\textgreater 8.0A+8.2A\textless /td\textgreater \textless /tr\textgreater \textless tr\textgreater \textless td rowspan="2"\textgreater 额定功率\textless /td\textgreater \textless td\textgreater 制冷\textless /td\textgreater \textless td\textgreater 1650W\textless /td\textgreater \textless /tr\textgreater \textless tr\textgreater \textless td\textgreater 制热\textless /td\textgreater \textless td\textgreater 1720W+1800W\textless /td\textgreater \textless /tr\textgreater \textless tr\textgreater \textless td  colspan="4"\textgreater 最大输入电流\textless /td\textgreater \textless td\textgreater 21.4A\textless /td\textgreater \textless /tr\textgreater \textless tr\textgreater \textless td  colspan="4"\textgreater 最大输入功率\textless /td\textgreater \textless td\textgreater 4000W\textless /td\textgreater \textless /tr\textgreater \textless tr\textgreater \textless td  colspan="4"\textgreater 电辅助加热输入电流\textless /td\textgreater \textless td\textgreater 8.2A\textless /td\textgreater \textless /tr\textgreater \textless tr\textgreater \textless td  colspan="4"\textgreater 电辅助加热输入功率\textless /td\textgreater \textless td\textgreater 1800W\textless /td\textgreater \textless /tr\textgreater \textless tr\textgreater \textless td  colspan="5"\textgreater \textless pFig\textgreater 图1\textless /pFig\textgreater \textless quad\textgreater (483,728),(608,800)\textless /quad\textgreater \textless /td\textgreater \textless /tr\textgreater \textless tr\textgreater \textless td  colspan="5"\textgreater 室内机出厂编号\textless br\textgreater 室内机制造日期\textless br\textgreater (见机身条形码)\textless br\textgreater \textless /td\textgreater \textless /tr\textgreater \textless tr\textgreater \textless td  colspan="5"\textgreater 广东美的制冷设备有限公司\textless /td\textgreater \textless /tr\textgreater \textless /table\textgreater\\

            生产日期 2009年08月\\

            机身编号: \textless pFig\textgreater 图2\textless /pFig\textgreater \textless quad\textgreater (452,938),(664,966)\textless /quad\textgreater 6UK4098118033325
        }
        \\
    \end{minipage}
    \caption{Robust Parsing Performance of HunyuanOCR in Complex Table Scenarios.}
    \label{fig:spotting_table_2}
\end{figure}

\renewcommand{\arraystretch}{1.5}
\begin{figure}[h]
    \centering
    \begin{tabular}{m{5cm}m{10cm}}
        \toprule
                
        \includegraphics[width=4.8cm]{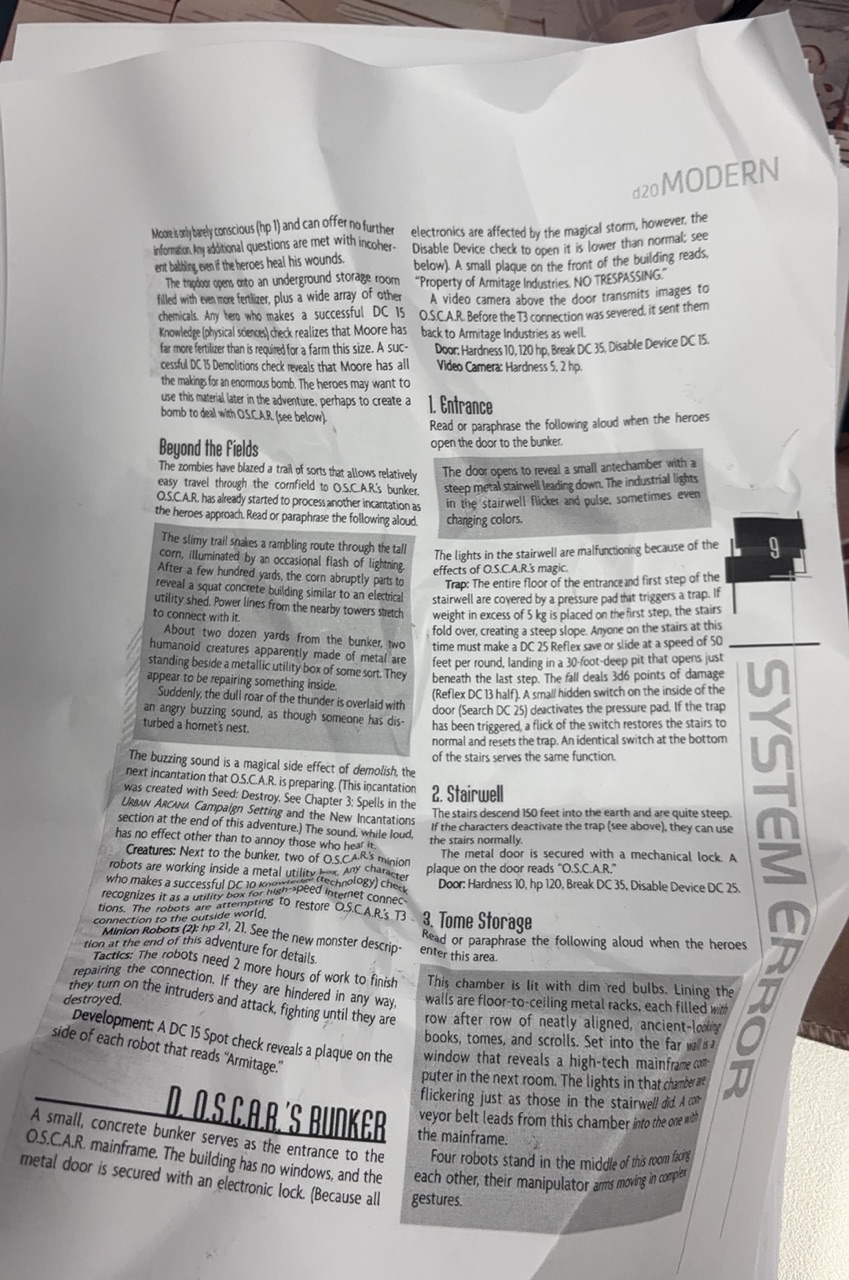} &
        {
        \textbf{Prompt:} {\footnotesize 提取文档图片中正文的所有信息用markdown格式表示，其中页眉、页脚部分忽略，表格用html格式表达，文档中公式用latex格式表示，按照阅读顺序组织进行解析。} \par
        \vspace{0.5em}
        \textbf{HunyuanOCR:}
         {\scriptsize
        ... 
        
Moore is only barely conscious (hp 1) and can offer no further information. Any additional questions are met with incoherent babbling, even if the heroes heal his wounds.

The trapdoor opens onto an underground storage room filled with even more fertilizer, plus a wide array of other chemicals. Any hero who makes a successful DC 15 Knowledge (physical sciences) check realizes that Moore has far more fertilizer than is required for a farm this size. A successful DC 15 Demolitions check reveals that Moore has all the makings for an enormous bomb. The heroes may want to use this material later in the adventure, perhaps to create a bomb to deal with O.S.C.A.R. (see below).

Beyond the Fields

The zombies have blazed a trail of sorts that allows relatively easy travel through the cornfield to O.S.C.A.R.’s bunker. O.S.C.A.R. has already started to process another incantation as the heroes approach. Read or paraphrase the following aloud.

The slimy trail snakes a rambling route through the tall corn, illuminated by an occasional flash of lightning. After a few hundred yards, the corn abruptly parts to reveal a squat concrete building similar to an electrical utility shed. Power lines from the nearby towers stretch to connect with it.

About two dozen yards from the bunker, two humanoid creatures apparently made of metal are standing beside a metallic utility box of some sort. They appear to be repairing something inside.

Suddenly, the dull roar of the thunder is overlaid with an angry buzzing sound, as though someone has disturbed a hornet’s nest.

The buzzing sound is a magical side effect of demolish, the next incantation that O.S.C.A.R. is preparing. (This incantation was created with Seed: Destroy. See Chapter 3: Spells in the Urban Arcan Campaign Setting and the New Incantations section at the end of this adventure.) The sound, while loud, has no effect other than to annoy those who hear it.

Creatures: Next to the bunker, two of O.S.C.A.R.’s minion robots are working inside a metal utility box. Any character who makes a successful DC 10 Knowledge (technology) check recognizes it as a utility box for high-speed internet connections. The robots are attempting to restore O.S.C.A.R.’s T3 connection to the outside world.

Minion Robots (2): hp 21, 21. See the new monster description at the end of this adventure for details.

Tactics: The robots need 2 more hours of work to finish repairing the connection. If they are hindered in any way, they turn on the intruders and attack, fighting until they are destroyed.

Development: A DC 15 Spot check reveals a plaque on the side of each robot that reads “Armitage.”

A small, concrete bunker serves as the entrance to the O.S.C.A.R. mainframe. The building has no windows, and the metal door is secured with an electronic lock. (Because all electronics are affected by the magical storm, however, the Disable Device check to open it is lower than normal; see below). A small plaque on the front of the building reads, “Property of Armitage Industries. NO TRESPASSING.”

A video camera above the door transmits images to O.S.C.A.R. Before the T3 connection was severed, it sent them back to Armitage Industries as well.

Door: Hardness 10, 120 hp, Break DC 35, Disable Device DC 15.

Video Camera: Hardness 5, 2 hp.

1. Entrance

Read or paraphrase the following aloud when the heroes open the door to the bunker.

The door opens to reveal a small antechamber with a steep metal stairwell leading down. The industrial lights in the stairwell flicker and pulse, sometimes even changing colors.

The lights in the stairwell are malfunctioning because of the effects of O.S.C.A.R.’s magic.

Trap: The entire floor of the entrance and first step of the stairwell are covered by a pressure pad that triggers a trap. If weight in excess of 5 kg is placed on the first step, the stairs fold over, creating a steep slope. Anyone on the stairs at this time must make a DC 25 Reflex save or slide at a speed of 50 feet per round, landing in a 30-foot-deep pit that opens just beneath the last step. The fall deals 3d6 points of damage (Reflex DC 13 half). A small hidden switch on the inside of the door (Search DC 25) deactivates the pressure pad. If the trap has been triggered, a flick of the switch restores the stairs to normal and resets the trap. An identical switch at the bottom of the stairs serves the same function.

2. Stairwell

The stairs descend 150 feet into the earth and are quite steep. If the characters deactivate the trap (see above), they can use the stairs normally.

The metal door is secured with a mechanical lock. A plaque on the door reads “O.S.C.A.R.”

Door: Hardness 10, hp 120, Break DC 35, Disable Device DC 25.

3. Tome Storage

Read or paraphrase the following aloud when the heroes enter this area.

This chamber is lit with dim red bulbs. Lining the walls are floor-to-ceiling metal racks, each filled with row after row of neatly aligned, ancient-looking books, tomes, and scrolls. Set into the far wall is a window that reveals a high-tech mainframe computer in the next room. The lights in that chamber are flickering just as those in the stairwell did. A conveyor belt leads from this chamber into the one with the mainframe.

Four robots stand in the middle of this room facing each other, their manipulator arms moving in complex gestures.                               }}                      \\
    \bottomrule
    \end{tabular}
    \caption{Robust Parsing Performance of HunyuanOCR in Wild-OmniDocBench.}
    \label{fig:omnidocbench}
\end{figure}

\renewcommand{\arraystretch}{1.5}
\begin{figure}[h]
    \centering
    \begin{tabular}{m{6cm}m{10cm}}
        \toprule
        \includegraphics[width=4.8cm]{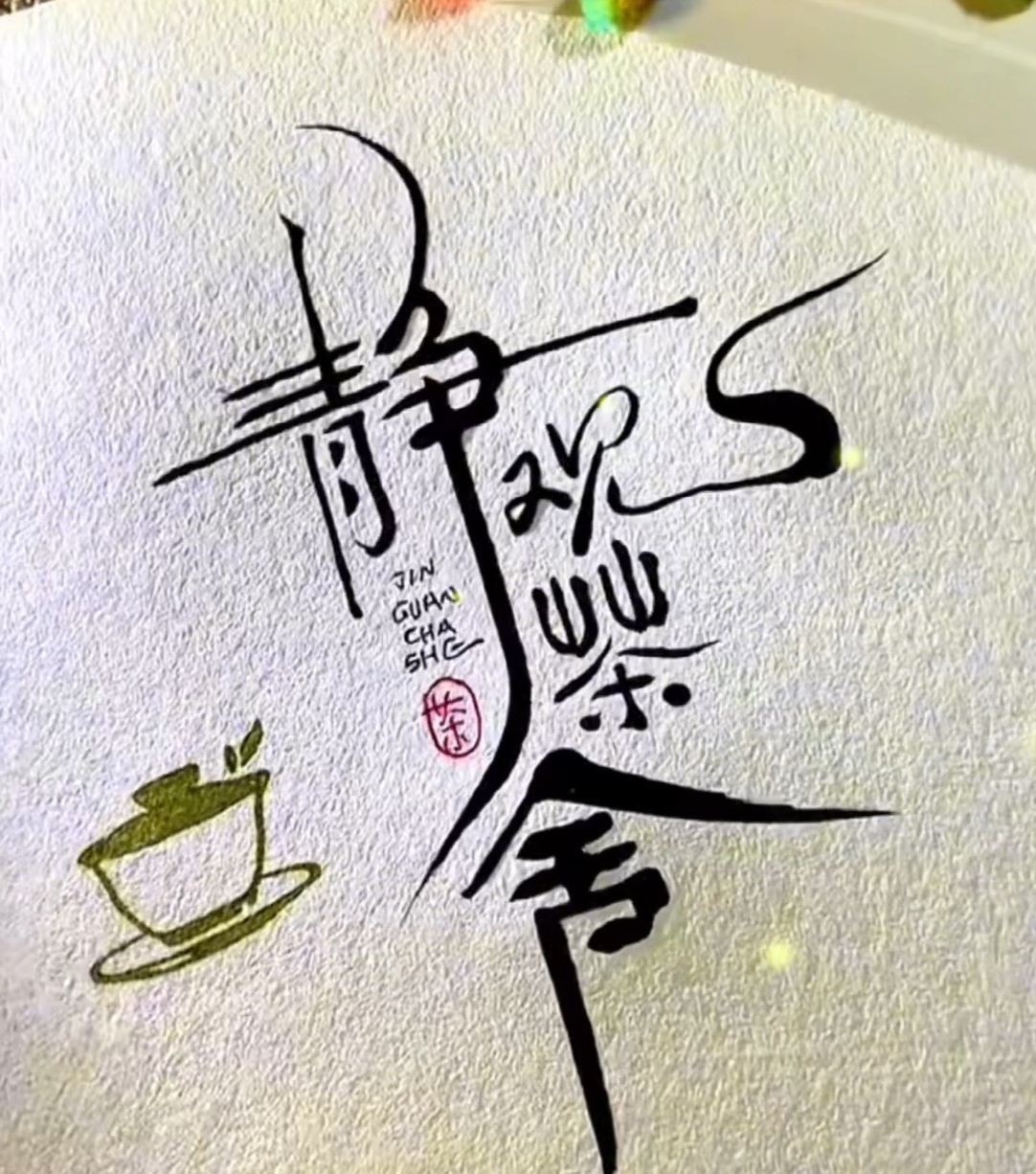}    &
        {
                \textbf{Prompt:}  {\footnotesize 提取图中的文字。}\par
                \vspace{0.5em}
                \textbf{HunyuanOCR:} 静观茶舍 \par JIN GUAN CHA SHE \par 茶

        }                                                                          \\
        \midrule
        \includegraphics[width=4.8cm]{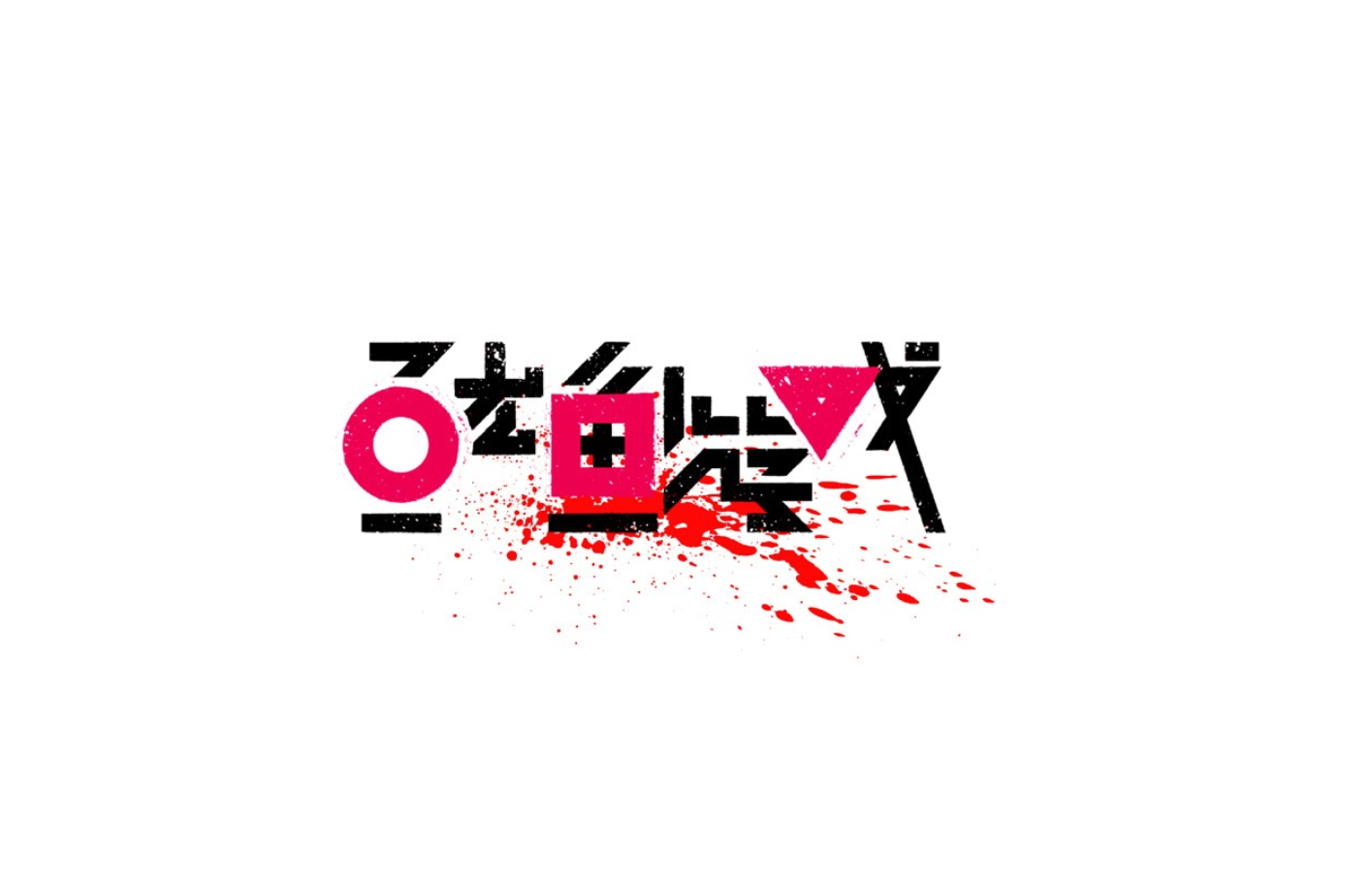} &
        {
                \textbf{Prompt:}  {\footnotesize 提取图中的文字。}\par
                \vspace{0.5em}
                \textbf{HunyuanOCR:} 鱿鱼游戏

        }                                                                          \\
        \midrule
        \includegraphics[width=4.8cm]{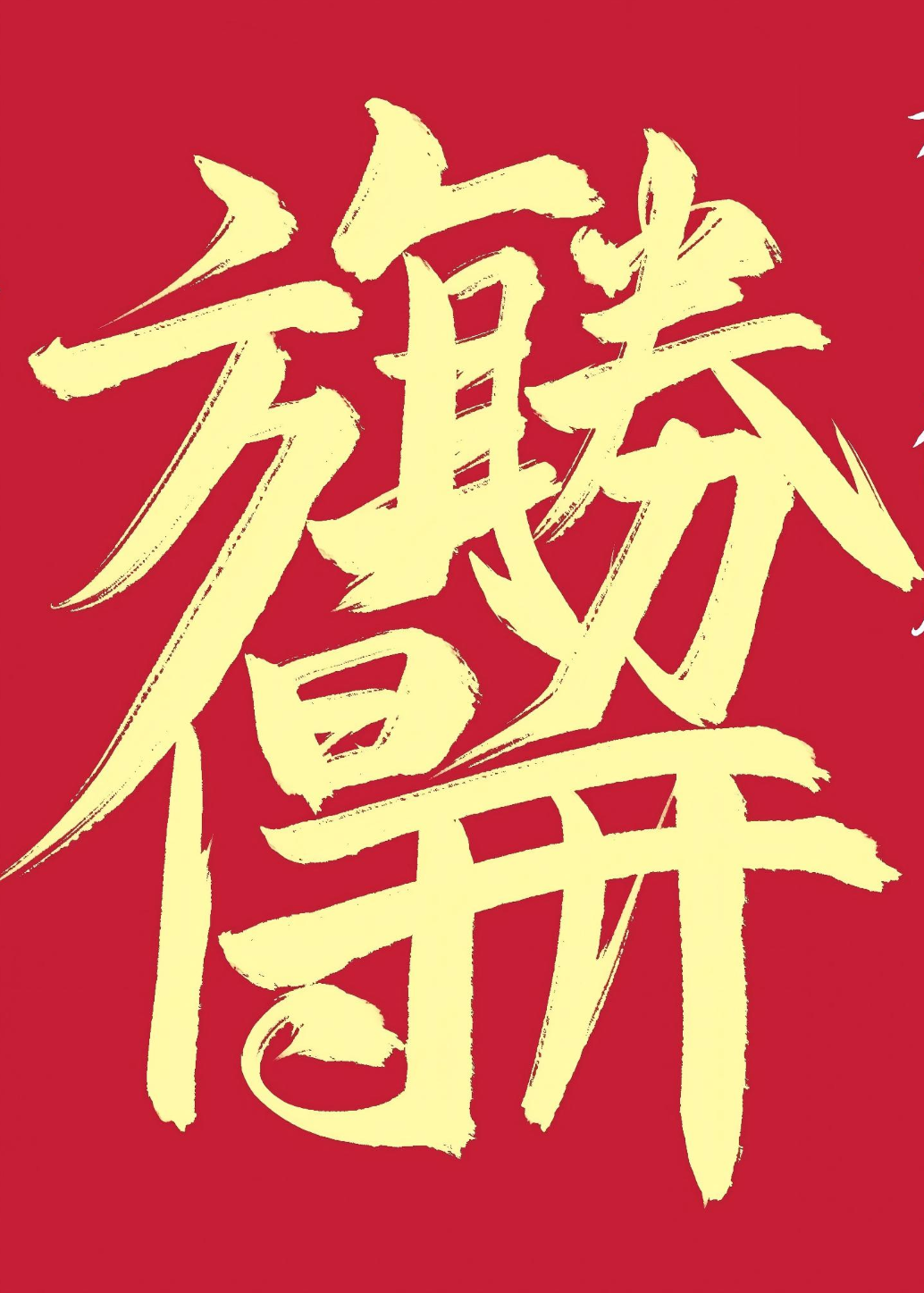}          &
        {
                \textbf{Prompt:}  {\footnotesize 提取图中的文字。}\par
                \vspace{0.5em}
                \textbf{HunyuanOCR:} \begin{CJK}{UTF8}{bsmi} 旗\end{CJK}开得\begin{CJK}{UTF8}{bsmi}勝\end{CJK}

        }                                                                          \\
        \midrule
        \includegraphics[width=4.8cm]{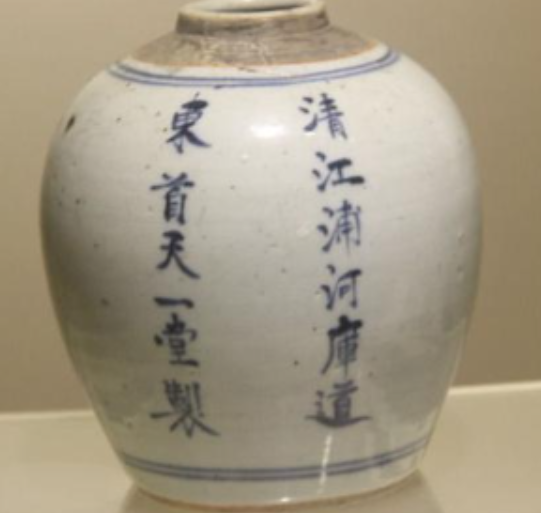}          &
        {
                \textbf{Prompt:}  {\footnotesize 提取图中的文字。}\par
                \vspace{0.5em}
                \textbf{HunyuanOCR:} \begin{CJK}{UTF8}{bsmi} 清江浦河庫道 \par 東首天一堂製 \end{CJK}

        }                                                                          \\

        \bottomrule
    \end{tabular}
    \caption{Robust Parsing Performance of HunyuanOCR in Diverse Real-World Visual Scenarios.}
    \label{fig:vis_real_world}
\end{figure}

\renewcommand{\arraystretch}{1.5}
\begin{figure}[h]
    \centering
    \begin{tabular}{m{6cm}m{10cm}}
        \toprule
        \includegraphics[width=4.8cm]{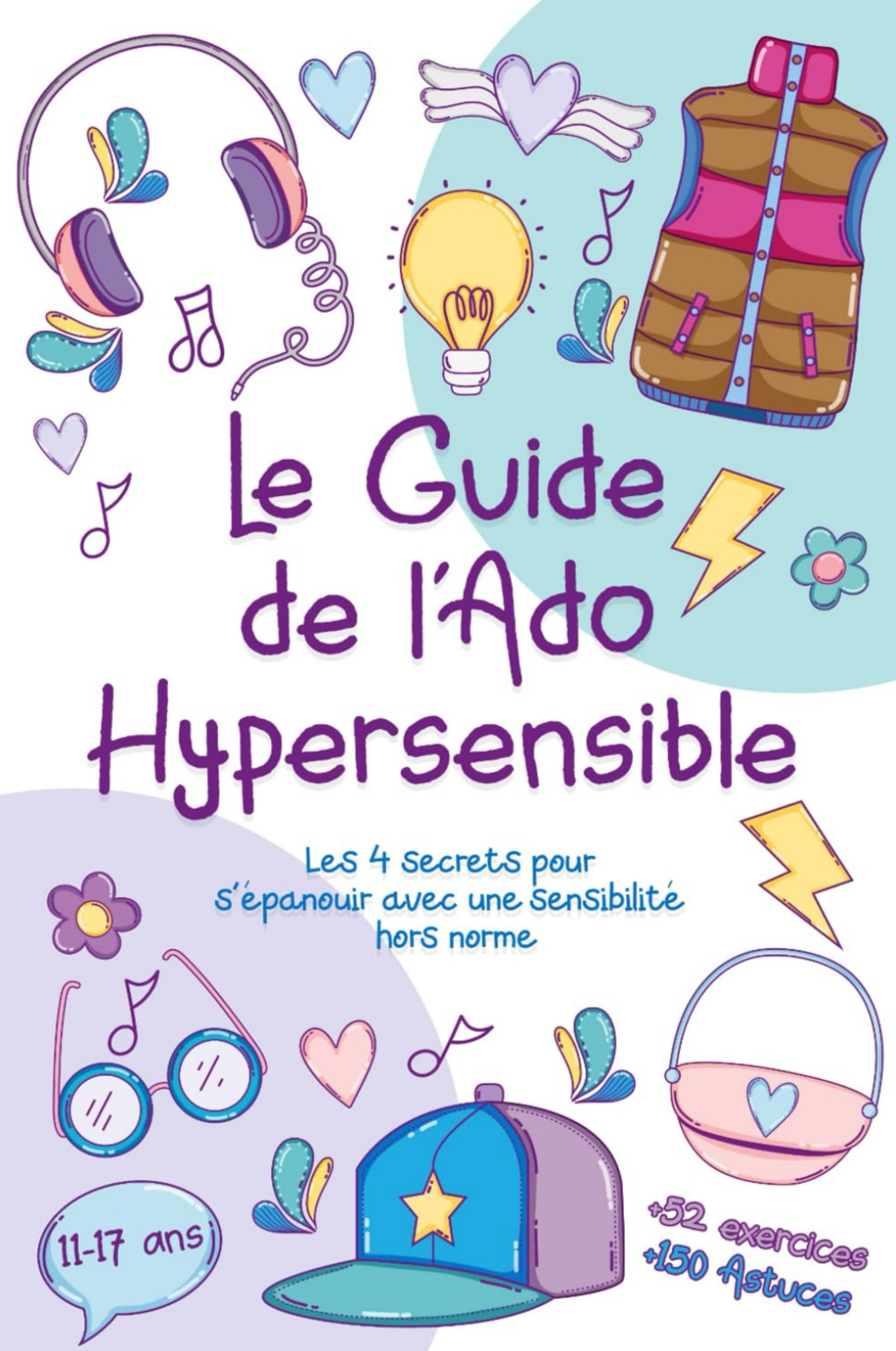}    &
        {
                \textbf{Prompt:}  {\footnotesize 提取图中的文字。}\par
                \vspace{0.5em}
                \textbf{HunyuanOCR:} \# Le Guide de l'Ado Hypersensible\par

Les 4 secrets pour s'épanouir avec une sensibilité hors norme\par

11-17 ans\par

+52 exercices +150 Astuces

        }                                                                          \\
        \midrule
        \includegraphics[width=4.8cm]{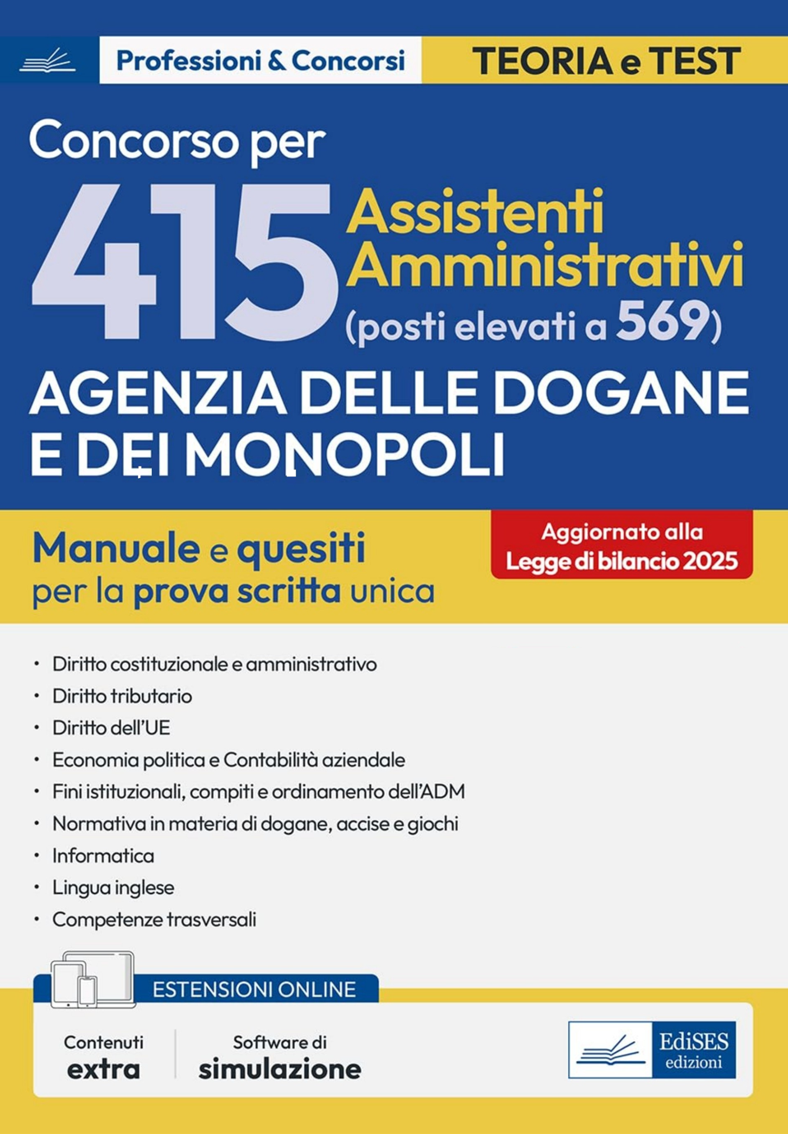} &
        {
                \textbf{Prompt:}  {\footnotesize 提取图中的文字。}\par
                \vspace{0.5em}
                \textbf{HunyuanOCR:} 
                {
            \scriptsize

            Professioni \& Concorsi

TEORIA e TEST

\#\#\# Concorso per

\# 415 Assistenti Amministrativi (posti elevati a 569)

\#\# AGENZIA DELLE DOGANE E DEI MONOPOLI

Manuale e quesiti per la prova scritta unica

Aggiornato alla Legge di bilancio 2025

Diritto costituzionale e amministrativo  

Diritto tributario  

Diritto dell’UE  

Economia politica e Contabilità aziendale 

Fini istituzionali, compiti e ordinamento dell’ADM  

Normativa in materia di dogane, accise e giochi  

Informatica

Lingua inglese  

Competenze trasversali

ESTENSIONI ONLINE

Contenuti extra

Software di simulazione

EdiSES edizioni
        }

        }                                                                          \\
        \midrule
        \includegraphics[width=4.8cm]{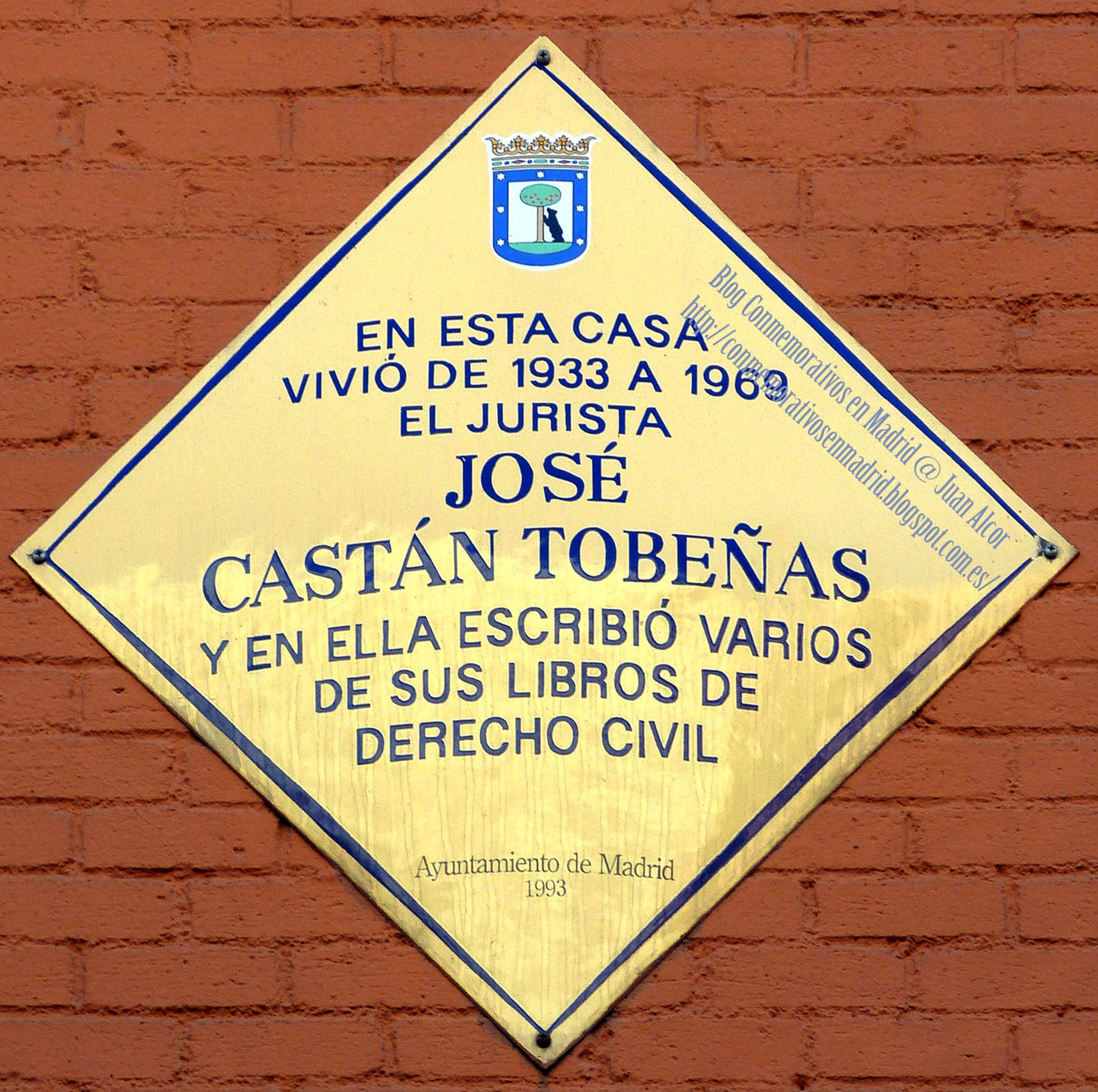}          &
        {
                \textbf{Prompt:}  {\footnotesize 提取图中的文字。}\par
                \vspace{0.5em}
                \textbf{HunyuanOCR:} EN ESTA CASA VIVIÓ DE 1933 A 1969 EL JURISTA JOSÉ CASTÁN TOBEÑAS Y EN ELLA ESCRIBIÓ VARIOSES DE SUS LIBROS DE DERECHO CIVIL\par

Ayuntamiento de Madrid  \par
1993\par

Blog Conmemorativos en Madrid @ Juan Alcor\par

http://conmemorativosenmadrid.blogspot.com.es/

        }                                                                          \\





        \bottomrule
    \end{tabular}
    \caption{Robust Parsing Performance of HunyuanOCR in DocML.}
    \label{fig:vis_ml}
\end{figure}

\begin{figure}[!h]
    \centering
    \begin{tabular}{@{}c@{\hspace{0.5cm}}c@{}}
        \begin{minipage}{0.45\textwidth}
            \centering
            Input image \\
            \includegraphics[height=0.92\textwidth]{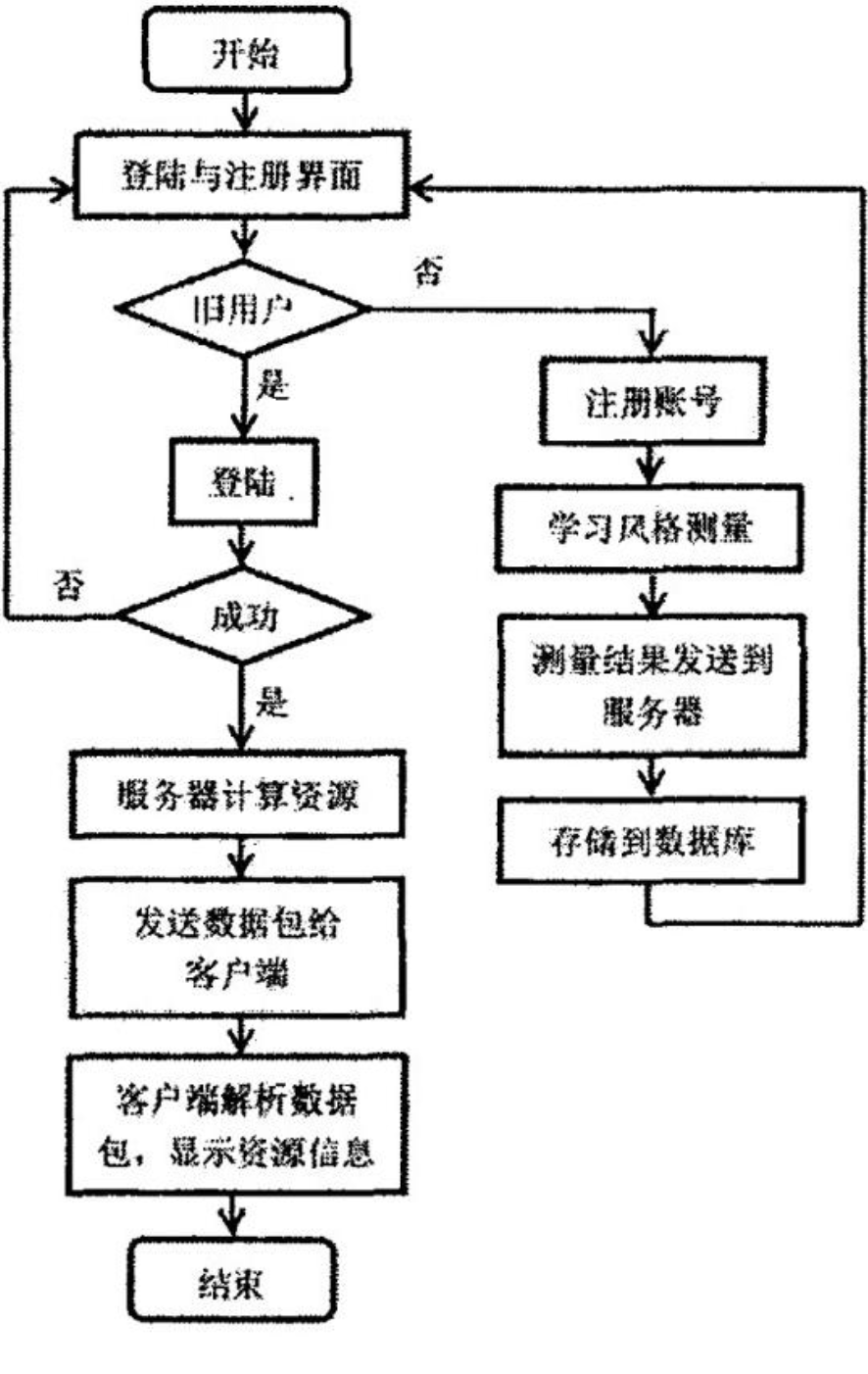}
        \end{minipage}
         &
        \begin{minipage}{0.45\textwidth}
            \centering
            Visualization of HunyuanOCR output \\
            \includegraphics[height=0.92\textwidth]{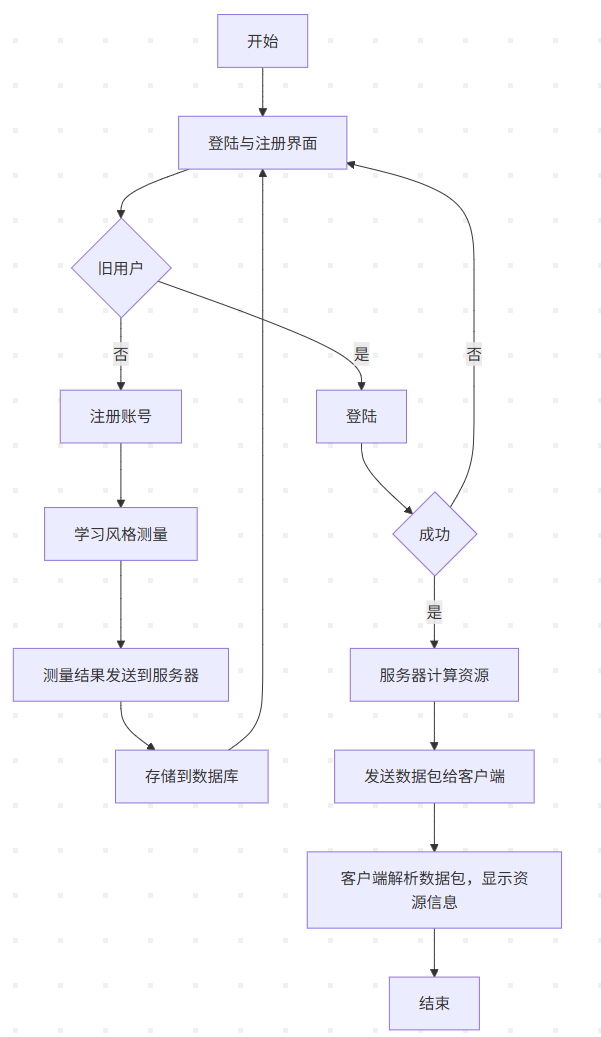}
        \end{minipage}
    \end{tabular}

    \vspace{0.5cm}
    \begin{minipage}{0.9\textwidth}
        \textbf{Prompt:} 解析图中的图表，对于流程图使用 Mermaid 格式表示，其他图表使用 Markdown 格式表示。\\
        \\
        \textbf{HunyuanOCR:}
        {
            \scriptsize

            flowchart TD

            A["开始"]

            B["登陆与注册界面"]

            C\{"旧用户"\}

            D["注册账号"]

            E["学习风格测量"]

            F["测量结果发送到服务器"]

            G["存储到数据库"]

            H["登陆"]

            I\{"成功"\}

            J["服务器计算资源"]

            K["发送数据包给客户端"]

            L["客户端解析数据包，显示资源信息"]

            M["结束"]

            A \texttt{--}\textgreater B

            B \texttt{--}\textgreater C

            C \texttt{--}\textgreater\textbar{}是\textbar{} H

            C \texttt{--}\textgreater\textbar{}否\textbar{} D

            D \texttt{--}\textgreater E

            E \texttt{--}\textgreater F

            F \texttt{--}\textgreater G

            G \texttt{--}\textgreater B

            H \texttt{--}\textgreater I

            I \texttt{--}\textgreater\textbar{}是\textbar{} J

            I \texttt{--}\textgreater\textbar{}否\textbar{} B

            J \texttt{--}\textgreater K

            K \texttt{--}\textgreater L

            L \texttt{--}\textgreater M
        }
        \\
    \end{minipage}
    \caption{Robust Parsing Performance of HunyuanOCR in Flowchart Scenarios.}
    \label{fig:vis_flowchart}
\end{figure}

\begin{figure}[!h]
    \centering
    \hrule
    \vspace{0.3cm}
    \begin{tabular}{@{}c@{\hspace{0.5cm}}c@{}}
        \begin{minipage}{0.45\textwidth}
            \centering
            Input image \\
            \includegraphics[height=0.52\textwidth]{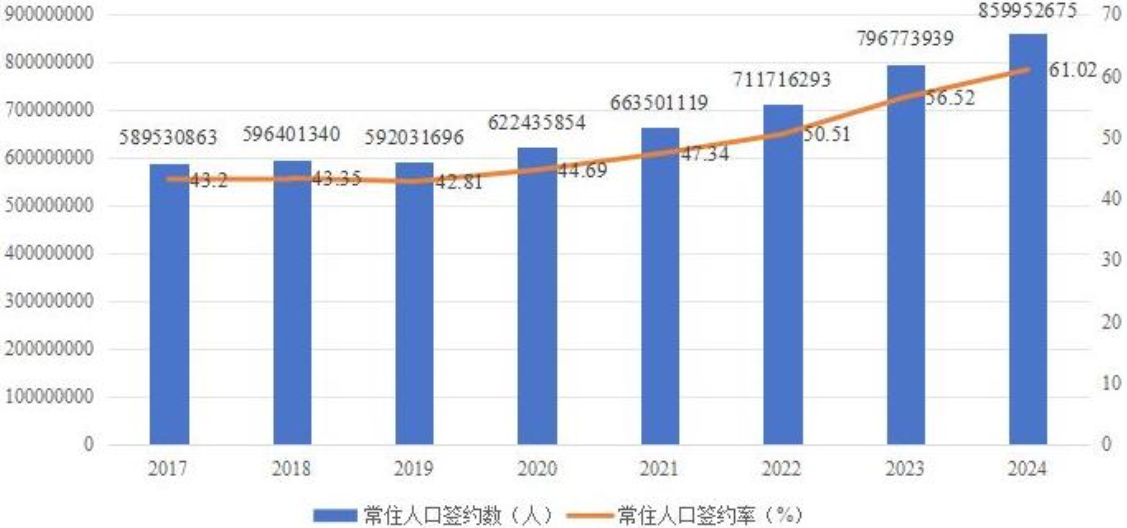}
        \end{minipage}
         &
        \begin{minipage}{0.45\textwidth}
            \centering
            Visualization of HunyuanOCR output \\
            \includegraphics[height=0.52\textwidth]{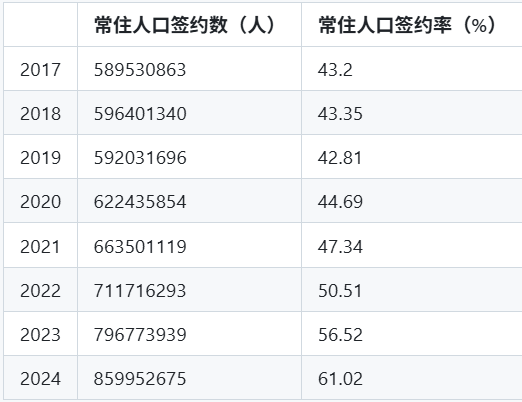}
        \end{minipage}
    \end{tabular}

    \vspace{0.5cm}
    \begin{minipage}{0.9\textwidth}
        \textbf{Prompt:} 解析图中的图表，对于流程图使用 Mermaid 格式表示，其他图表使用 Markdown 格式表示。\\
        \\
        \textbf{HunyuanOCR:}
        {
            \scriptsize

            \textbar{}  \textbar{} 常住人口签约数（人） \textbar{} 常住人口签约率（\%） \textbar{}

            \textbar{} \texttt{---} \textbar{} \texttt{---} \textbar{} \texttt{---} \textbar{}

            \textbar{} 2017 \textbar{} 589530863 \textbar{} 43.2 \textbar{}

            \textbar{} 2018 \textbar{} 596401340 \textbar{} 43.35 \textbar{}

            \textbar{} 2019 \textbar{} 592031696 \textbar{} 42.81 \textbar{}

            \textbar{} 2020 \textbar{} 622435854 \textbar{} 44.69 \textbar{}

            \textbar{} 2021 \textbar{} 663501119 \textbar{} 47.34 \textbar{}

            \textbar{} 2022 \textbar{} 711716293 \textbar{} 50.51 \textbar{}

            \textbar{} 2023 \textbar{} 796773939 \textbar{} 56.52 \textbar{}

            \textbar{} 2024 \textbar{} 859952675 \textbar{} 61.02 \textbar{}
        } \\
        
    \end{minipage}

    \vspace{0.3cm}
    \hrule
    \vspace{0.5cm}
    
    \begin{tabular}{@{}c@{\hspace{0.5cm}}c@{}}
        \begin{minipage}{0.45\textwidth}
            \centering
            Input image \\
            \includegraphics[height=0.52\textwidth]{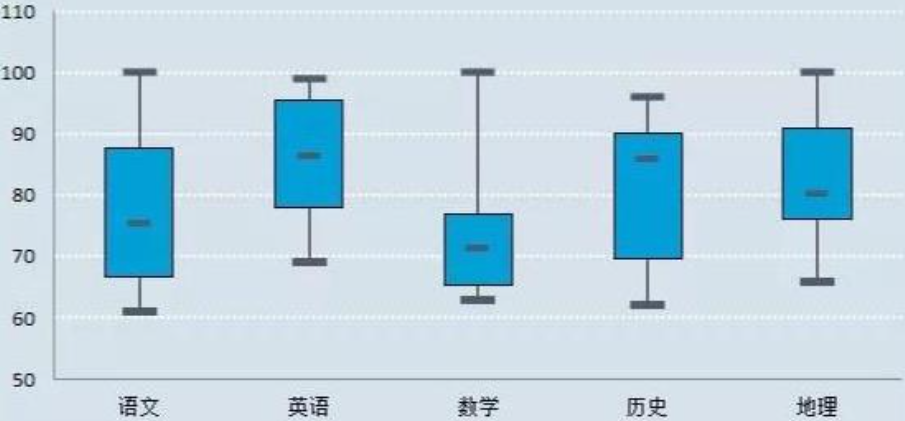}
        \end{minipage}
         &
        \begin{minipage}{0.45\textwidth}
            \centering
            Visualization of HunyuanOCR output \\
            \includegraphics[height=0.52\textwidth]{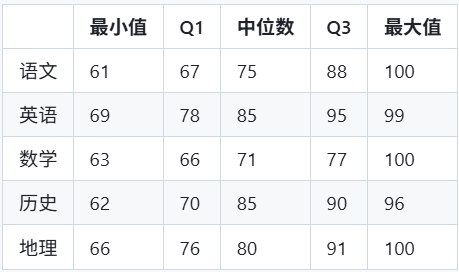}
        \end{minipage}
    \end{tabular}

    \vspace{0.5cm}
    \begin{minipage}{0.9\textwidth}
        \textbf{Prompt:} 解析图中的图表，对于流程图使用 Mermaid 格式表示，其他图表使用 Markdown 格式表示。\\
        \\
        \textbf{HunyuanOCR:}
        {
            \scriptsize

            \textbar{}  \textbar{} 最小值 \textbar{} Q1 \textbar{} 中位数 \textbar{} Q3 \textbar{} 最大值 \textbar{}

            \textbar{} \texttt{---} \textbar{} \texttt{---} \textbar{} \texttt{---} \textbar{} \texttt{---} \textbar{} \texttt{---} \textbar{} \texttt{---} \textbar{}

            \textbar{} 语文 \textbar{} 61 \textbar{} 67 \textbar{} 75 \textbar{} 88 \textbar{} 100 \textbar{}

            \textbar{} 英语 \textbar{} 69 \textbar{} 78 \textbar{} 85 \textbar{} 95 \textbar{} 99 \textbar{}

            \textbar{} 数学 \textbar{} 63 \textbar{} 66 \textbar{} 71 \textbar{} 77 \textbar{} 100 \textbar{}



            \textbar{} 历史 \textbar{} 62 \textbar{} 70 \textbar{} 85 \textbar{} 90 \textbar{} 96 \textbar{}

            \textbar{} 地理 \textbar{} 66 \textbar{} 76 \textbar{} 80 \textbar{} 91 \textbar{} 100 \textbar{}
        }
        \\
    \end{minipage}
    \hrule
    \vspace{0.3cm}
    \caption{Robust Parsing Performance of HunyuanOCR in Chart Scenarios.}
    \label{fig:vis_chart}
\end{figure}

%% file: sec/appendix/translation.tex
\renewcommand{\arraystretch}{1.5}
\begin{figure}[h]
    \centering
    \begin{tabular}{m{5cm}m{10cm}}
        \toprule
        \includegraphics[width=4.8cm]{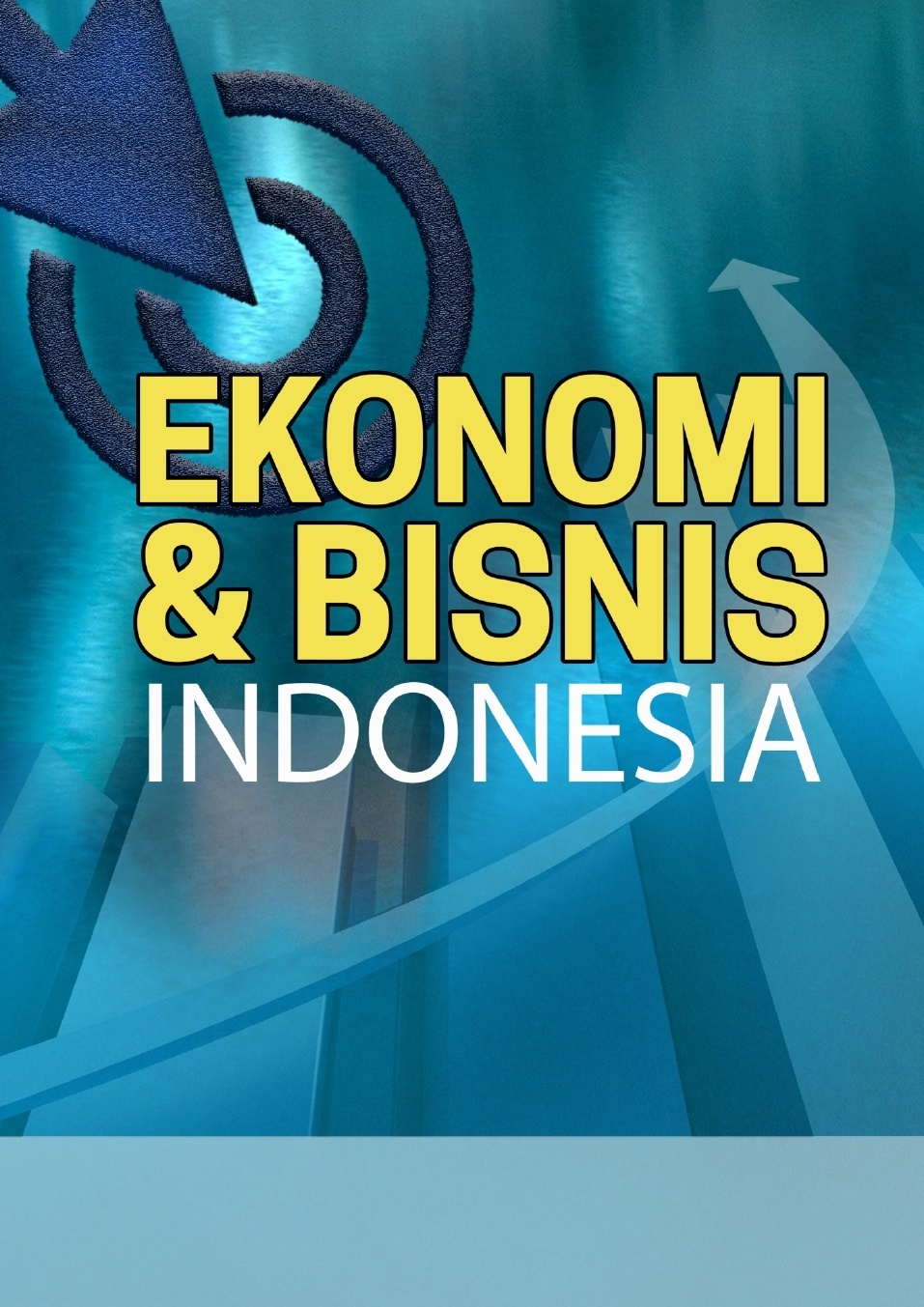}         &
        {
                \textbf{Prompt:}  {\footnotesize 将图中文字翻译为中文。}\par
                \vspace{0.5em}
                \textbf{HunyuanOCR:} 
                {
                经济与商业  
                
                印度尼西亚
                } 
        }                                                                   \\
        \midrule
        \includegraphics[width=4.8cm]{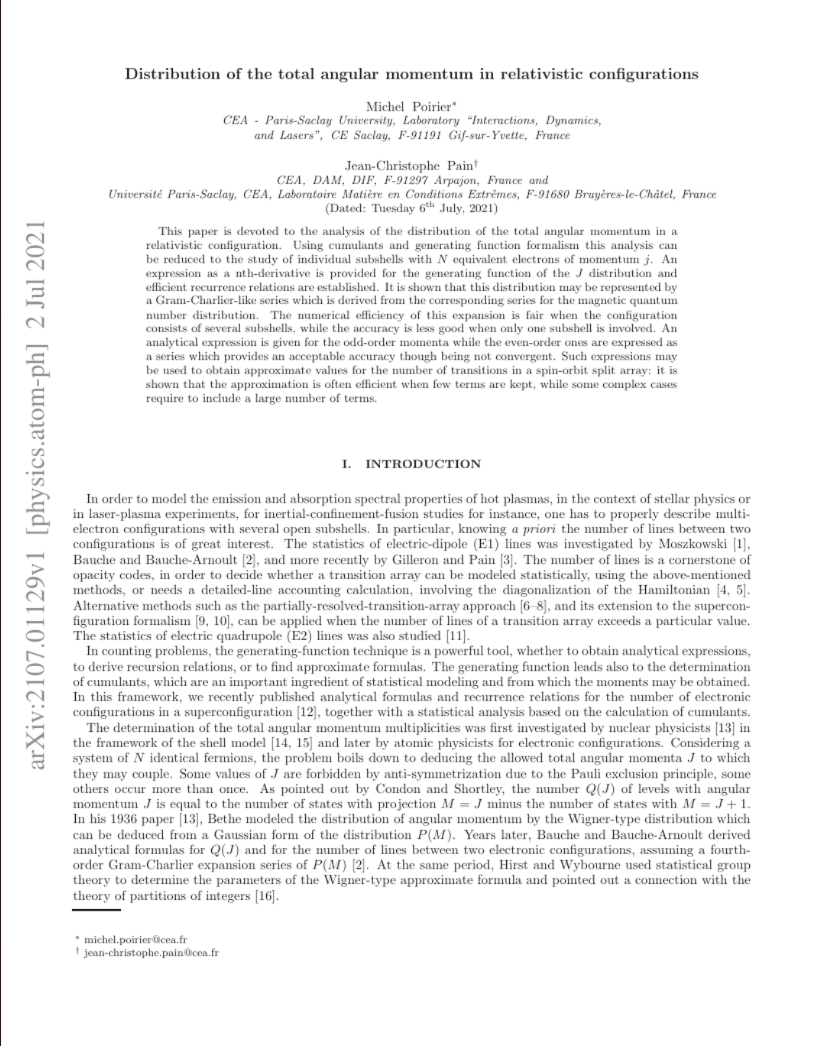} &
        {
        \textbf{Prompt:} {\footnotesize 先解析文档，再将文档内容翻译为中文，其中页眉、页脚忽略，公式用latex格式表示，表格用html格式表示。} \par
        \vspace{0.5em}
        \textbf{HunyuanOCR:}
         {\scriptsize
        ... 
        
翻译成中文为

\#\# 相对论构型中总角动量的分布

Michel Poirier*

CEA - 巴黎-萨克雷大学，“相互作用、动力学和激光”实验室，CE Saclay, F-91191 Gif-sur-Yvette, 法国

Jean-Christophe Pain†

CEA, DAM, DIF, F-91297 Arpajon, 法国 和

巴黎-萨克雷大学, CEA, 极端条件物质实验室, F-91680 Bruyères-le-Châtel, 法国

(2021年7月$ 6^{th} $日 星期二)

本文致力于分析相对论构型中总角动量的分布。利用累积量和生成函数形式，该分析可以简化为对具有 $ N $ 个等效动量为 $ j $ 的电子的单个子壳层进行研究。为 $ J $ 分布的生成函数提供了 $ n $ 阶导数形式的表达式，并建立了有效的递推关系。结果表明，该分布可以用一种类似 Gram-Charlier 的级数来表示，该级数来源于磁量子数分布的相应级数。当构型包含多个子壳层时，此展开的数值效率尚可，而当只涉及一个子壳层时，精度则较差。给出了奇数阶矩的解析表达式，而偶数阶矩则表示为级数，虽然不收敛，但提供了可接受的精度。此类表达式可用于获得自旋轨道分裂阵列中跃迁数的近似值：结果表明，当保留的项数较少时，该近似通常是有效的，而某些复杂情况则需要包含大量项。

\#\# I. 引言

为了在恒星物理学或激光等离子体实验（例如惯性约束聚变研究）的背景下模拟热等离子体的发射和吸收光谱特性，需要适当地描述具有多个开放子壳层的多电子构型。特别是，\textit{先验地}了解两种构型之间的谱线数量具有重要意义。电偶极子 (E1) 谱线的统计特性由 Moszkowski [1]、Bauche 和 Bauche-Arnoult [2] 以及最近由 Gilleron 和 Pain [3] 研究。谱线数量是不透明度代码的基石，用于决定是使用上述方法对跃迁阵列进行统计建模，还是需要涉及哈密顿量对角化的详细谱线计算 [4, 5]。当跃迁阵列的谱线数量超过特定值时，可以应用部分分辨跃迁阵列方法 [6–8] 及其对超构型形式的扩展 [9, 10] 等替代方法。电四极子 (E2) 谱线的统计特性也得到了研究 [11]。

在计数问题中，生成函数技术是一种强大的工具，无论是为了获得解析表达式、推导递推关系还是寻找近似公式。生成函数还能确定累积量，累积量是统计建模的重要组成部分，可以从中获得矩。在此框架下，我们最近发表了超构型中电子构型数量的解析公式和递推关系 [12]，以及基于累积量计算的统计分析。

总角动量多重性的确定最早由核物理学家 [13] 在壳模型 [14, 15] 框架内进行研究，后来由原子物理学家用于电子构型。考虑一个包含 $ N $ 个相同费米子的系统，问题归结为推导它们可以耦合的允许的总角动量 $ J $。由于泡利不相容原理导致的反对称性，某些 $ J $ 值是被禁止的，而另一些则出现多次。正如 Condon 和 Shortley 指出的那样，具有角动量 $ J $ 的能级数量 $ Q(J) $ 等于投影 $ M=J $ 的态的数量减去投影 $ M=J+1 $ 的态的数量。在1936年的论文 [13] 中，Bethe 通过 Wigner 型分布对角动量分布进行建模，该分布可以从 $ P(M) $ 分布的高斯形式推导出来。多年后，Bauche 和 Bauche-Arnoult 推导出了 $ Q(J) $ 和两种电子构型之间谱线数量的解析公式，假设 $ P(M) $ 采用四阶 Gram-Charlier 展开级数 [2]。同期，Hirst 和 Wybourne 使用统计群论来确定 Wigner 型近似公式的参数，并指出了与整数划分理论的联系 [16]。                                   }}                      \\
    \bottomrule
    \end{tabular}
    \caption{Translation Performance of HunyuanOCR.}
    \label{fig:translation}
\end{figure}

\renewcommand{\arraystretch}{1}
\clearpage

%% file: sec/appendix/ie_vqa.tex

\renewcommand{\arraystretch}{1.5}
\begin{figure}[h]
    \centering
    \begin{tabular}{m{5cm}m{8cm}}
        \toprule
        \\
        \includegraphics[width=4cm]{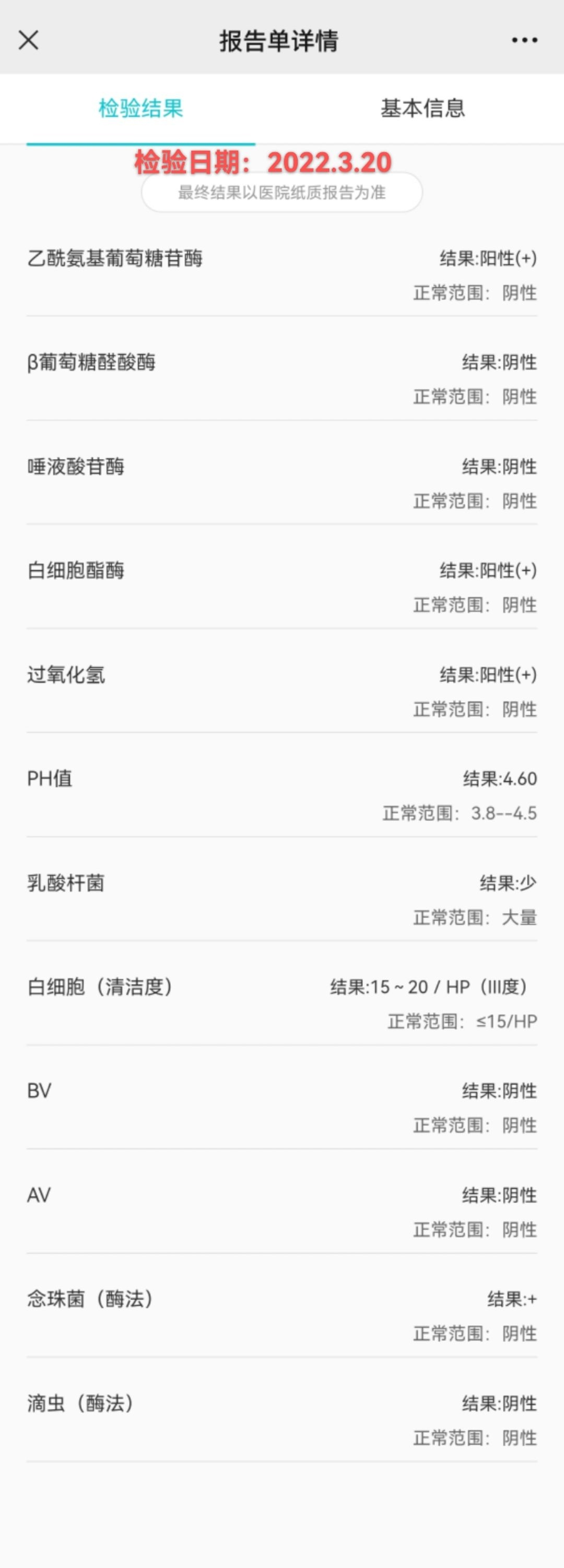}         &
        {
                \textbf{Prompt:} 输出检验日期对应的值 \par
                \vspace{0.5em}
                \textbf{HunyuanOCR:} 2022.03.20
        }                                                                   \\
        \midrule
        \includegraphics[width=4cm]{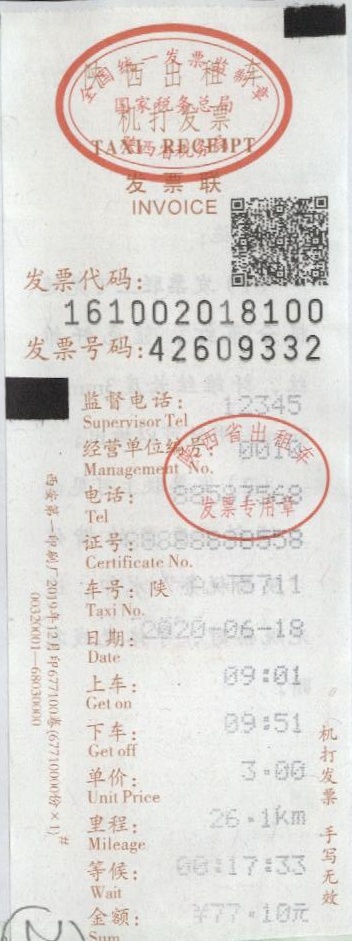} &
        {
        \textbf{Prompt:} 提取图片中的：[`单价', `上车时间', `发票号码', `省前缀', `总金额', `发票代码', `下车时间', `里程数'] 的字段内容，并且按照JSON格式返回。 \par
        \vspace{0.5em}
        \textbf{HunyuanOCR:}
    ```json
        \{\par
        ``单价'': ``3.00'',\par
        ``上车时间'': ``09:01'',\par
        ``发票号码'': ``42609332'',\par
        ``省前缀'': ``陕'',\par
        ``总金额'': ``￥77.10元'',\par
        ``发票代码'': ``161002018100'',\par
        ``下车时间'': ``09:51'',\par
        ``里程数'': ``26.1km''\par
        \}
        }                                                                       \\
        \bottomrule
    \end{tabular}
    \caption{Information Extraction (IE) Performance of HunyuanOCR on Receipts.}
    \label{fig:ie2}
\end{figure}
\renewcommand{\arraystretch}{1}
\clearpage

\renewcommand{\arraystretch}{1.5}
\begin{figure}[h]
    \centering
    \begin{tabular}{m{7cm}m{6cm}}
        \toprule
        \\
        \includegraphics[width=7cm]{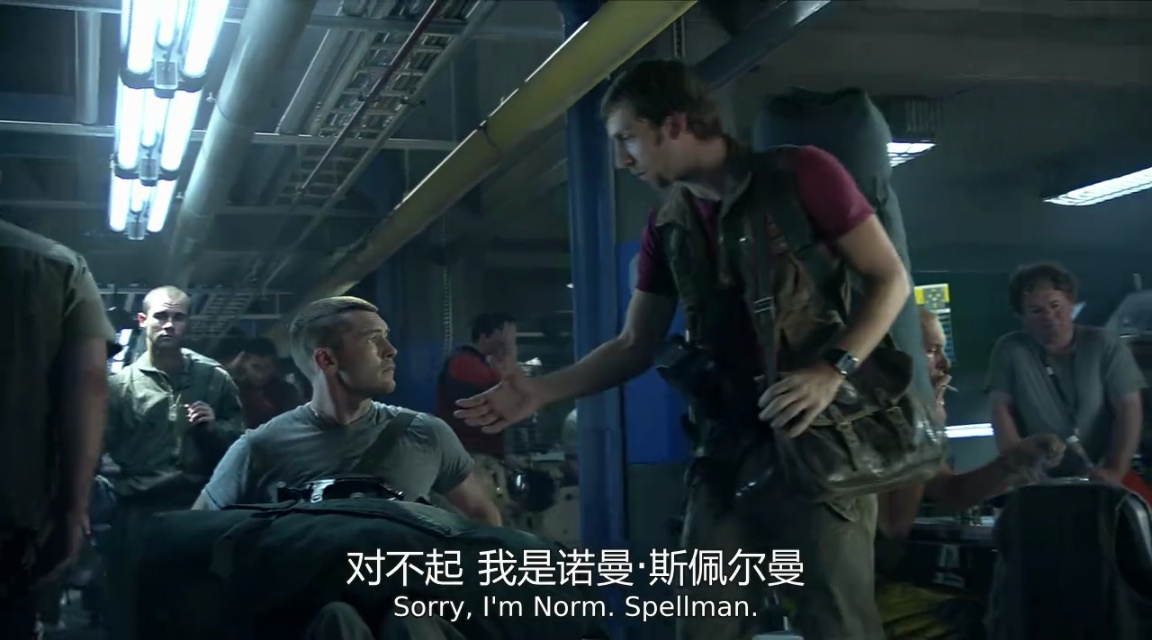} & {
                \textbf{Prompt:} 提取图中的字幕。 \par
                \vspace{0.5em}
                \textbf{HunyuanOCR:} 对不起 我是诺曼.斯佩尔曼  \par Sorry, I'm Norm. Spellman.
        }    \\                                                            
        \midrule
        \\
        \includegraphics[width=7cm]{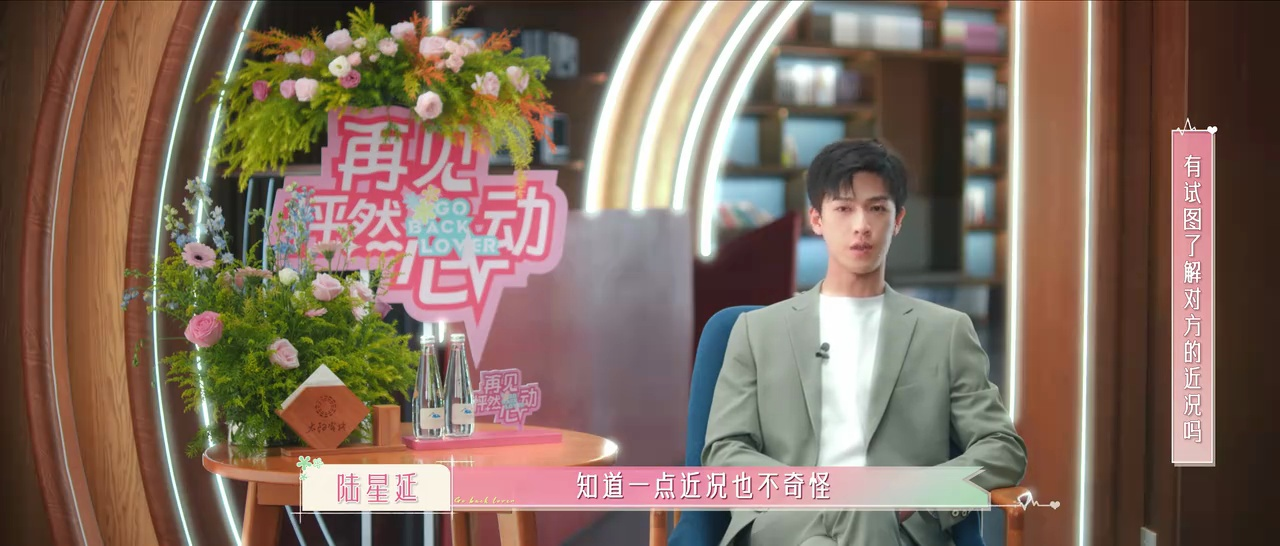} & {
                \textbf{Prompt:} 提取图中的字幕。 \par
                \vspace{0.5em}
                \textbf{HunyuanOCR:} 知道一点近况也不奇怪 \par
        }                                                               \\
        \midrule
        \\

        \includegraphics[width=7cm]{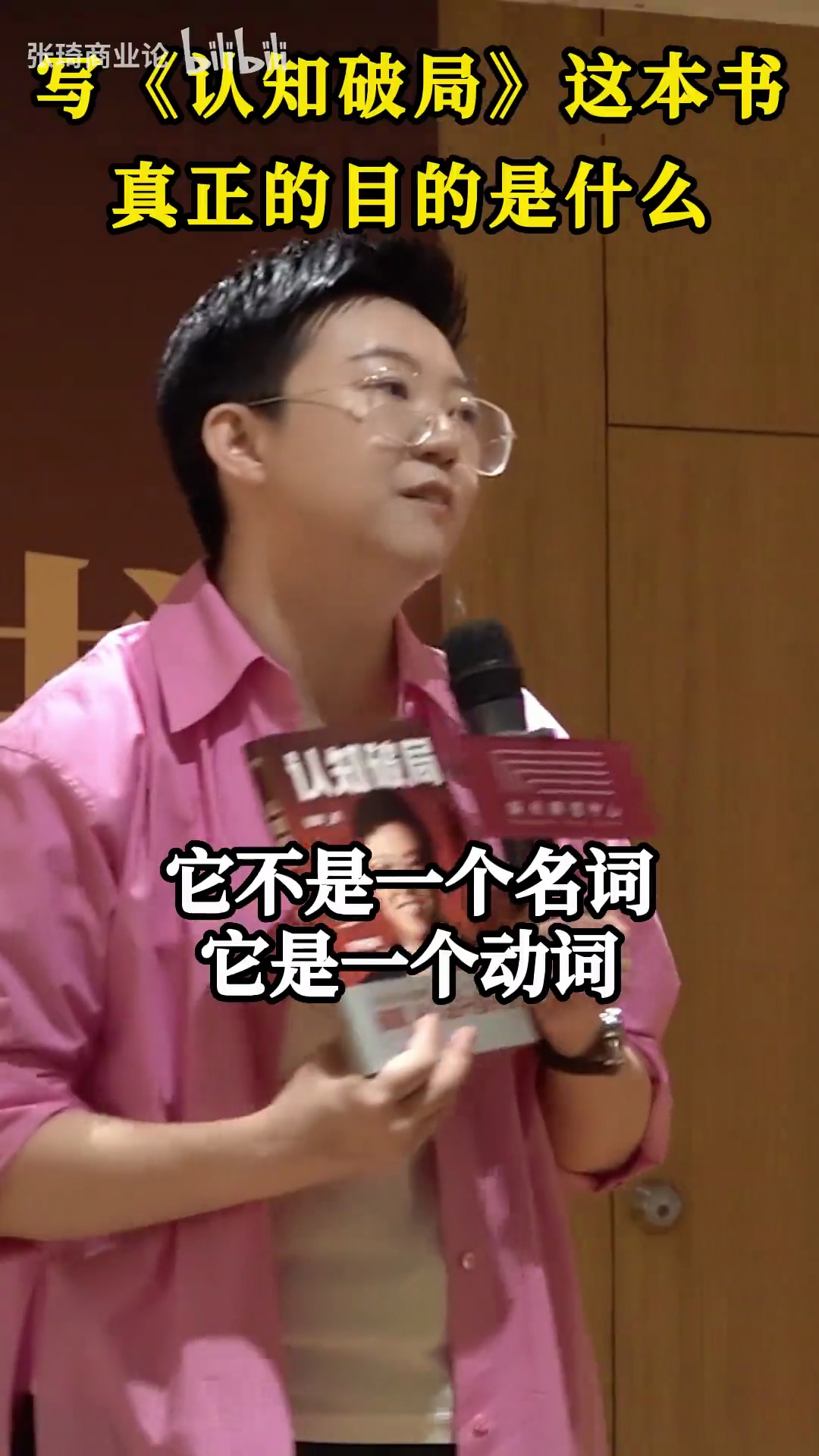} & {
                \textbf{Prompt:} 提取图中的字幕。 \par
                \vspace{0.5em}
                \textbf{HunyuanOCR:} 它不是一个名词\par它是一个动词
        }                                                               \\
        \bottomrule
    \end{tabular}
    \caption{Video Subtitle Extraction Performance of HunyuanOCR.}
    \label{fig:ie3}
\end{figure}
\renewcommand{\arraystretch}{1}
\clearpage







\renewcommand{\arraystretch}{1.5}
\begin{figure}[h]
    \centering
    \begin{tabular}{m{8cm}m{7cm}}
        \toprule
        \\
        \includegraphics[width=8cm]{imgs/appendix/IE_VQA/gzyh0227_9.png}      &
        {
                \textbf{Prompt:} What is the factory name? \par
                \vspace{0.5em}
                \textbf{HunyuanOCR:} Fort Morgan
        }                                                                       \\
        \midrule
        \\
        \includegraphics[width=8cm]{imgs/appendix/IE_VQA/multi_col_60150.png} &
        {
                \textbf{Prompt:} What is the highest life expectancy at birth of male? \par
                \vspace{0.5em}
                \textbf{HunyuanOCR:} 80.7.
        }                                                                       \\
        \midrule
        \\

        \includegraphics[width=8cm]{imgs/appendix/IE_VQA/two_col_23773.png}   &
        {
                \textbf{Prompt:} What platform did Samsung have the largest market share in 2018? \par
                \vspace{0.5em}
                \textbf{HunyuanOCR:} Tizen.
        }                                                                       \\
        \bottomrule
    \end{tabular}
    \caption{Document \& Chart Visual Question Answering (VQA) performance of HunyuanOCR.}
    \label{fig:ie6}
\end{figure}

\renewcommand{\arraystretch}{1}
\clearpage